\documentclass[10pt,twocolumn,letterpaper]{article}

\usepackage{iccv}              
\usepackage{times}
\usepackage{epsfig}
\usepackage{graphicx}
\usepackage{amsmath}
\usepackage{amssymb}
\usepackage{xcolor}
\usepackage{capt-of}
\usepackage{multirow}
\usepackage{booktabs}
\usepackage{tabularx}
\usepackage{amsmath}

\usepackage{algorithm}
\usepackage{algpseudocode}
\usepackage[export]{adjustbox} 
\usepackage[margin=1in]{geometry}
\usepackage{graphicx}
\usepackage{booktabs}  
\usepackage{caption}
\usepackage{float}
\usepackage{arydshln}
\usepackage{subcaption}
\usepackage{tikz}
\usepackage{graphicx}
\usepackage[dvipsnames]{xcolor}
\usepackage{graphicx}   
\usepackage{array}      

\newbox\jsavebox

\long\def\ignorethis#1{}

\usepackage[T1]{fontenc} 
\usepackage{iftex} 

\ifLuaTeX
\protected\def\pdfmapline {\pdfextension mapline }
\fi

\newcommand{\whitetxt}[1]{{\color{white}#1}\normalfont}

\pdfmapline{+delphine < Delphine.ttf <T1-WGL4.enc}

\usepackage{makecell}

\definecolor{iccvblue}{rgb}{0.21,0.49,0.74}
\usepackage[pagebackref=true,breaklinks=true,letterpaper=true,colorlinks,bookmarks=false]{hyperref}

\def\paperID{3206} 
\def\confName{ICCV}
\def\confYear{2025}

\title{Supercharging Floorplan Localization with Semantic Rays}

\author{
Yuval Grader\\
Tel Aviv University\\
\and
Hadar Averbuch-Elor\\
Cornell University\\
}

\begin{document}


\twocolumn[{
\maketitle
 \vspace{-35pt}
  \begin{center}
    \small\url{https://tau-vailab.github.io/SemRayLoc/}
  \end{center}
  \vspace{1em}
}]

\begin{abstract}
Floorplans provide a compact representation of the building’s structure, revealing not only layout information but also detailed semantics such as the locations of windows and doors. However, contemporary floorplan localization techniques  mostly focus on matching depth-based structural cues, ignoring the rich semantics communicated within floorplans. In this work, we introduce a semantic-aware localization framework that jointly estimates depth and semantic rays, consolidating over both for predicting a structural-semantic probability volume. Our probability volume is constructed in a coarse-to-fine manner: We first sample a small set of rays to obtain an initial low-resolution probability volume. We then refine these probabilities by performing a denser sampling only in high-probability regions and process the refined values for predicting a 2D location and orientation angle. We conduct an evaluation on two standard floorplan localization benchmarks. Our experiments demonstrate that our approach substantially outperforms state-of-the-art methods, achieving significant improvements in recall metrics compared to prior works. Moreover, we show that our framework can easily incorporate additional metadata such as room labels, enabling additional gains in both accuracy and efficiency. 
\end{abstract}

\section{Introduction}
\label{sec:intro}

\definecolor{gtarrow}{RGB}{255,0,255}

\begin{figure}[t]
  \centering

\begin{tabular}{@{} m{0.5cm} >{\centering\arraybackslash}m{0.9\linewidth} @{}}

    \hfill
    \rotatebox[origin=c]{90}{ Image} &
    \includegraphics[height=1.86cm]{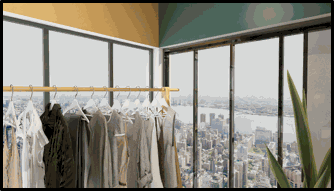} \hfill  
    \\[1ex]

    \hfill
    \rotatebox[origin=c]{90}{Raw Floorplan (w/ ~\cite{chen2024f3loc})} &
    \begin{tabular}{@{}c@{\quad}c@{}}
      \includegraphics[width=0.46\linewidth]{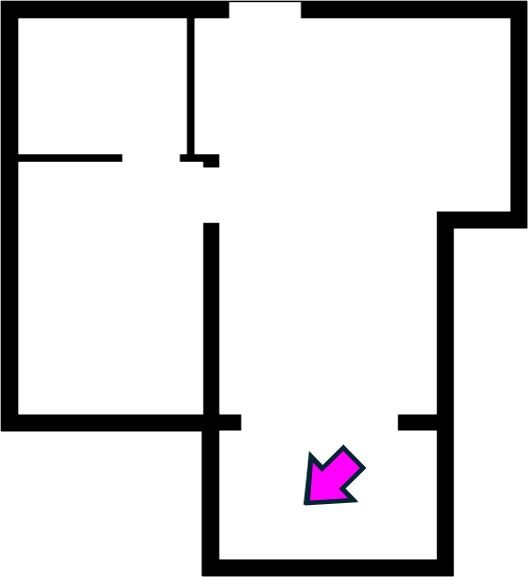} &
      \includegraphics[width=0.46\linewidth]{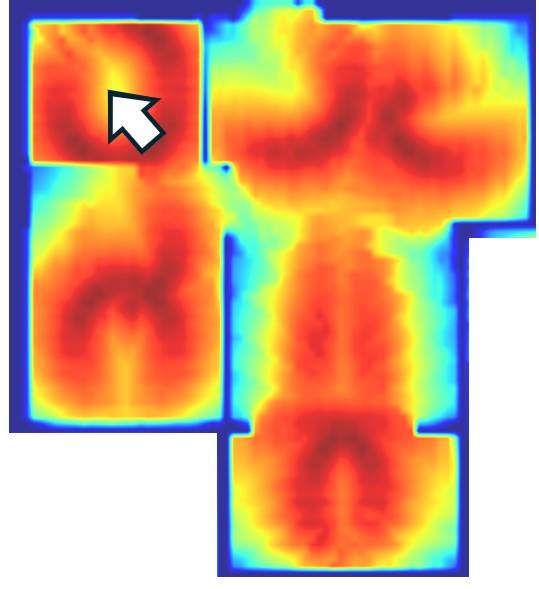}
      \hfill
    \end{tabular}
    \\[1ex]

    \hfill
    \rotatebox[origin=c]{90}{ \whitetxt{xxxxx}+\emph{Semantics} (w/ Ours)\whitetxt{p}} &
    \begin{minipage}[c]{0.46\linewidth}
      \hspace{-6pt}
      \includegraphics[width=\linewidth]{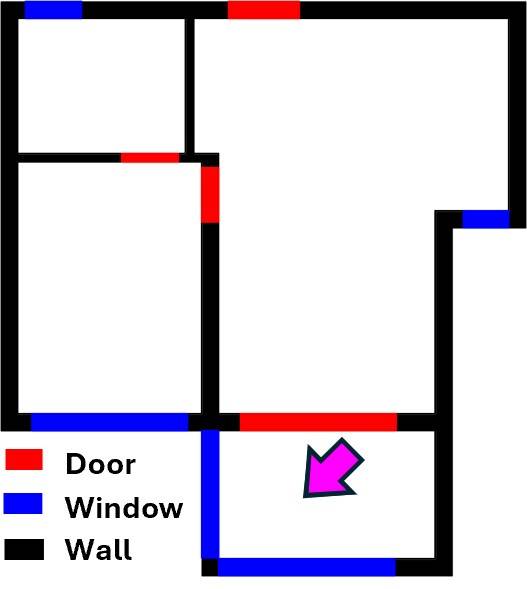}
      \par Input \& \colorbox{gtarrow}{GT}
    \end{minipage}
   \hspace{1pt}
    \begin{minipage}[c]{0.46\linewidth}
      \includegraphics[width=\linewidth]{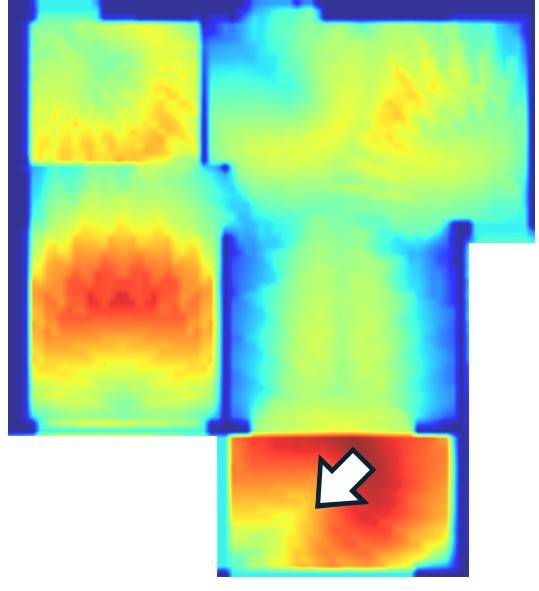}
      \par Output Localization
    \hfill
    \end{minipage}
    
    \\

  \end{tabular}
    \vspace{-15pt}
    \caption{Floorplan localization using a raw binary floorplan (middle row) often yields ambiguous predictions.
In this work, we utilize a richer, yet readily available,  representation: a floorplan enhanced with semantic labels (bottom row). We present an approach that supercharges floorplan localization with semantic rays, enabling for resolving localization ambiguities, as illustrated by the comparison on the right. 
    } 
  \label{fig:teaser}
\end{figure}

Camera localization is a longstanding problem in computer vision, with significant applications in 3D reconstruction~\cite{liu2017efficient,sattler2011fast,sattler2012improving,sattler2016efficient,sarlin2019coarse}, augmented reality~\cite{shotton2013scene,valentin2015exploiting,kendall2015posenet,glocker2013real,newcombe2011kinectfusion}, and navigation~\cite{walch2017image,wu2017delving,thoma2019mapping,niwa2022spatio,mathur2022sparse}.
Localization within indoor environments is especially challenging due to the absence of reliable GPS signals and the complexity of reasoning across multiple floors and layers. Hence, to bypass complicated 3D model-based localization techniques, prior work ~\cite{chu2015you,winterhalter2015accurate,chen2024f3loc,howard2021lalaloc,howard2022lalaloc++, karkus2018particle} has explored the problem of localizing camera observations within a provided 2D floorplan map by matching depth-based structural cues.

However, while floorplans offer a compact and lightweight scene representation,  structural cues within floorplans often correlate with multiple candidate locations, particularly for environments with repetitive or symmetric layouts.
Consider the example in Figure~\ref{fig:teaser}. \emph{Can you localize the input image within the floorplan?} Provided with just a \emph{raw} (walls only) floorplan, room corners are indistinguishable and hence localization is highly ambiguous, as can also be observed by the probabilities predicted by the state-of-the-art F3Loc~\cite{chen2024f3loc} technique (middle row, right). To resolve such ambiguities, we are interested in utilizing a slightly different representation: a \emph{semantics}-aware floorplan, such as the one illustrated on the bottom row of Figure \ref{fig:teaser}. 

Accordingly, we introduce a semantic‐aware ray‐based localization framework that integrates semantic cues with depth‐based predictions. In particular, we propose a semantic prediction network that predicts accurate semantic ray representations along with optional additional metadata (such as room labels) from a single RGB image with a limited field‐of‐view. We process these rays to compute a semantic probability volume, which is then fused with depth information for constructing a \emph{structural--semantic} probability volume. 

Our framework follows a coarse-to-fine localization strategy. We first operate over a low-resolution image for an efficient floorplan search. This initial search yields the Top-$k$ candidate locations. Finally, we refine these candidates and select the best match using a high-resolution ray representation. By adopting this coarse-to-fine approach, our method effectively constrains the localization search to the most promising regions while keeping the computation cost feasible.

Our experiments demonstrate that our approach yields improvements by factors ranging from two to three across most metrics compared to the state-of-the-art technique~\cite{chen2024f3loc}, upon which our method is built. This substantial gain underscores the effectiveness of fusing semantic and depth ray predictions into a unified probability volume. We further show that our coarse-to-fine strategy offers a flexible tradeoff between accuracy and computational cost, with performance consistently improving as larger candidate sets (Top-$k$) are evaluated—making our method adaptable to diverse task requirements. Moreover, incorporating additional metadata further enhances precision, leading to significantly improved localization accuracies.

Explicitly stated, our contributions are: 
\begin{itemize}
    \item We introduce a  semantic ray prediction network that receives a single RGB image as input.
    \item We propose an efficient and unified framework that fuses semantic and depth ray predictions into a structural--semantic probability volume, which effectively resolves localization ambiguities.
    \item Results that demonstrate that significant improvements over state-of-the-art methods.
\end{itemize}
\section{Related Work}
\label{sec:related_work}
\noindent \textbf{Visual Localization}. The task of visual localization has received ongoing attention throughout the past several decades. Traditional approaches often rely on image retrieval or on a 3D Structure-from-Motion~(SfM) model of the environment. In the \emph{image retrieval} paradigm, methods such as NetVLAD~\cite{arandjelovic2016netvlad} or RelocNet~\cite{balntas2018relocnet} compare a query image against a database of labeled images. Once the closest match is found, the query pose is approximated by the retrieved image’s pose. 
Other methods explicitly construct a 3D SfM model of a scene to establish 2D--3D correspondences between the query image and the reconstructed 3D structure~\cite{liu2017efficient,sarlin2019coarse,sattler2011fast,sattler2012improving,sattler2016efficient}. After matching local image descriptors to 3D points, robust solvers estimate the 6-DoF pose.

Recent \emph{learning-based} pipelines deviate from classical 2D--3D matching. Scene-coordinate regression methods predict dense 3D coordinates for every pixel in the query image~\cite{brachmann2017dsac,shotton2013scene,valentin2015exploiting}, whereas pose regression methods directly estimate the 6-DoF camera pose via neural networks~\cite{kendall2015posenet,walch2017image,wu2017delving}. Although promising for single-scene scenarios, these methods must be retrained or fine-tuned to handle new environments.

\medskip \noindent
\textbf{Floorplan  Localization}. Prior work addressing the task of floorplan localization primarily focused on \emph{depth-based cues}, leveraging image-derived depth predictions or sensors, to match depth obtained from floorplans. In particular, LiDAR-based methods \cite{boniardi2017robust,boniardi2019pose,wang2019glfp,li2020online} utilize precise laser scans but restrict usability on most mobile devices. Alternative sources of geometric cues, including semi dense visual odometry (SDVO)~\cite{chu2015you}, or point clouds from depth cameras \cite{ito2014wrgdb}, can circumvent heavy LiDAR hardware. 

Earlier works compare extracted room edges directly to the 2D layout~\cite{boniardi2019robot}, often assuming knowledge of camera or room height \cite{boniardi2019robot,chu2015you}. Other approaches, are embedding RGB images and floorplans into a shared metric space in order to do the localization. For instance, LaLaLoc~\cite{howard2021lalaloc} uses panoramic depth layout image which is rendered at known heights at different locations within the floorplan, where the localization is done by doing similarity in the embedded space. LaLaLoc++~\cite{howard2022lalaloc++} removes the hight assumption by embedding the entire floorplan into the feature space.  However, these approaches often require an upright camera pose ~\cite{howard2021lalaloc,howard2022lalaloc++}, making them less flexible for hand-held or head-mounted devices.
F3Loc~\cite{chen2024f3loc} localizes by predicting depth rays from a given image and generating probability volumes that indicate the likelihood of each the depth prediction to a particular location on the floorplan. Although effectively leverages geometric cues from depth maps, it does not incorporate semantic information, which may reduce its robustness in environments with repetitive structures or occlusions.

\emph{Semantic-based cues} are less commonly used for indoor localization, but several works have utilized them for this task. Wang et al.\cite{wang2015lost} extracts scene texts from images in large indoor spaces and performs localization by matching this text to the floorplan. 
SeDAR~\cite{mendez2020sedar} uses a CNN to extract semantic labels from an image and perform Monte Carlo localization. By contrast to our single-image localization framework, it depends on sequential images and depth sensors. 
PF-Net~\cite{karkus2018particle} applies a differentiable particle filter with a learned observation model, relying on a computationally-intensive embedding process that yields limited performance for the single-image scenario. SPVLoc \cite{Gard2024_SPVLOC} matches captured images to panoramic renderings to estimate location, leveraging additional height information present in these renderings. Similarly, Kim et al.\ \cite{Kim2024_FullyGeometric} perform localization by using a pre-captured 3D map of the environment. Our approach does not assume the availibity of these additional cues.
%
LASER~\cite{min2022laser} treats the floorplan as a set of points and synthesizes view-dependent features, including the semantic label of each point, for matching purposes. They embed images into circular features, which are then compared to pose features in the same embedded space. By contrast, we model the semantics as fine-grained ray predictions. Furthermore, unlike these prior works, our approach can operate over images with non-zero pitch and roll. 


\begin{figure*}[t]
    \centering
    \includegraphics[width=\textwidth]{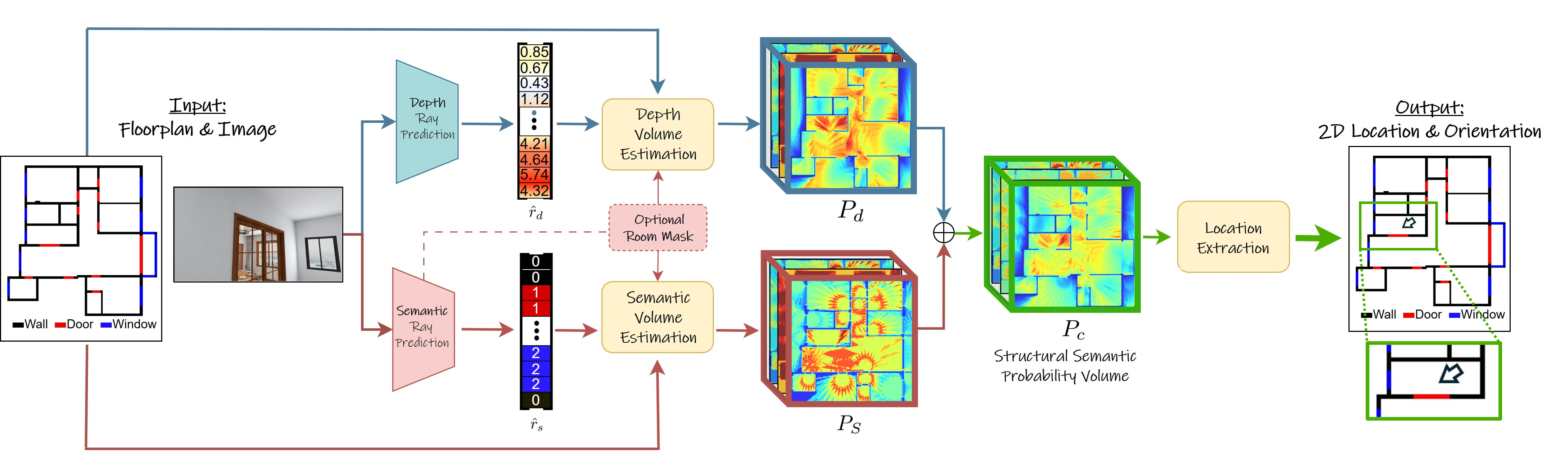}
        \caption{Overview of our pipeline. The input image is processed to generate depth rays, semantic rays, and optionally additional metadata (e.g., room type prediction). We interpolate the ray predictions to a low-resolution representation and generate the depth probability volume \(P_d\) and the semantic probability volume \(P_s\) (optionally masked according to the room type prediction). These probability volumes are then fused to form the structural-semantic probability volume \(P_c\) for efficient coarse localization. Finally, we  refine the candidate poses using high-resolution ray predictions and predict the final 2D camera location and orientation, visualized with an arrow on the right. 
        }
    \label{fig:methods_pipeline}
\end{figure*}


\section{Method}
\label{sec:methods}

In this work, we propose a floorplan localization framework that jointly estimates semantic and depth rays to infer the 2D camera location and orientation relative to a given floorplan. Specifically, we assume that we are provided with an RGB Image \( I \in \mathbb{R}^{h \times w \times 3} \), where \( h \) and \( w \) denote the height and width of the image, respectively, and a 2D floorplan map \(F \in \{0,1,2,...,C\}^{H \times W}\), represented as a matrix of dimensions \(H \times W\), where each element is assigned a semantic label from $C$ unique semantic categories, with zero denoting \emph{empty space}. The semantic categories we consider in our work are \emph{wall}, \emph{window} and \emph{door}, but our framework could easily incorporate additional categories (e.g, staircases, columns). 

Our objective is to predict the camera's 2D location \( (x,y) \) and orientation angle \( \theta \) at which the image \( I \) was captured. That is, given the observation \( O_{I,F} = (I, F) \), our goal is to infer the location parameters \( S_{I,F} = (x,y,\theta) \). To this end, we adopt a probabilistic framework by modeling the posterior distribution \( p(S_{I,F} \mid O_{I,F}) \). We discretize the camera pose space as \( \mathcal{S} = \{S_i\} \) and define a probability volume \( P \in \mathbb{R}^{\hat{H} \times \hat{W} \times O} \) where each element \( P(S_i) \) represents the posterior probability \( p(S_i \mid O_{I,F}) \) for a candidate pose \( S_i \). Here, \(\hat{H}\) and \(\hat{W}\) denote the number of discretized cells in the \(x\) and \(y\) dimensions, respectively, and \(O\) represents the number of orientation bins. The final predicted camera pose is then given by
\[
    \hat{S}_{I,F} = \operatorname*{arg\,max}_{S_i \in \mathcal{S}} \; p(S_i \mid O_{I,F}).
\]

We proceed to provide  background (Section \ref{sec:F3Loc}), prior to introducing our semantic prediction network (Section \ref{sec:semantic_prediction}), which constructs a semantic probability volume. We then describe how it is fused with depth cues to perform floorplan localization (Section \ref{sec:fusion}). Finally, we provide training and implementation details (Section \ref{sec:training}); additional details can be found in the supplementary material. 
An overview of our approach is provided in Figure \ref{fig:methods_pipeline}.

\subsection{Background: F3Loc}
\label{sec:F3Loc}




Our work builds upon F3Loc~\cite{chen2024f3loc}, a recent technique that estimates depth rays for performing floorplan localization given a single image or image sequence. We briefly outline several key components from their work that provide background for our framework. For additional details, we refer readers to their work.

\medskip \noindent \textbf{Depth Rays Prediction.} Given a query image, a depth prediction network estimates per-column depth values that capture the distance from the camera to the nearest wall along specific angles. These values are then linearly interpolated to produce a fixed set of equiangular depth rays $\hat{r}_d \in \mathbb{R}^l$ that represent the floorplan depth, with $l$ denoting the number of predicted rays. 

\medskip \noindent \textbf{Estimating Depth Probability Volume.} For every candidate location \((x,y)\) on the floorplan and each discrete orientation \(\theta\), a corresponding set of reference rays is generated based on the floorplan’s geometry. The predicted depth rays are compared with these reference rays to compute a likelihood score for each grid cell and orientation, resulting in a three-dimensional probability volume \(P_d \in [0,1]^{[\hat{H}, \hat{W}, O]}\).  For instance, given a $10$\,m~$\times$~$7$\,m floorplan discretized at $0.1$\,m with $10^\circ$ increments in orientation yields a probability volume $P_d\in[0,1]^{[100,\,70,\,36]}$.

The final camera 2D location and orientation are determined by selecting the grid cell with the highest likelihood in the probability volume. This process maximizes the posterior probability, thereby estimating the camera's \(x\) and \(y\) coordinates as well as its orientation.

\subsection{Adding a Semantic Prediction Network}
\label{sec:semantic_prediction}
To utilize semantic cues for performing floorplan localization, we propose to add a semantic prediction network that first predicts semantic rays, and then processes these for constructing a semantic probability volume. We provide additional details in what follows, and then present an optional room type prediction component, which can be utilized if room labels are available. 

\begin{figure}[t]
\centering
\resizebox{\linewidth}{!}{%
    \includegraphics[width=\linewidth]{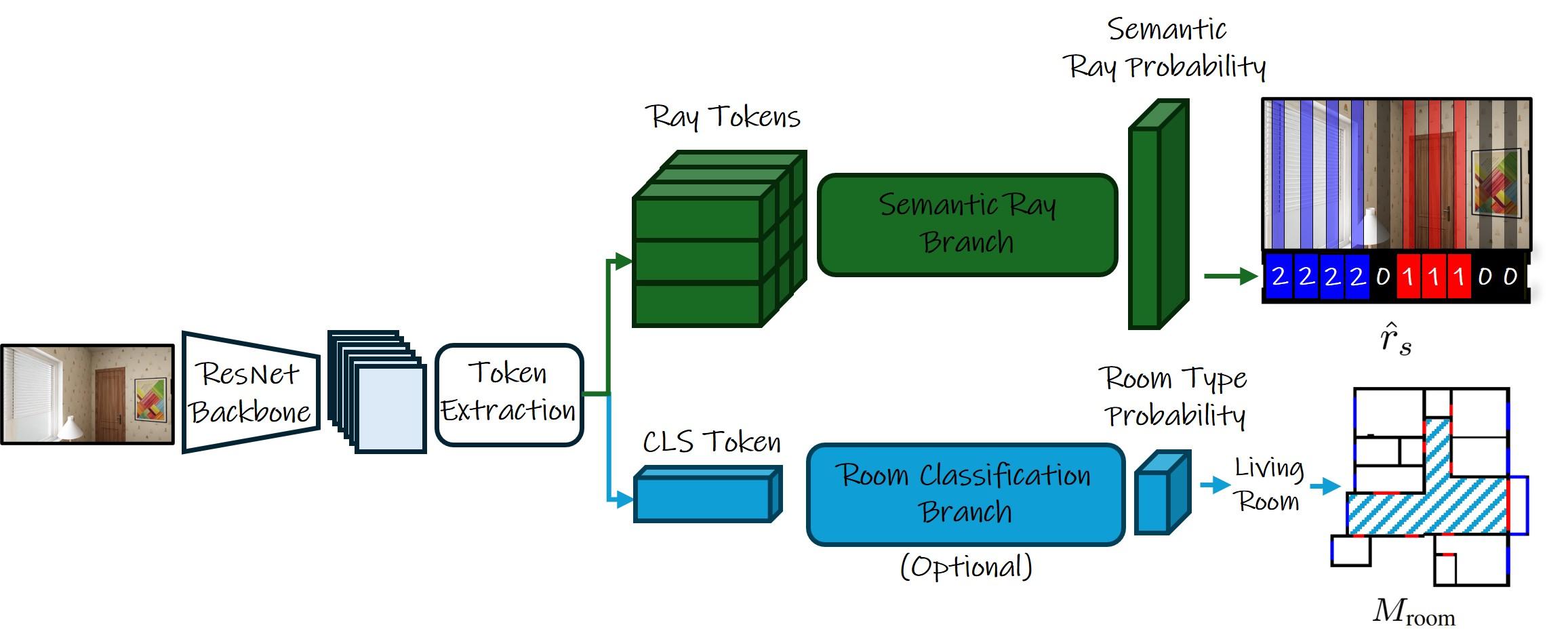}
}
\caption{Overview of our semantic prediction network that predicts a set of semantic rays through the \emph{Semantic Ray Branch} (top) and an optional room type value---\emph{e.g.}, \texttt{Living Room}---through the \emph{Room Classification Branch} (bottom). The room type is used for extracting the mask $M_{\text{room}}$, as visualized on the bottom right. 
}
\label{fig:semantic_network}
\end{figure}

\medskip \noindent \textbf{Semantic Rays Prediction.} 
Unlike the continuous depth values estimated in prior work, the semantic rays should correspond to semantic categories, which are represented as a set of discrete classes. Therefore, we construct a network that produces a semantic ray representation \(\hat{r}_s \in \{1,\ldots,C\}^l\) from the image, where each ray is classified into one of \(C\) semantic categories. We provide an overview of our semantic ray prediction network in Figure \ref{fig:semantic_network}.

As illustrated in the figure, our semantic network architecture leverages a pretrained ResNet50 backbone to extract robust, high-level features from an input RGB image \(I\). After reducing the feature channels using a CNN and projecting them into a lower-dimensional subspace, positional encodings are computed to preserve spatial information. Two sets of learnable tokens are introduced: a set of \(l\) ray tokens responsible for predicting the semantic ray representation \(\hat{r}_s\) and a single (optional) CLS token dedicated for representing \emph{global} room classification information.

A single-head cross attention module integrates these tokens with the flattened spatial features, yielding refined tokens that capture both global context and local details. In the ray branch, the refined ray tokens are first processed by a self-attention block that enables each token to interact with all others, thereby aggregating complementary contextual information. The enriched tokens are then passed through an MLP to produce per-token semantic logits, which after normalization form the final semantic ray vector \(\hat{r}_s\). If room labels are available in the dataset, a similar network processes the CLS token for room type prediction, as we further detail later.

\medskip \noindent \textbf{Estimating Semantic Probability Volume.}
\label{Localization Process}
To obtain the semantic probability map, \(P_s\in [0,1]^{[\hat{H},\hat{W},O]}\), we first need to interpolate the \(l\) predicted semantic rays. Regular linear interpolation---which prior work used for depth estimation---is unsuitable in the context of discrete labeling since interpolating between class labels can produce non-valid or semantically meaningless results. Instead, we propose a voting-based interpolation scheme: We reduce the original equiangular rays to the desired count by applying a majority vote within a small neighborhood. We use a window of three rays, assigning the label that appears most frequently in that window to the center target ray; see the supplementary material for the full algorithm. Next, we compute the score for each set of rays by taking the $L_1$ difference between the predicted semantic labels and the reference labels. The score is then exponentiated and normalized to form the semantic probability volume \(P_s\), which quantifies the likelihood of each candidate pose based on the alignment between the semantic rays and the candidate pose.


\medskip \noindent \textbf{Room Type Prediction.} In addition to predicting semantic rays, our semantic network can optionally also predict the room type, which corresponds to the room from which the input image was taken. This is achieved by processing the CLS token in the semantic network (see Figure~\ref{fig:semantic_network}). If the predicted room probability exceeds a threshold \(T_{\text{room}}\), the predicted room type is then used to construct a mask \(M_{\text{room}}\) from the polygons associated with that room type. For example, if the model predicts a \texttt{Living Room} label with high confidence, the mask \(M_{\text{room}}\) retains only the regions corresponding to the living room by setting all other areas to zero. Let \(\tilde{P}\) denote the original probability volume, then the masked probability volume $P$ is computed as:

\[
    P = M_{\text{room}} \odot \Tilde{P},
\]
where \(\odot\) denotes element-wise multiplication, thereby filtering out all the probabilities which are not in the living room and substantially narrowing down the search space. A detailed analysis of room type distributions and the model's prediction accuracy is provided in the supplementary material.

\subsection{Floorplan Localization via a Structural--Semantic Probability Volume}
\label{sec:fusion}

\begin{figure}[t!]
    \centering
    \resizebox{\linewidth}{!}{%
\centering
\renewcommand{\arraystretch}{1.4}
\begin{tabular}{%
  >{\centering\arraybackslash}m{0.20\textwidth}%
  >{\centering\arraybackslash}m{0.15\textwidth}%
  >{\centering\arraybackslash}m{0.2\textwidth}%
  >{\centering\arraybackslash}m{0.2\textwidth}%
  >{\centering\arraybackslash}m{0.2\textwidth}%
  >{\centering\arraybackslash}m{0.2\textwidth}%
}
\noalign{\vskip 3pt}
\hline
\large\textbf{Image} & \large\textbf{Floorplan} & \multicolumn{4}{c}{\large\textbf{Rays}} \\
\cline{3-6}
 &  & \multicolumn{2}{c}{\large Low Resolution} & \multicolumn{2}{c}{\large High Resolution} \\
\cline{3-6}
 &  & \textcolor[HTML]{FF00FF}{\large\textbf{GT}} & \textcolor[HTML]{FFC000}{\large\textbf{Predicted}} & \textcolor[HTML]{FF00FF}{\large\textbf{GT}} & \textcolor[HTML]{47D45A}{\large\textbf{Predicted}} \\
\hline
\noalign{\vskip 5pt}
\parbox[c]{\linewidth}{%
  \centering
  \includegraphics[width=\linewidth]{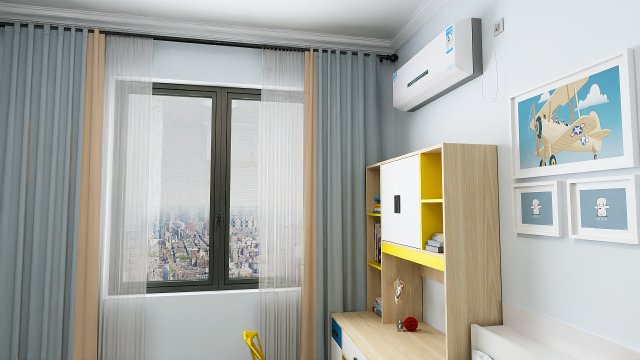}} &
\parbox[c]{\linewidth}{%
  \centering
  \includegraphics[width=\linewidth]{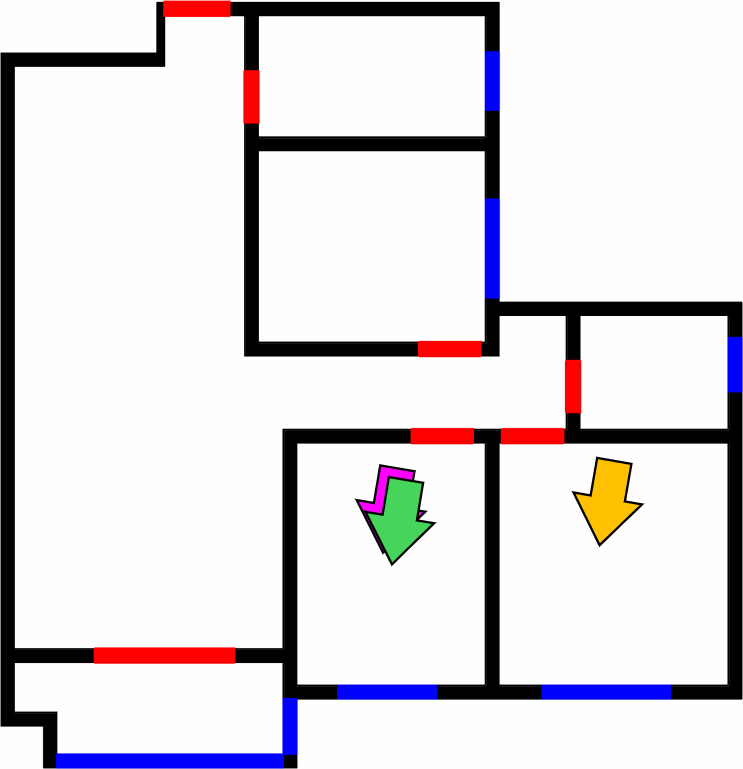}} &
\parbox[c]{\linewidth}{%
  \centering
  \includegraphics[width=\linewidth]{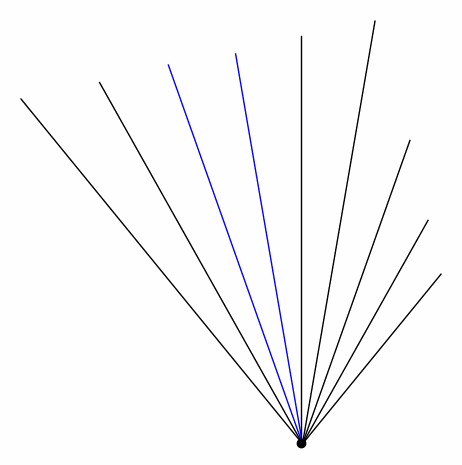}} &
\parbox[c]{\linewidth}{%
  \centering
  \includegraphics[width=\linewidth]{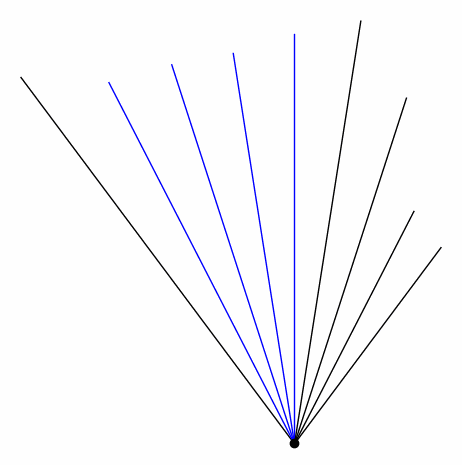}} &
\parbox[c]{\linewidth}{%
  \centering
  \includegraphics[width=\linewidth]{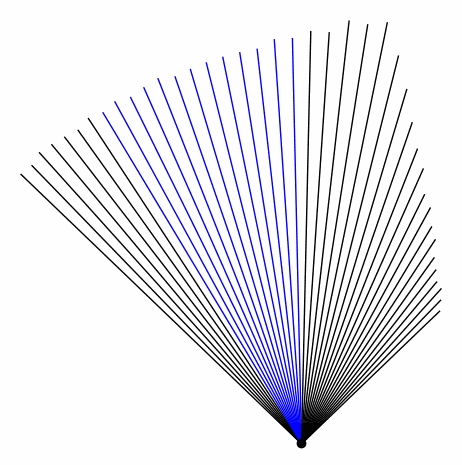}} &
\parbox[c]{\linewidth}{%
  \centering
  \includegraphics[width=\linewidth]{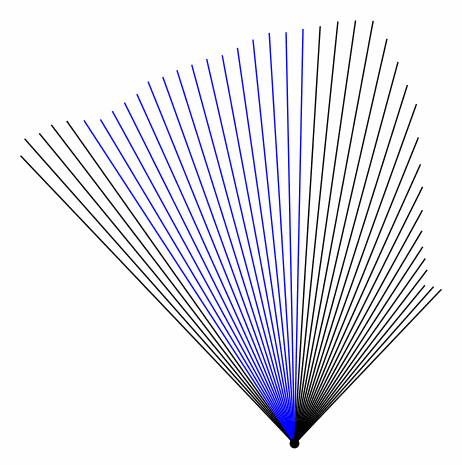}} \\
\noalign{\vskip 5pt}
\parbox[c]{\linewidth}{%
  \centering
  \includegraphics[width=\linewidth]{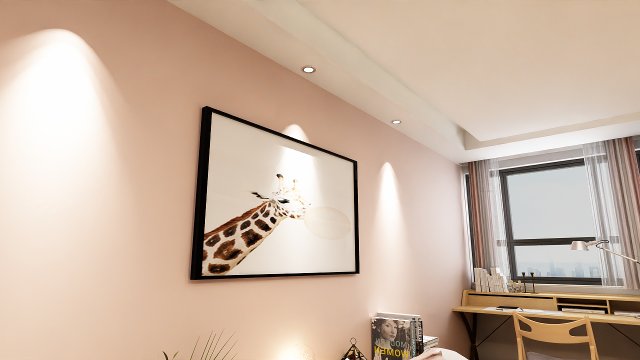}} &
\parbox[c]{\linewidth}{%
  \centering
  \includegraphics[width=\linewidth]{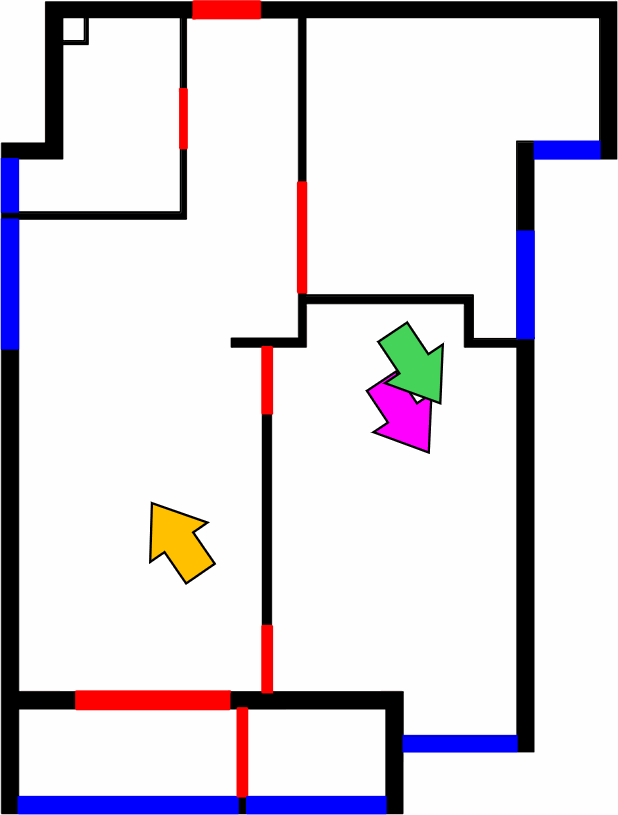}} &
\parbox[c]{\linewidth}{%
  \centering
  \includegraphics[width=\linewidth]{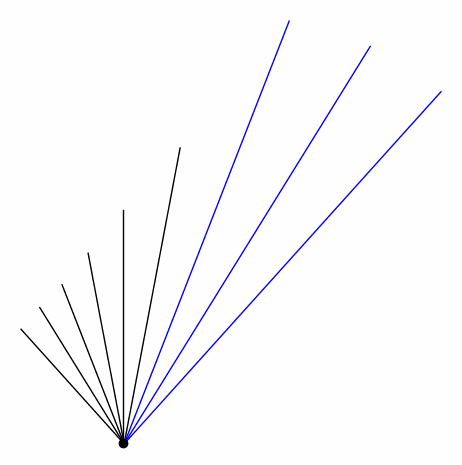}} &
\parbox[c]{\linewidth}{%
  \centering
  \includegraphics[width=\linewidth]{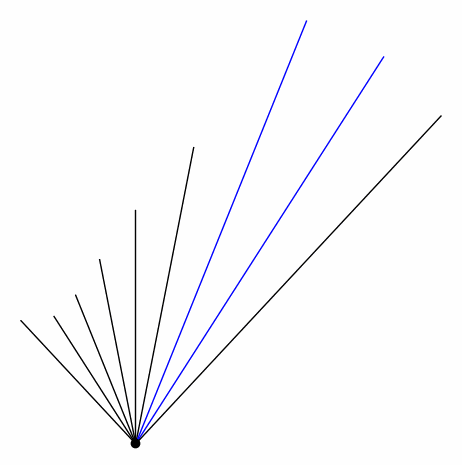}} &
\parbox[c]{\linewidth}{%
  \centering
  \includegraphics[width=\linewidth]{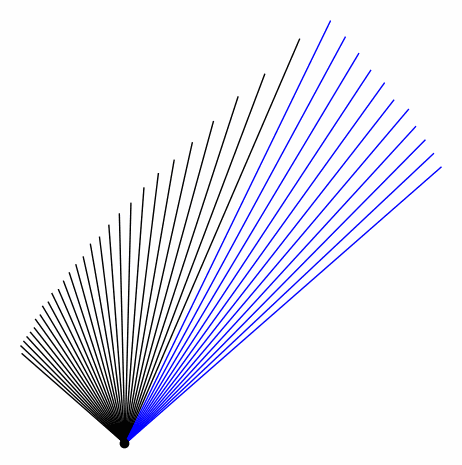}} &
\parbox[c]{\linewidth}{%
  \centering
  \includegraphics[width=\linewidth]{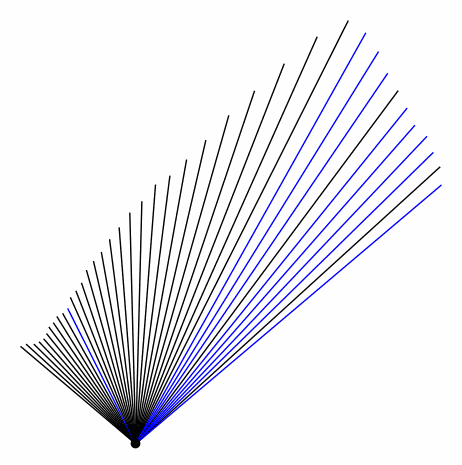}} \\
\noalign{\vskip 3pt}
\hline
\end{tabular}

    }
    \vspace{-8pt}
        \caption{Comparing low-resolution and high-resolution rays with our coarse-to-fine approach. Given an input image (left), we construct a structural--semantic probability volume by comparing low resolution \textcolor[HTML]{FF00FF}{ground-truth} and \textcolor[HTML]{FFC000}{predicted} rays (center). Location extraction from this coarse volume directly often yields significant errors, as illustrated with the yellow arrows. Results are refined only for the Top-k candidate poses by comparing the high resolution \textcolor[HTML]{FF00FF}{ground-truth} and \textcolor[HTML]{47D45A}{predicted} rays (right). This allows for efficiently extracting more accurate predictions, as further detailed in Section \ref{sec:fusion}.   
        }        
    \label{fig:rays_comparison}
\end{figure}

We obtain the final probability map by generating the depth probability map \(P_d\), following the procedure described in Section~\ref{sec:F3Loc}, and the semantic probability map \(P_s\), according to the approach detailed in Section~\ref{Localization Process}. To leverage both semantic and geometric cues, we fuse the two probability maps using a weighted combination:
\[
    P_c = w_s \cdot P_s + w_d \cdot P_d,
    \label{eq:combined_ray}
\]
where \(w_s\) denotes the weight given to the semantic cues, while \(w_d=1 - w_s\) represents the weight of the depth cues. These weights are determined using a held-out validation set, as further detailed in Section \ref{sec:results}. Note that as both \(P_s\) and \(P_d\) are generated from interpolated rays, which we refer to as low-resolution rays (see Figure \ref{fig:rays_comparison}, \textcolor[HTML]{FFC000}{predicted}), and hence the initial probability volume \(P_c\) is also in low resolution.

\medskip \noindent \textbf{Location Extraction}. Our approach follows a coarse-to-fine strategy to achieve precise localization while maintaining computational efficiency. Given \(P_c\), we first extract the Top-$k$ candidate poses from the structural-semantic probability volume based on their scores, ensuring that each candidate is separated by at least \(\delta_{\text{res}}\) in translation. Next, for each candidate, we generate an augmented set of orientations by perturbing its original angle in increments of \(\pm\delta_{\text{ang}}\) up to a maximum deviation of \(\Delta_{\text{max}}\), resulting in the final augmented set:
\[
[0, \pm\delta_{\text{ang}}, \pm2\delta_{\text{ang}}, \ldots, \pm\Delta_{\text{max}}].
\]

At each candidate location, we compute the corresponding high-resolution ground-truth depth and semantic ray representations, yielding \(l\) rays per candidate location. These ground-truth rays are then compared against the original predicted \(l\) rays using a similarity metric (see supplementary material for metric details). For a visual comparison of the high-resolution \textcolor[HTML]{FF00FF}{ground-truth} and \textcolor[HTML]{47D45A}{predicted} rays, please refer to Figure \ref{fig:rays_comparison}. The candidate with the highest similarity score is selected as the final predicted location. 

This coarse-to-fine refinement process effectively leverages the full resolution of the predictions, which were initially interpolated for runtime efficiency, to achieve more precise localization. Note that there is a tradeoff between accuracy and computation time: finer resolutions yield improved precision at the expense of increased computational load. This tradeoff is illustrated in Figure~\ref{fig:rays_comparison}. As can be observed, comparing low resolution rays often yields inaccurate predictions. 
For instance, environments with multiple semantic objects can suffer from loss of critical details (e.g., the left door is omitted in the middle row) in the low-resolution prediction, which can lead to misclassification, further motivating the refinement step. We present a quantitative analysis of this tradeoff in the supplementary material.

\subsection{Training and Implementation Details}
\label{sec:training}
Both networks are trained in an end-to-end manner. For depth prediction, an \(L_1\) loss supervises the predicted depths against ground-truth depth maps. For semantic prediction, a cross-entropy loss is applied to the predicted semantic labels. If room labels \(R\) are available in the dataset, an additional cross-entropy loss is used for the room label, and the network is trained to optimize both objectives jointly. As in prior work, data augmentation techniques—including virtual roll-pitch augmentation—are employed to improve robustness to non-upright camera poses. The networks are optimized using the Adam optimizer with an initial learning rate of \(1 \times 10^{-3}\).
%
We use a floorplan resolution of $0.1$\,m and an angular granularity of $10^\circ$. We set the number of predicted rays to $l=40$ and interpolate these to 7 rays during the coarse stage of localization. For the \emph{Location Extraction} module, we use $\delta_{\text{res}}=0.1$\,m, $\delta_{\text{ang}}=5^\circ$, and $\Delta_{\text{max}}=5^\circ$, and refine our results using Top-$5$ poses. 

\section{Results}
\label{sec:results}
In this section, we present a comprehensive evaluation of our  localization method. We begin by introducing the datasets we use in our experiments, followed by a discussion of the baselines we compare against and the evaluation metrics. The main results are reported in Section \ref{sec:comp}. 
We conduct an ablation study to assess the contributions of various components of our approach in Section \ref{sec:ablations}. We report results for our approach under two conditions: one in which room labels are not utilized (denoted as $\text{Ours}_s$) and another in which room labels are employed to further refine the predictions (denoted as $\text{Ours}_r$). Additional experiments and qualitative results are provided in the supplementary material.

\medskip \noindent \textbf{Datasets.} We conduct experiments on two popular datasets: Structured 3D (S3D)~\cite{Structured3D} and ZInD~\cite{ZInD}. S3D is a synthetic dataset containing realistic 3D renders of 3,296 houses. We use the fully furnished version of S3D, as employed in previous works. ZInD consists of 1,575 unfurnished homes containing only panoramic images. We crop these panoramas to perspective views with an 80\textdegree{} field of view (as was done in S3D), and generate a fixed-size dataset from the resulting images. For both datasets, we follow their official train and test splits.

\medskip \noindent \textbf{Baselines}. We compare our approach against two baselines: F3Loc~\cite{chen2024f3loc} (considering the single-image localization component) and LASER~\cite{min2022laser}. We also report performance using an \textsc{Oracle} ray prediction. This oracle ray prediction simulates the best possible performance achievable by our pipeline using ground truth depth and semantic rays. Note that the Oracle ray prediction baseline does not incorporate room-aware features.
For F3Loc, results on the ZIND dataset are obtained from our experiments by running their publicly available code on the dataset, as the original paper does not include results on this dataset. We also use the publicly available implementation of LASER. Additional details are provided in the supplementary material.

\medskip \noindent \textbf{Metrics.} Following prior work~\cite{chen2024f3loc,min2022laser}, we report recall metrics computed at distance thresholds of 0.1\,m, 0.5\,m, and 1\,m. We also report recall for predictions with an orientation error bounded to less than 30° (at the 1\,m threshold). Recall is defined as the percentage of predictions that fall within these thresholds.

\begin{table}[t]
\centering
\begin{tabular}{lcccc}
\toprule
\multicolumn{5}{c}{\textbf{S3D R@}} \\
\midrule
Method & 0.1m & 0.5m & 1m & 1m 30° \\
\midrule
LASER & 0.7 & 6.4 & 10.4 & 8.7 \\
F3Loc & 1.5 & 14.6 &  22.4 &  21.3 \\
$\text{Ours}_s$ & 5.42 &  41.87 & 53.52 & 52.61   \\
$\text{Ours}_r$ & \textbf{5.70} & \textbf{45.53} & \textbf{58.78} & \textbf{57.49} \\
\hdashline[0.5pt/1pt]
\textsc{Oracle} & 61.00 & 93.84 & 94.87 & 94.57 \\
\midrule
\multicolumn{5}{c}{\textbf{ZInD R@}} \\
\midrule
Method & 0.1m & 0.5m & 1m & 1m 30° \\
\midrule
LASER & 1.38 & 11.06 & 17.55 & 13.64 \\
F3Loc & 0.67 & 7.90 & 15.07 & 11.46 \\
$\text{Ours}_s$  &2.98 & 24.00 & 33.96 & 29.30 \\
$\text{Ours}_r$ & \textbf{3.31} & \textbf{26.60} & \textbf{38.01} & \textbf{31.86} \\
\hdashline[0.5pt/1pt]
\textsc{Oracle } & 26.42 & 60.85 & 67.69 & 65.13 \\
\bottomrule
\end{tabular}
\vspace{-8pt}
\caption{Recall performance on the S3D and ZInD datasets. The table reports recall at thresholds of 0.1\,m, 0.5\,m, 1\,m, and 1\,m with a 30° orientation tolerance for LASER, F3Loc, our approach without room labels ($\text{Ours}_s$), our approach with room labels ($\text{Ours}_r$), and an Oracle ray prediction.}
\label{tab:comparison_s3d_and_zind}
\end{table}
\subsection{Quantitative Evaluation} \label{sec:comp}



Results are reported in Table~\ref{tab:comparison_s3d_and_zind}. On both S3D and ZinD, our method more than doubles F3Loc's and LASER performance across all thresholds. Notably, considering S3D over the R@1m30° metric---which reflects the quality of matches between the predicted and actual camera views---in comparison to F3Loc, our method improves by more than three times. Room type predictions yield improvements of $9.2\%$ in R@1m30° on the S3D dataset and $8.7\%$ on the ZInD dataset. 
We include an additional experiment in the supplementary material that evaluates F3Loc with our refinement module, demonstrating that both the semantic rays and our refinement module provide significant performance gains. 

We observe that significant performance improvements are also achieved for a very fine-grained recall metric of 0.1\,m, boosting performance from 1.5\% (F3Loc) to 5.7\%  with our approach on S3D. We attribute this to our coarse-to-fine strategy, which effectively refines the coarse location estimates into precise predictions, as further validated in our ablation study.

\begin{figure}[t!]
\centering
\resizebox{\linewidth}{!}{%
\newcolumntype{C}[1]{>{\centering\arraybackslash}m{#1}}

\begin{tabular}{C{0.25\textwidth} C{0.15\textwidth} C{0.15\textwidth} C{0.15\textwidth} C{0.15\textwidth}}
\hline
\large\textbf{Input Image} & \large\textbf{Floorplan} & \large\textbf{Ours} & \large\textbf{F3loc} & \large\textbf{LASER} \\
\hline\hline
\includegraphics[width=0.25\textwidth]{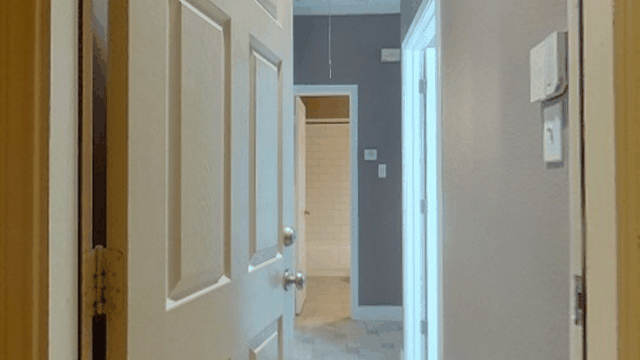} &
\includegraphics[width=0.15\textwidth]{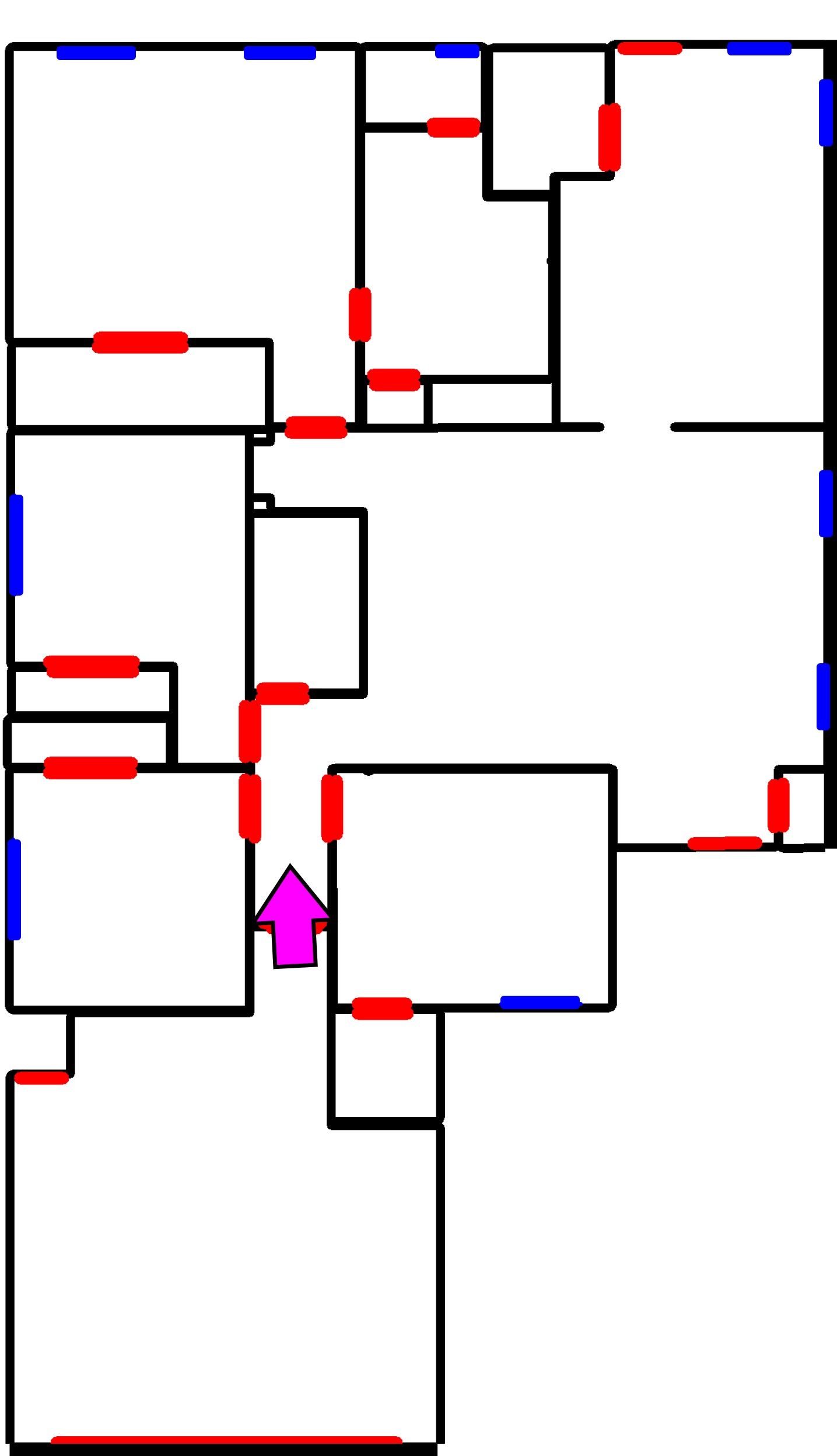} &
\includegraphics[width=0.15\textwidth]{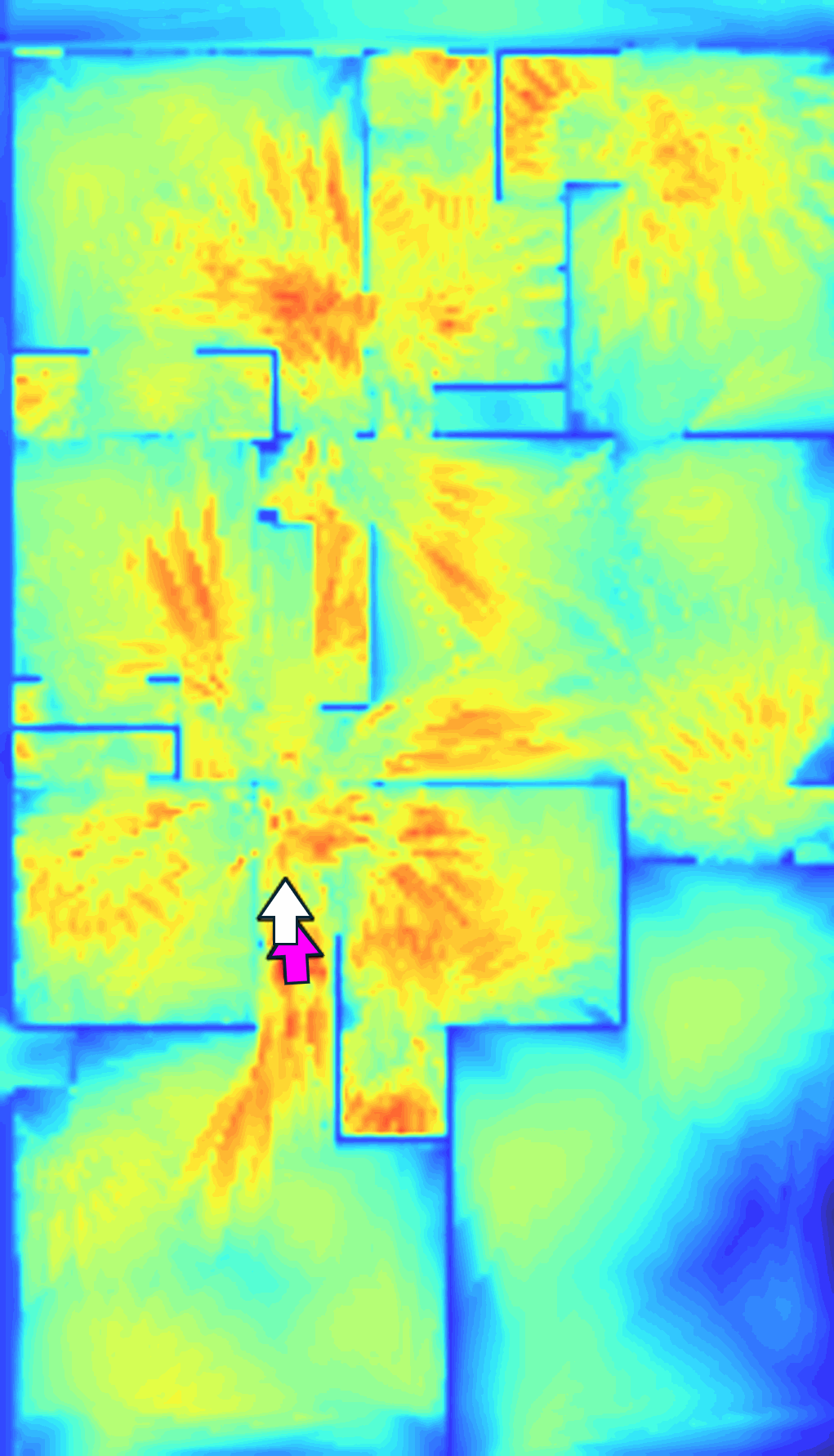} &
\includegraphics[width=0.15\textwidth]{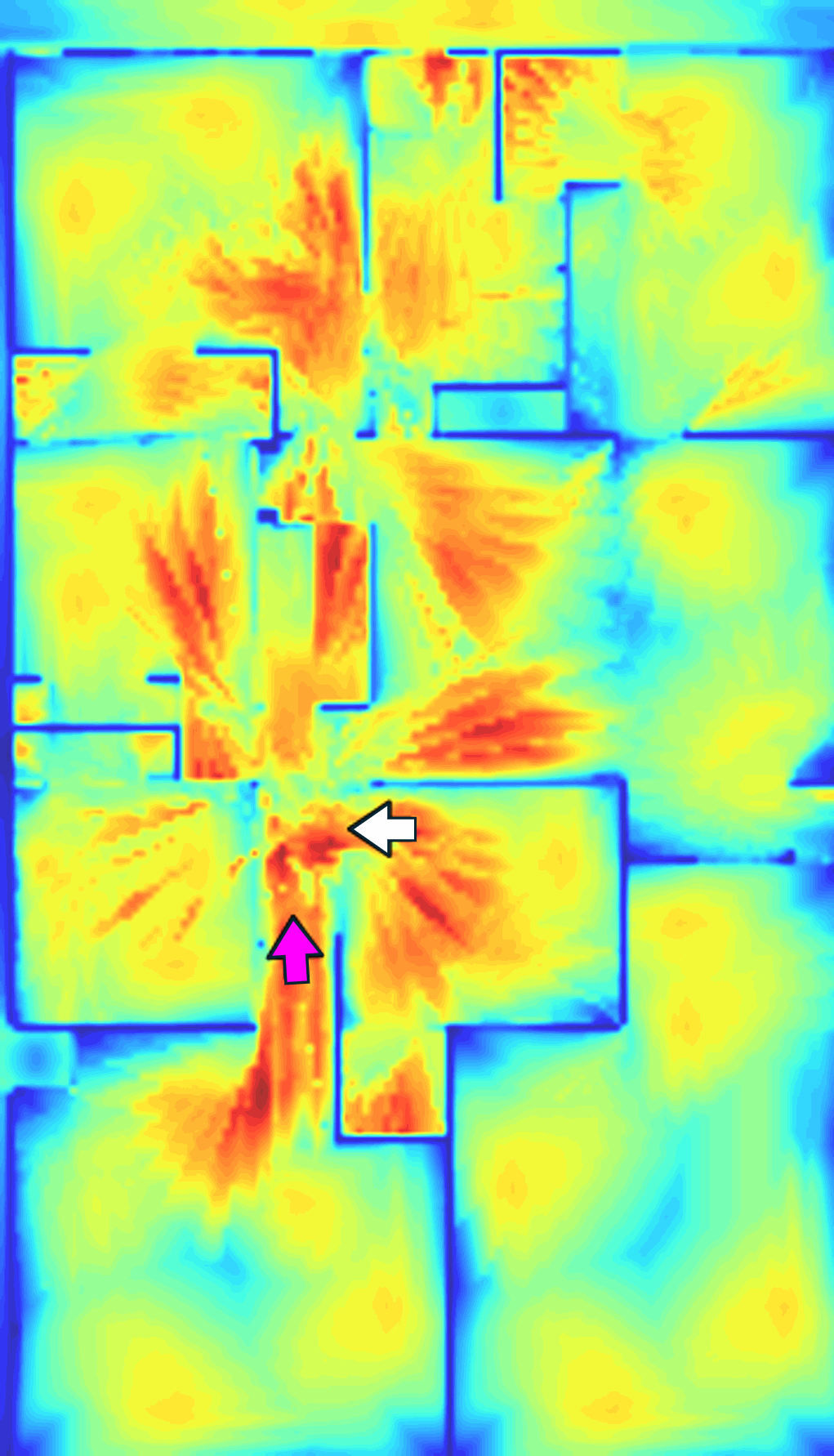} &
\includegraphics[width=0.15\textwidth]{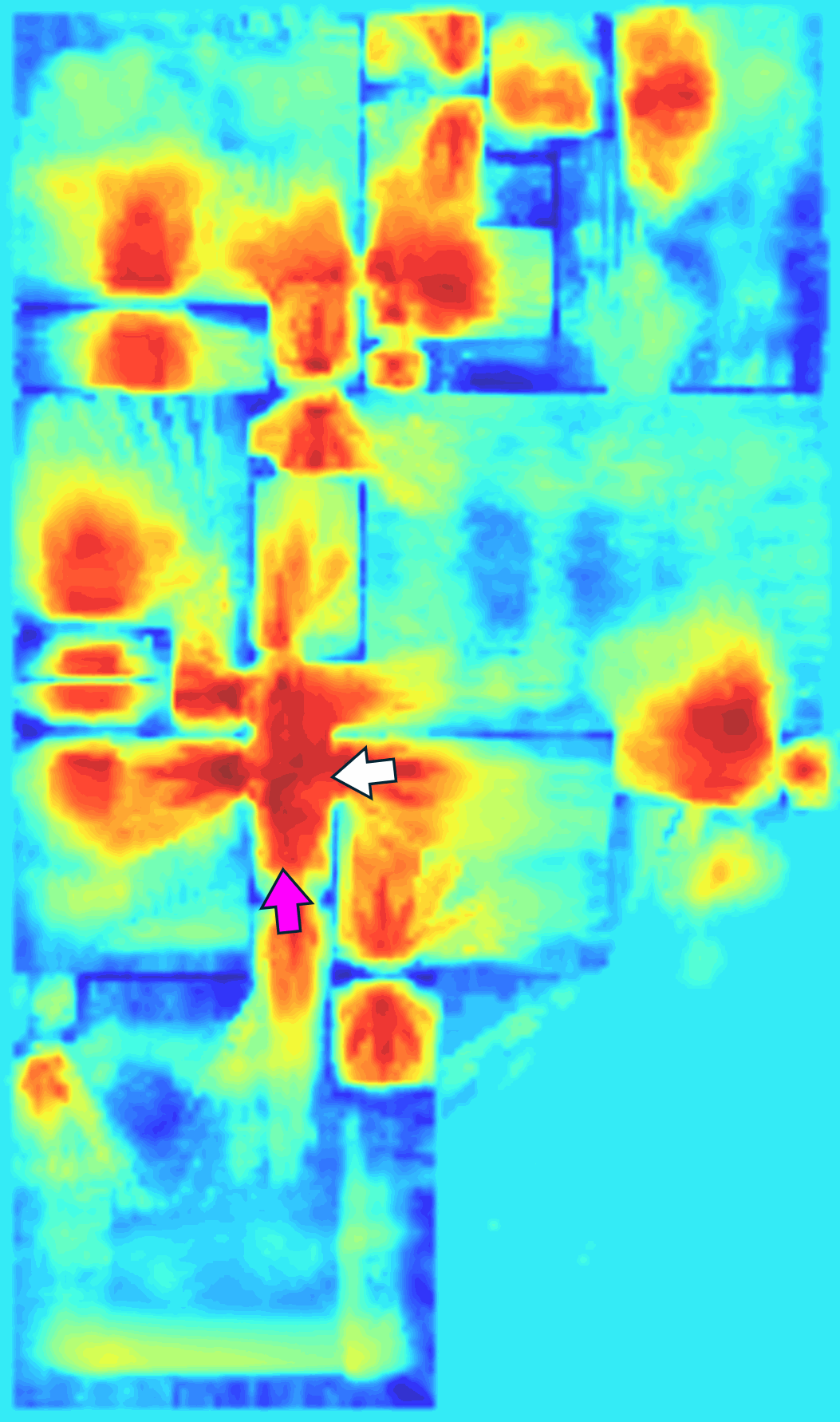} \\
\multicolumn{2}{c}{} & $(0.43m, 17^\circ)$ & $(3.32m, 97^\circ)$ & $(2.57m, 92^\circ)$ \\

\includegraphics[width=0.25\textwidth]{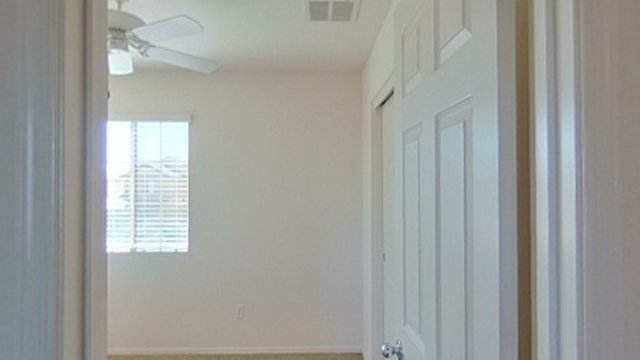} &
\includegraphics[width=0.15\textwidth]{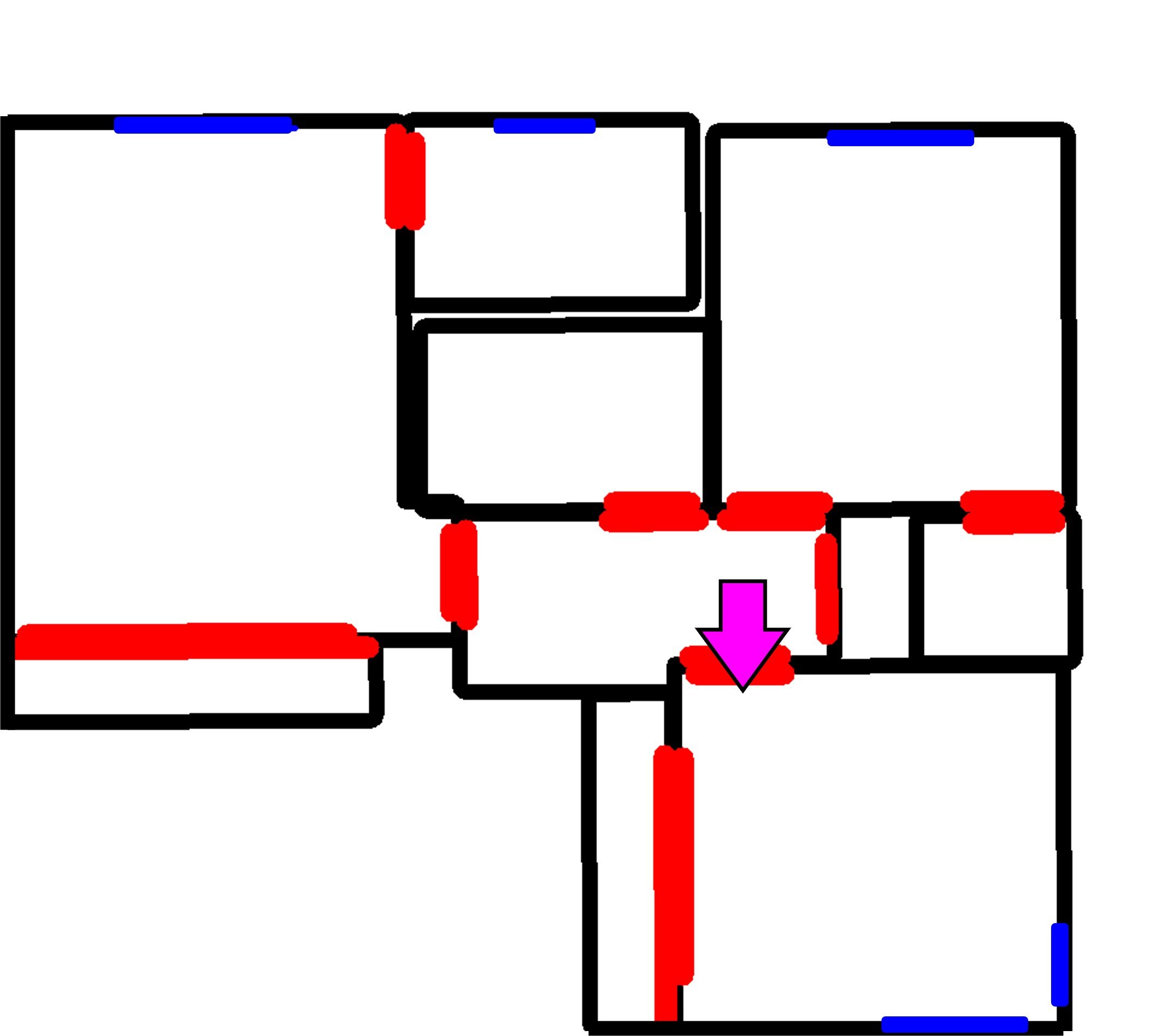} &
\includegraphics[width=0.15\textwidth]{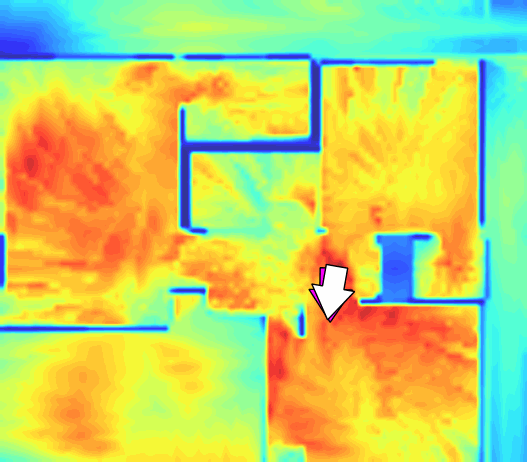} &
\includegraphics[width=0.15\textwidth]{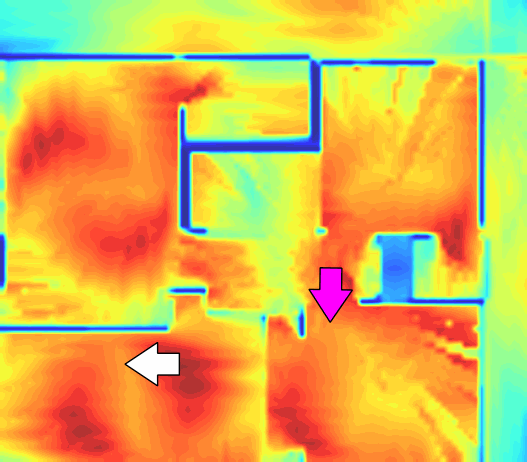} &
\includegraphics[width=0.15\textwidth]{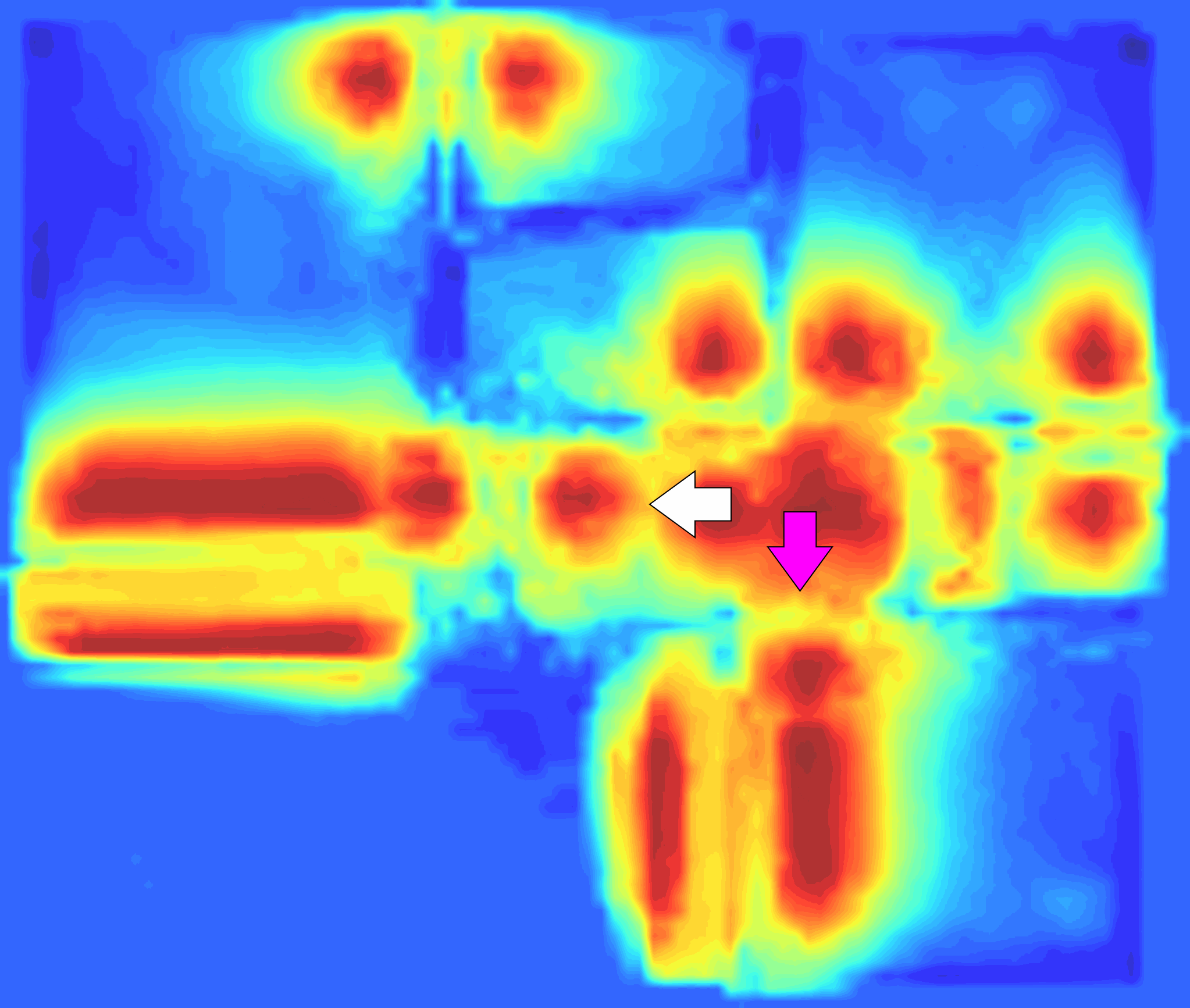} \\
\multicolumn{2}{c}{} & $(0.17m, 07^\circ)$ & $(3.74m, 90^\circ)$ & $(0.70m, 89^\circ)$ \\

\includegraphics[width=0.25\textwidth]{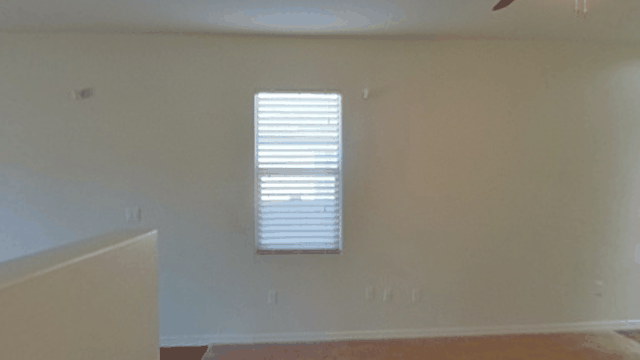} &
\includegraphics[width=0.15\textwidth]{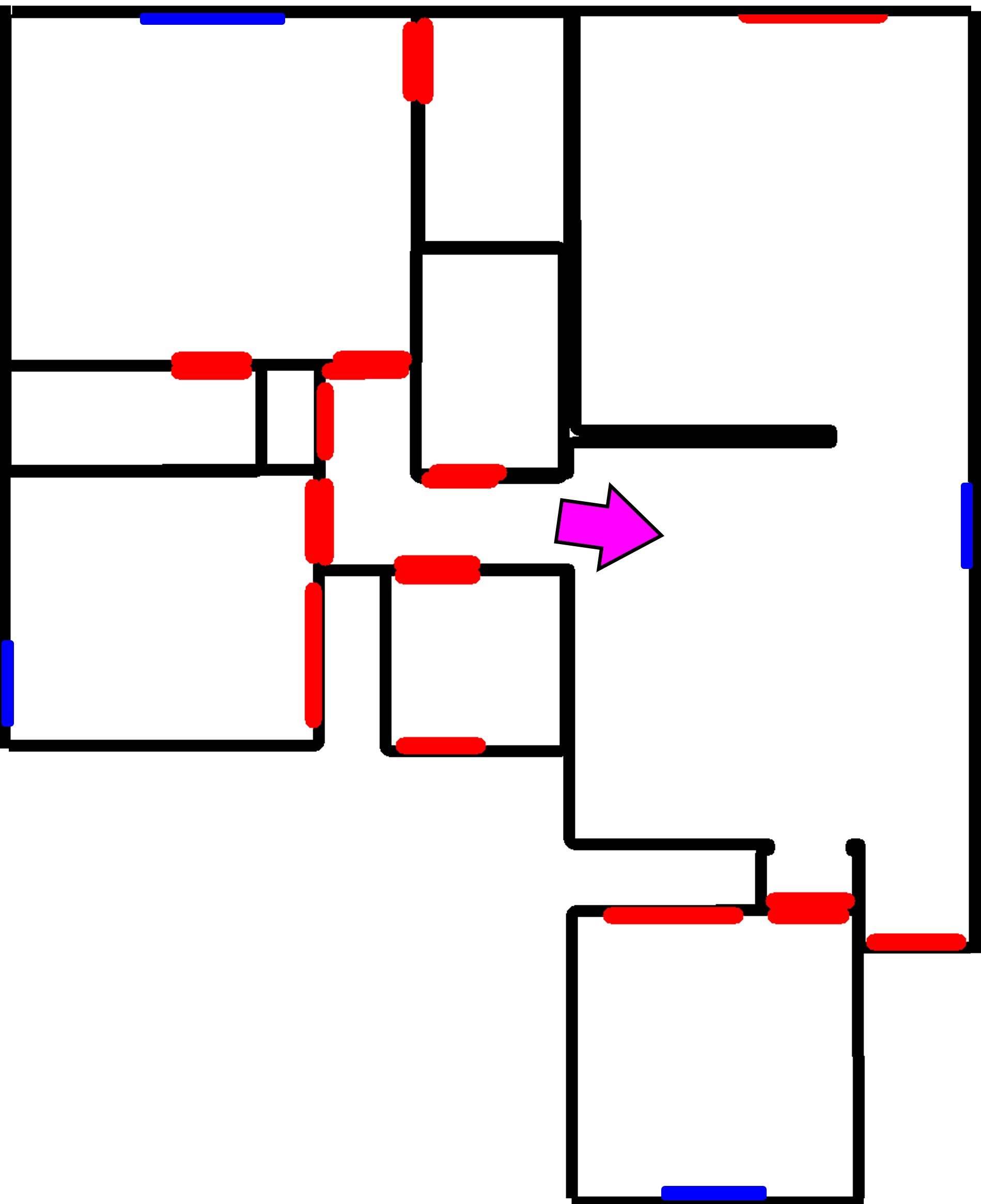} &
\includegraphics[width=0.15\textwidth]{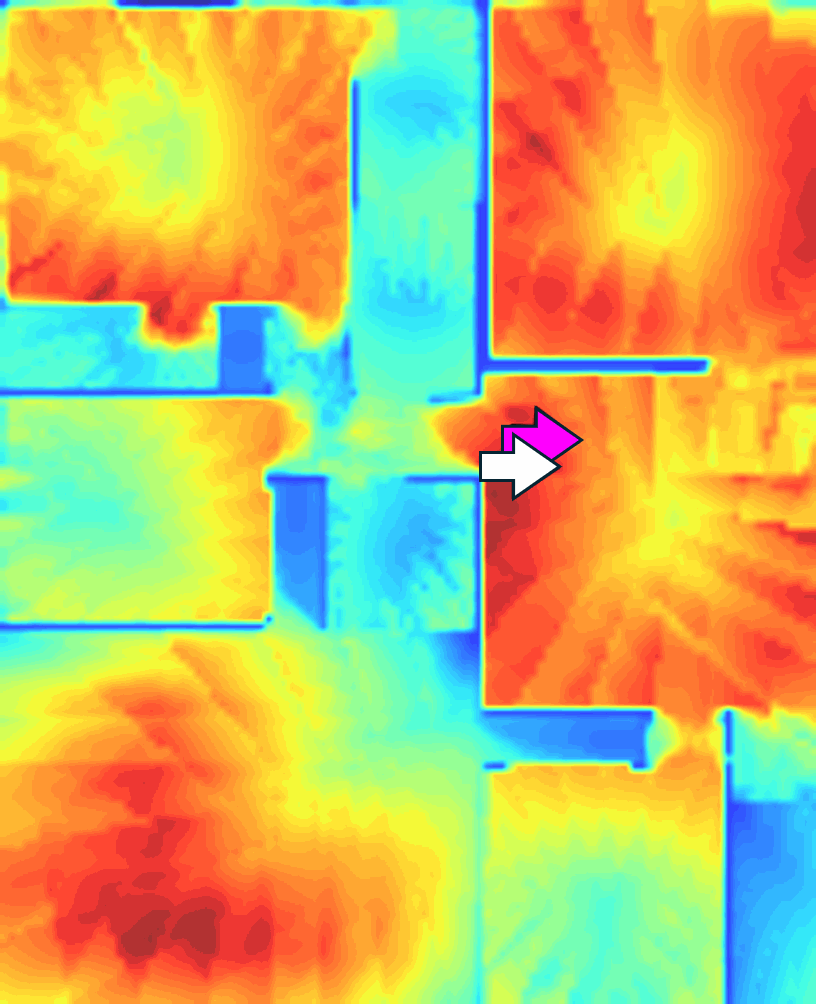} &
\includegraphics[width=0.15\textwidth]{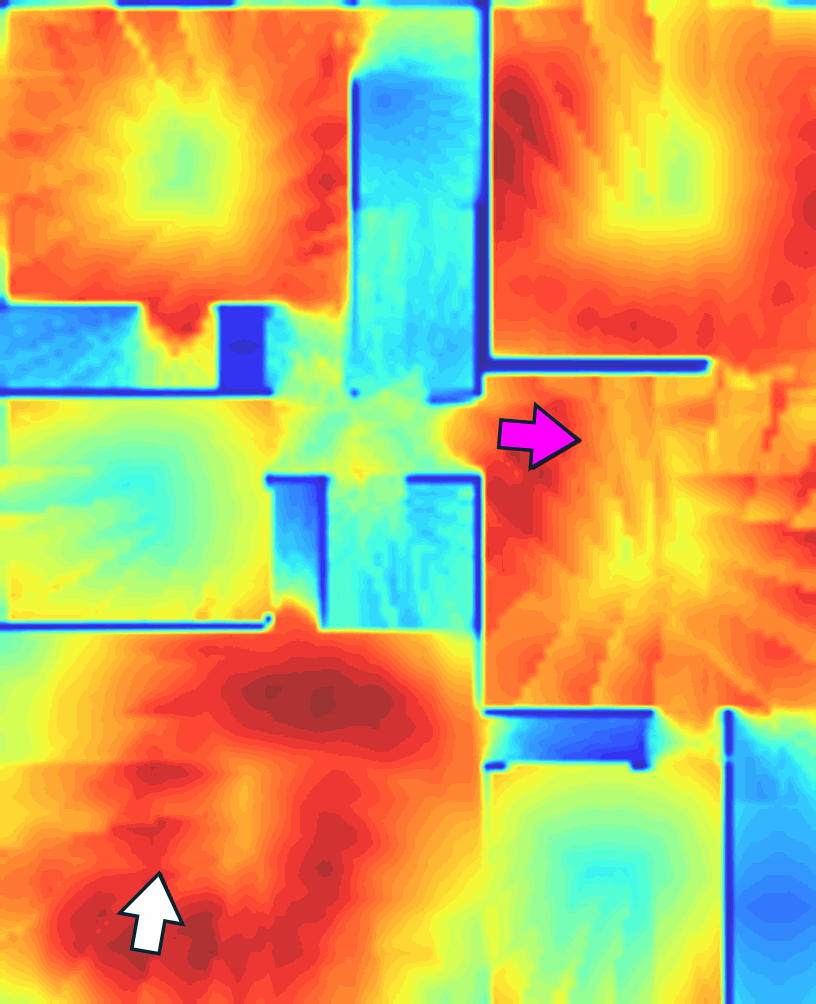} &
\includegraphics[width=0.15\textwidth]{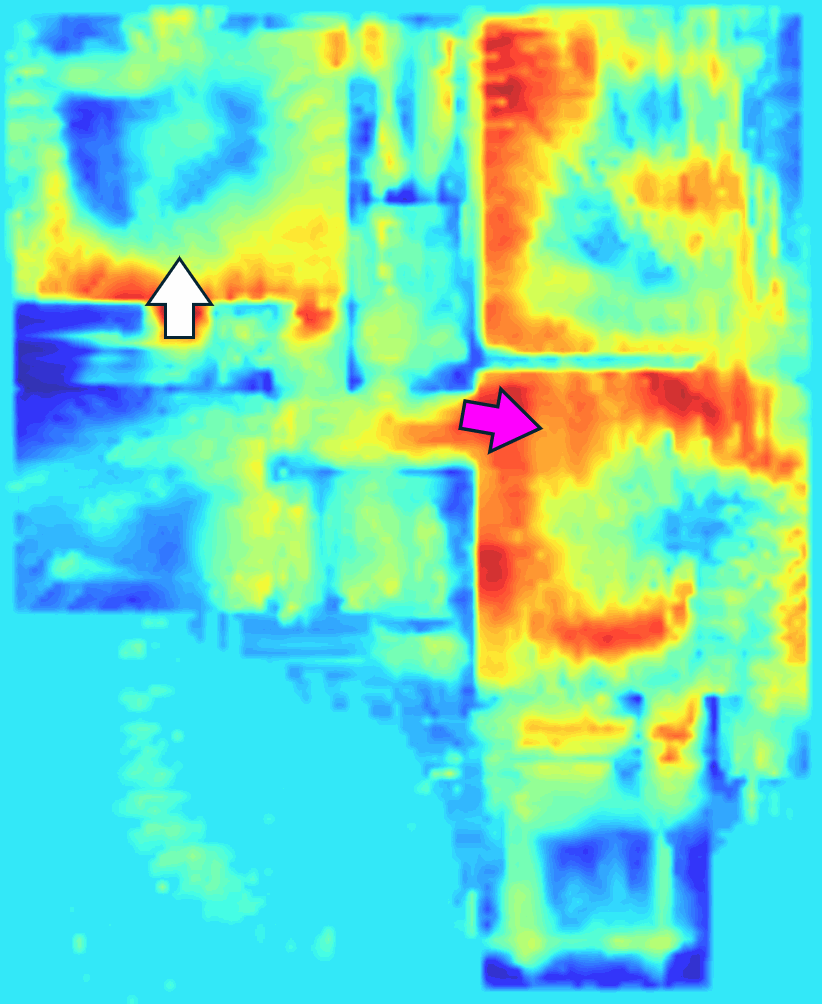} \\
\multicolumn{2}{c}{} & $(0.47m, 3^\circ)$ & $(7.99m, 83^\circ)$ & $(4.81m, 95^\circ)$ \\
\hline\hline
\\
\end{tabular}

}
\vspace{-14pt}
\caption{Qualitative comparison of our method with F3Loc and LASER. Warmer colors correspond to regions with higher probabilities. Below each map we report the localization error in meters and degrees. We use arrows to visualize the ground truth location (\textcolor{magenta}{magenta}) and the predicted location (white).}
\label{fig:compare_results_with_other_methods}
\end{figure}

Figure~\ref{fig:compare_results_with_other_methods} presents qualitative examples that illustrate how the integration of floorplan semantics with precise depth cues enables our pipeline to effectively resolve localization ambiguities. For instance, in the third row, we can see that F3Loc, which does not use semantics, interprets this image as a blank wall, while LASER misinterprets the window size and predicts an incorrect location.

\subsection{Ablation Study}
\label{sec:ablations}

We demonstrate the effect of incorporating each component of our method on overall localization performance. Specifically, we conduct the following ablations: (1) \emph{Base}, which corresponds to computing the structural-semantic probability volume and selecting the maximum probability without any additional refinement. (2) \emph{Removing semantic interpolation} (denoted as --Interpolation), where we replace our majority voting interpolation with a simple linear interpolation followed by rounding. (3) \emph{Adding room predictions} (denoted as +Room), where we assess the effect of integrating room type predictions into our localization pipeline. (4) \emph{Adding refinement} (denoted as +Refine), which assesses our coarse-to-fine approach, which refines our localization extraction via the Top-K candidate poses. (5) \emph{Adding room predictions and refinement} (denoted as +Room\&Refine), which combines both components. 

From the results in Table~\ref{tab:ablation_methods},
we can see that on the 1m 30° metric, the addition of our refinement module improves performance by 8.6\% relative to the baseline. This indicates that a substantial amount of information is lost during the initial interpolation process if not properly addressed, thereby strongly motivating the use of coarse-to-fine strategies in our Location Extraction module. We also observe a significant improvement of 11.5\% from the room prediction component, which is discussed in further detail in the supplementary material and visualized in Figure~\ref{fig:ablation_fig_room_aware}. Finally, our majority voting interpolation approach contributes an additional gain of 0.9\% compared to a simple interpolation strategy. When combining all these improvements, our method achieves an overall enhancement of 18.6\% on the 1m 30° metric relative to our base model.

\begin{table}[t]
    \centering
    \begin{tabular}{lcccc}
        \hline
        \textbf{Method} & \textbf{0.1m} & \textbf{0.5m} & \textbf{1m} & \textbf{1m 30°} \\
        \hline
        Base      &    4.65 & 38.35 & 49.40 & 48.44 \\
        -- Interpolation & 4.73 & 38.44 & 48.91 & 47.99\\
        + Room& 5.12 & 42.92 & 55.57 & 54.04 \\   
        + Refine & 5.42 & 41.87 & 53.52 & 52.61 \\
        + Room\&Refine & \textbf{5.70} & \textbf{45.53} & \textbf{58.78} & \textbf{57.49} \\
        \hline
    \end{tabular}
    \vspace{-8pt}
    \caption{Ablation study, evaluating the effect of incorporating the different components in our pipeline on the S3D dataset. 
    }
    \label{tab:ablation_methods}
\end{table}

\begin{figure}[t!]
\centering
\resizebox{\linewidth}{!}{%
\centering
\normalsize
\begin{tabular}{%
  >{\centering\arraybackslash}m{0.25\textwidth}%
  >{\centering\arraybackslash}m{0.15\textwidth}%
  >{\centering\arraybackslash}m{0.15\textwidth}%
  >{\centering\arraybackslash}m{0.15\textwidth}%
  >{\centering\arraybackslash}m{0.15\textwidth}%
}
\hline
\large\textbf{Input Image} &\large \textbf{Floorplan} & \large\textbf{Base} & \large\textbf{+Room} & \large\textbf{+Refinement} \\
\hline\hline
\includegraphics[width=\linewidth]{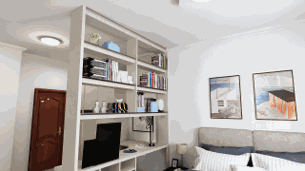} &
\includegraphics[width=\linewidth]{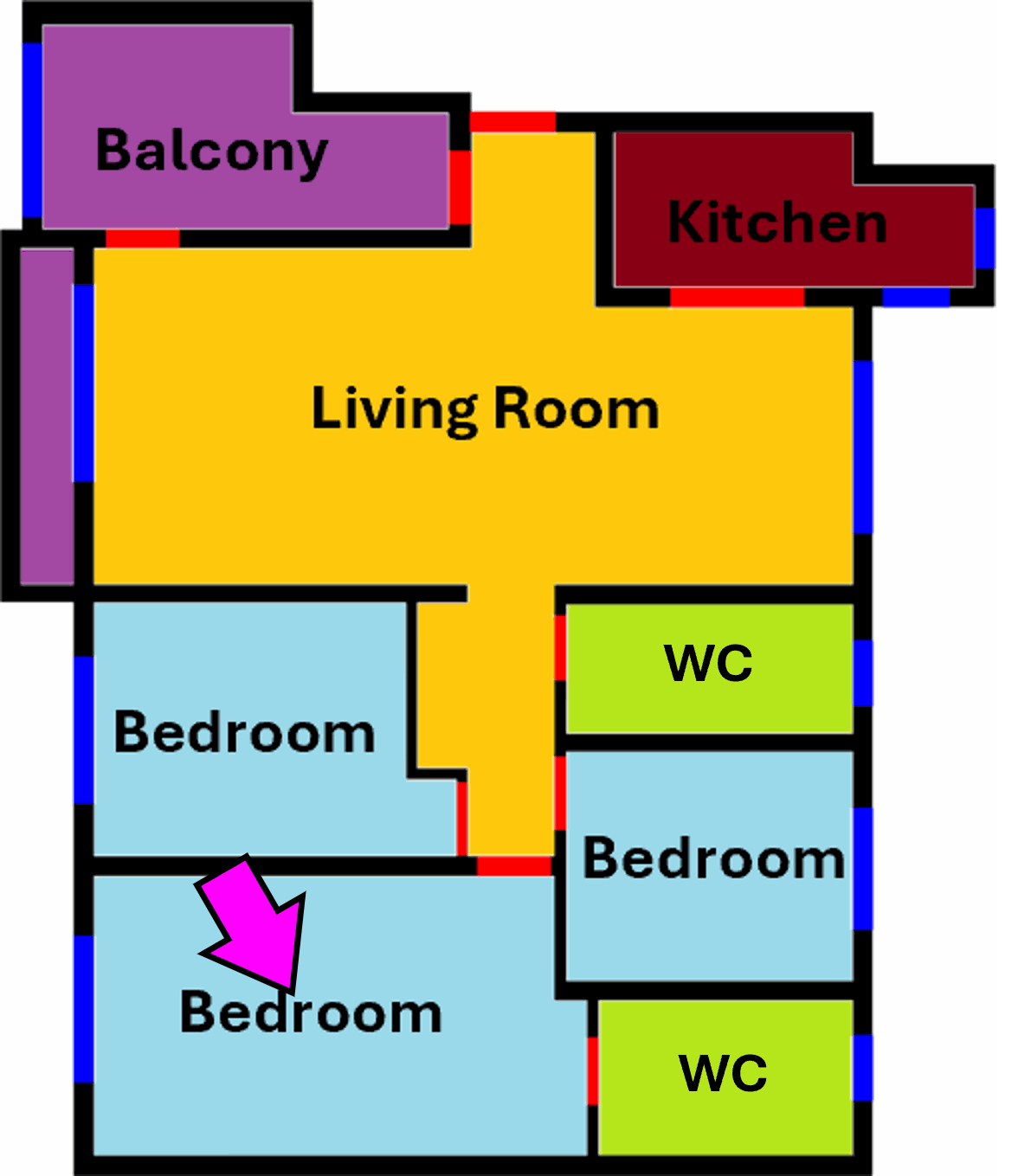} &
\includegraphics[width=\linewidth]{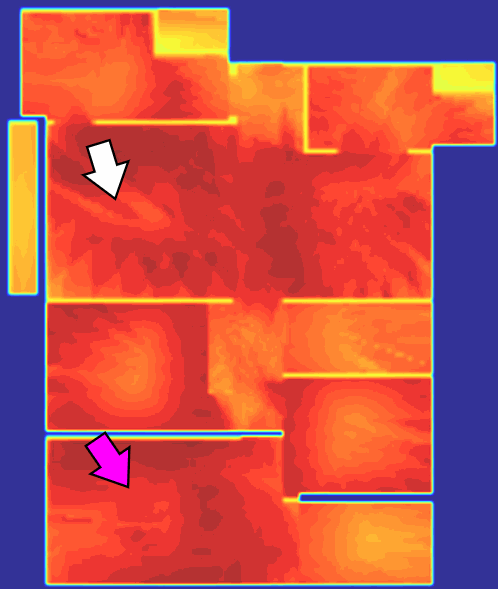} &
\includegraphics[width=\linewidth]{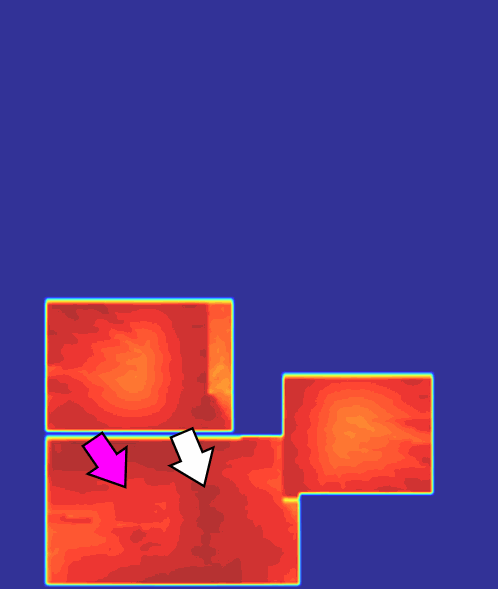} &
\includegraphics[width=\linewidth]{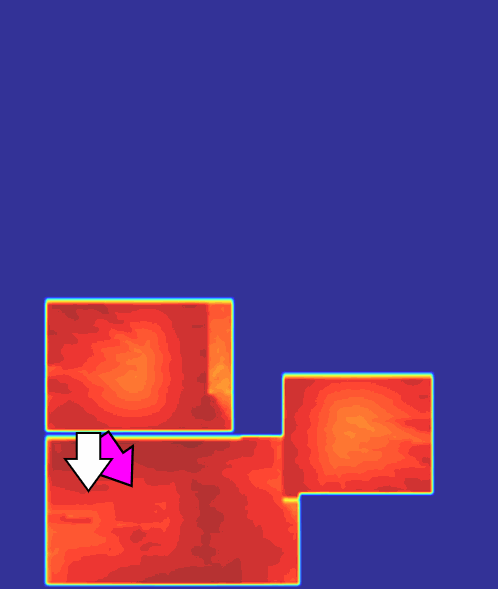} \\
Bedroom &
&
$(6.84m, 15^\circ)$ &
$(2.2m, 16^\circ)$ &
$(0.16m, 25^\circ)$ \\
\includegraphics[width=\linewidth]{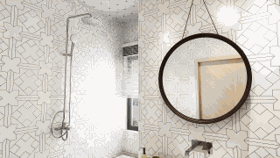} &
\includegraphics[width=\linewidth]{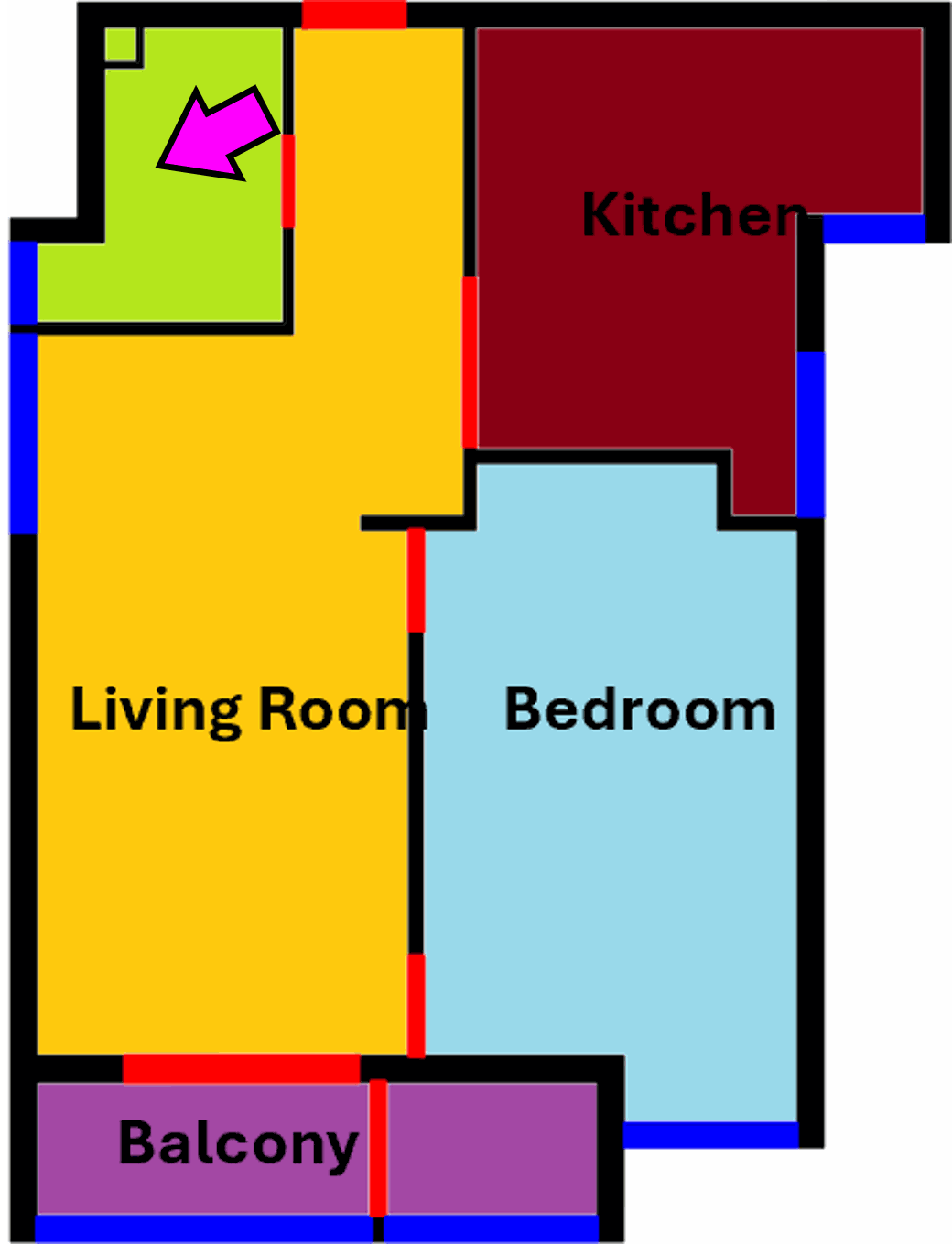} &
\includegraphics[width=\linewidth]{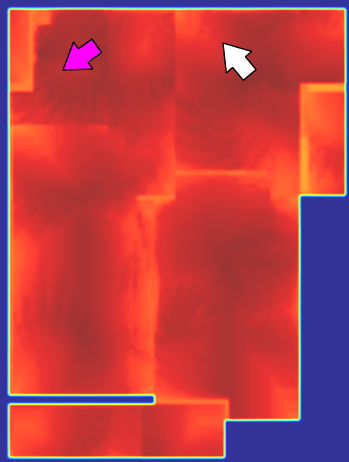} &
\includegraphics[width=\linewidth]{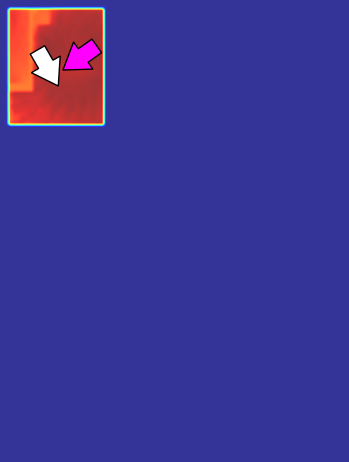} &
\includegraphics[width=\linewidth]{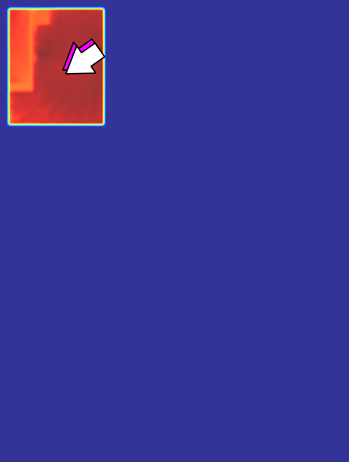} \\
W/C &
&
$(3.74m, 85^\circ)$ &
$(1.22m, 84.4^\circ)$ &
$(0.12m, 1^\circ)$ \\
\hline\hline
\end{tabular}

}
\vspace{-8pt}
\caption{Qualitative comparison of using room predictions and our coarse-to-fine refinement approach. Below each map we report the localization error in meters and degrees. Warmer colors correspond to regions with higher probabilities. Overlaid on the estimated probabilities, we show the ground truth location (\textcolor{magenta}{magenta}) and the predicted location.}
\label{fig:ablation_fig_room_aware}
\end{figure}

In Figure~\ref{fig:ablation_fig_room_aware}, we illustrate the impact of incorporating room type predictions alongside the location extraction module. The figure clearly demonstrates how these components refine the probability volume by narrowing down the candidate poses, which in turn improves overall localization accuracy.

\begin{figure}[t]
  \centering
  \includegraphics[width=\linewidth]{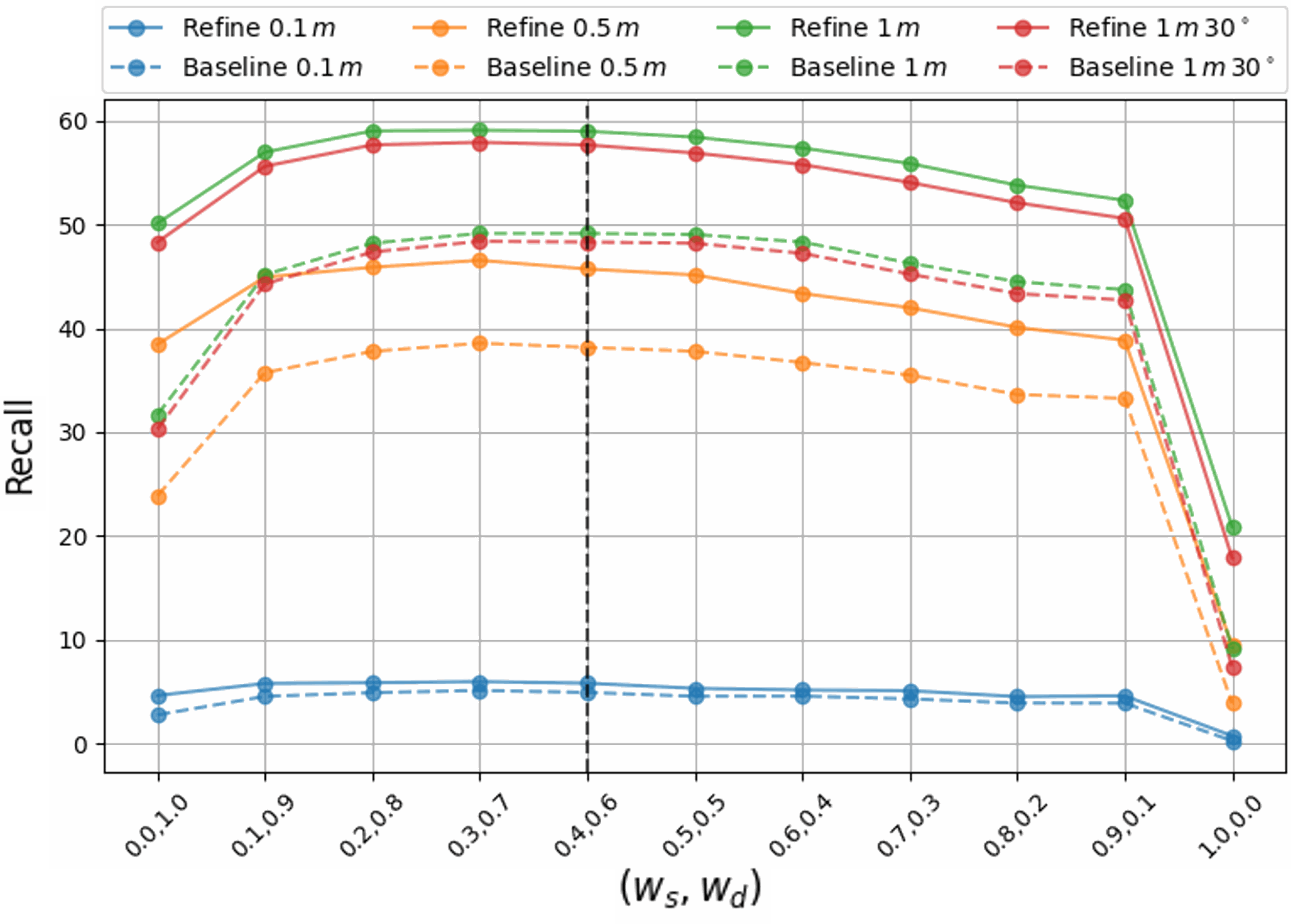}
\caption{Recall vs. weight combinations on the S3D test set. The plot shows the recall metrics for four different thresholds: \textcolor[HTML]{1f77b4}{$0.1\,m$}, \textcolor[HTML]{ff7f0e}{$0.5\,m$},  \textcolor[HTML]{2ca02c}{$1\,m$}, and  \textcolor[HTML]{d62728}{$1\,m\,30^\circ$}. The x-axis displays the weight combinations in the order $(w_s, w_d)$. A vertical dashed line at $w_s=0.4$, $w_d=0.6$ highlights the weight combination selected using the validation set.}
  \label{fig:recall_plot}
\end{figure}
\begin{figure}[t!]
    \centering
    \resizebox{\linewidth}{!}{%
\centering
\renewcommand{\arraystretch}{1.2}
\begin{tabular}{%
  >{\centering\arraybackslash}m{0.25\textwidth}%
  >{\centering\arraybackslash}m{0.15\textwidth}%
  >{\centering\arraybackslash}m{0.15\textwidth}%
  >{\centering\arraybackslash}m{0.15\textwidth}%
  >{\centering\arraybackslash}m{0.15\textwidth}%
}
\hline
\large\textbf{Input Image} & \large\textbf{Floorplan} & \multicolumn{3}{c}{\large\textbf{ Probability Volumes}} \\
\cline{3-5}
 & & \large\textbf{Depth} & \large\textbf{Semantic} & \large\textbf{Structural Semantic} \\
\hline\hline
\parbox[c]{\linewidth}{%
  \centering
  \includegraphics[width=\linewidth]{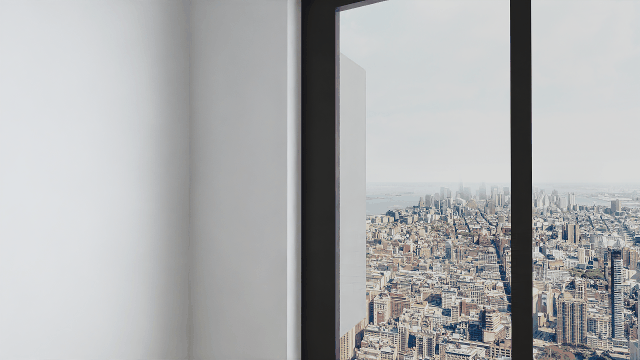}\\[0.5ex]} &
\parbox[c]{\linewidth}{%
  \centering
  \includegraphics[width=\linewidth]{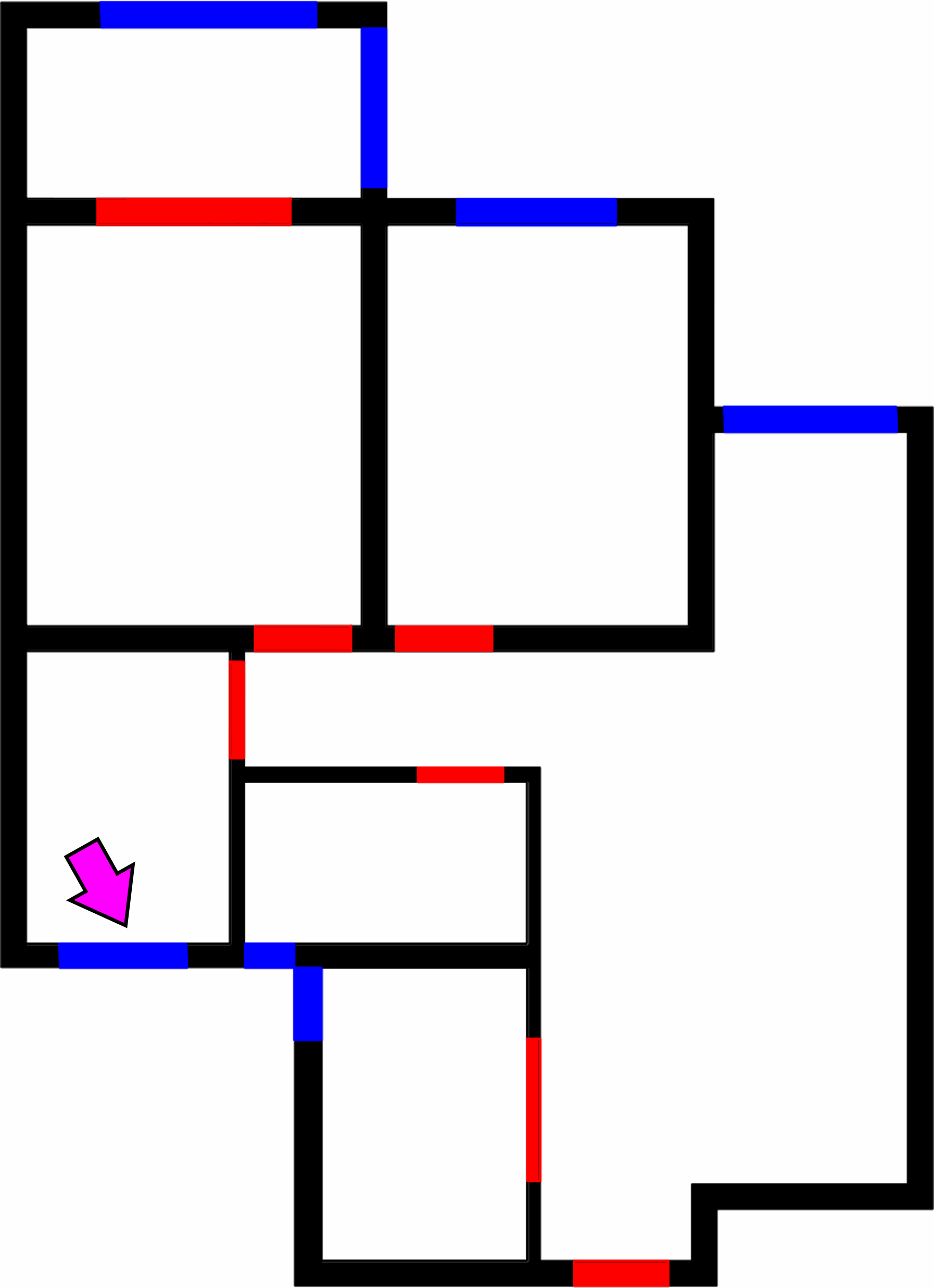}\\[0.5ex]
  \vphantom{$(3.67m, 178^\circ)$}} &
\parbox[c]{\linewidth}{%
  \centering
  \includegraphics[width=\linewidth]{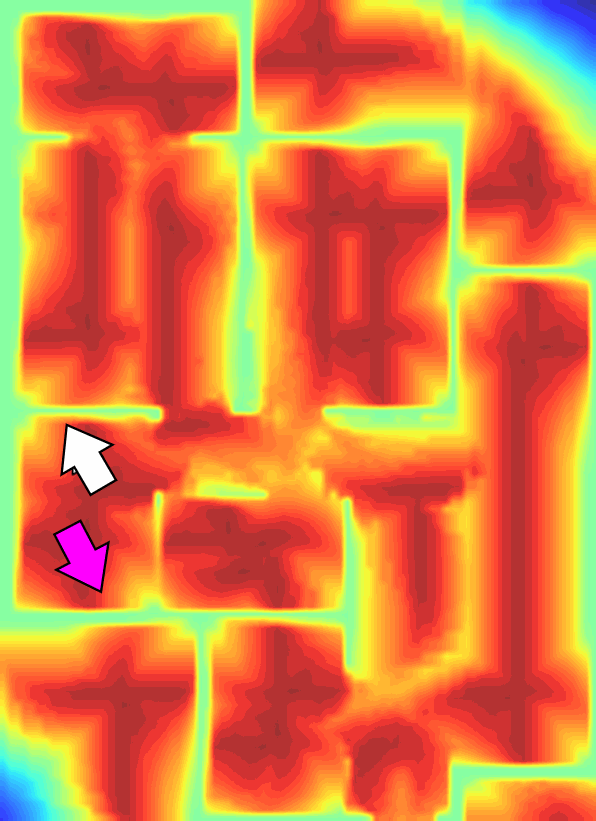}\\[0.5ex]
  $(3.67m, 178^\circ)$} &
\parbox[c]{\linewidth}{%
  \centering
  \includegraphics[width=\linewidth]{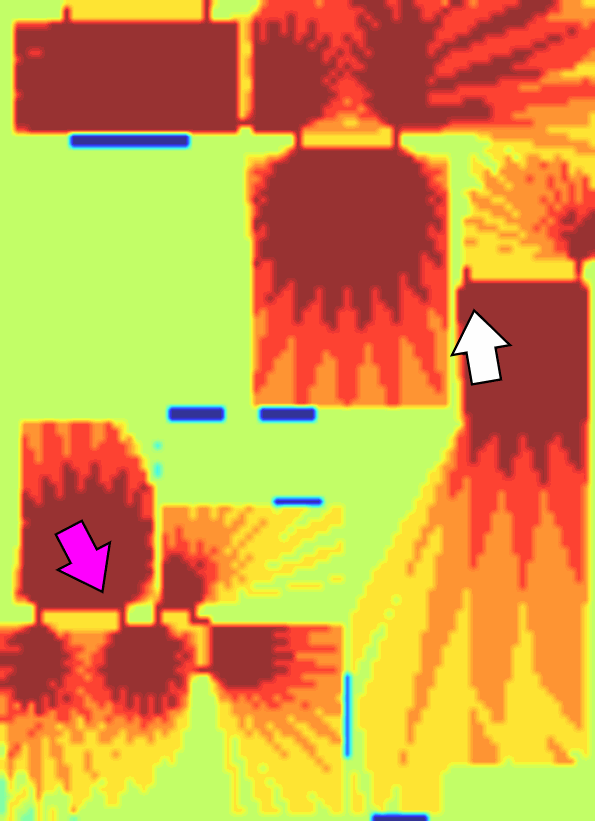}\\[0.5ex]
  $(1.94m, 162^\circ)$} &
\parbox[c]{\linewidth}{%
  \centering
  \includegraphics[width=\linewidth]{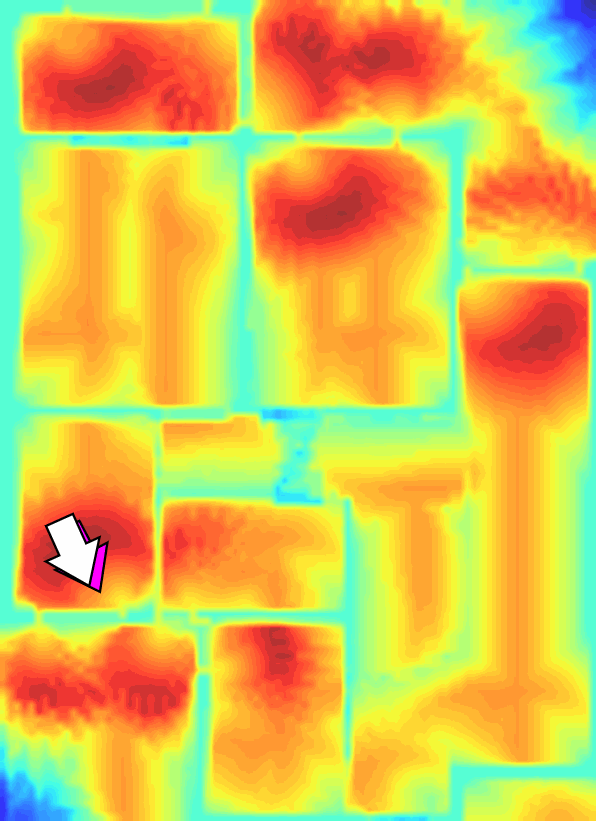}\\[0.5ex]
  $(0.09m, 2^\circ)$} \\
\parbox[c]{\linewidth}{%
  \centering
  \includegraphics[width=\linewidth]{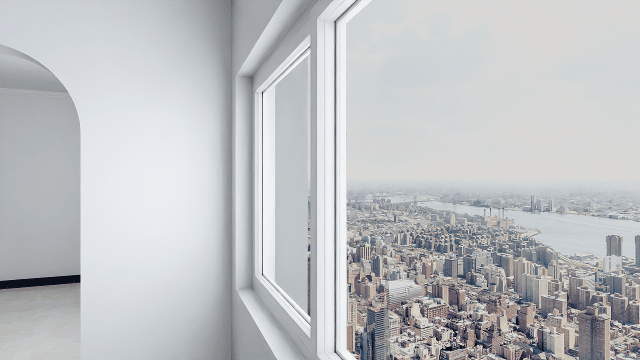}\\[0.5ex]} &
\parbox[c]{\linewidth}{%
  \centering
  \includegraphics[width=\linewidth]{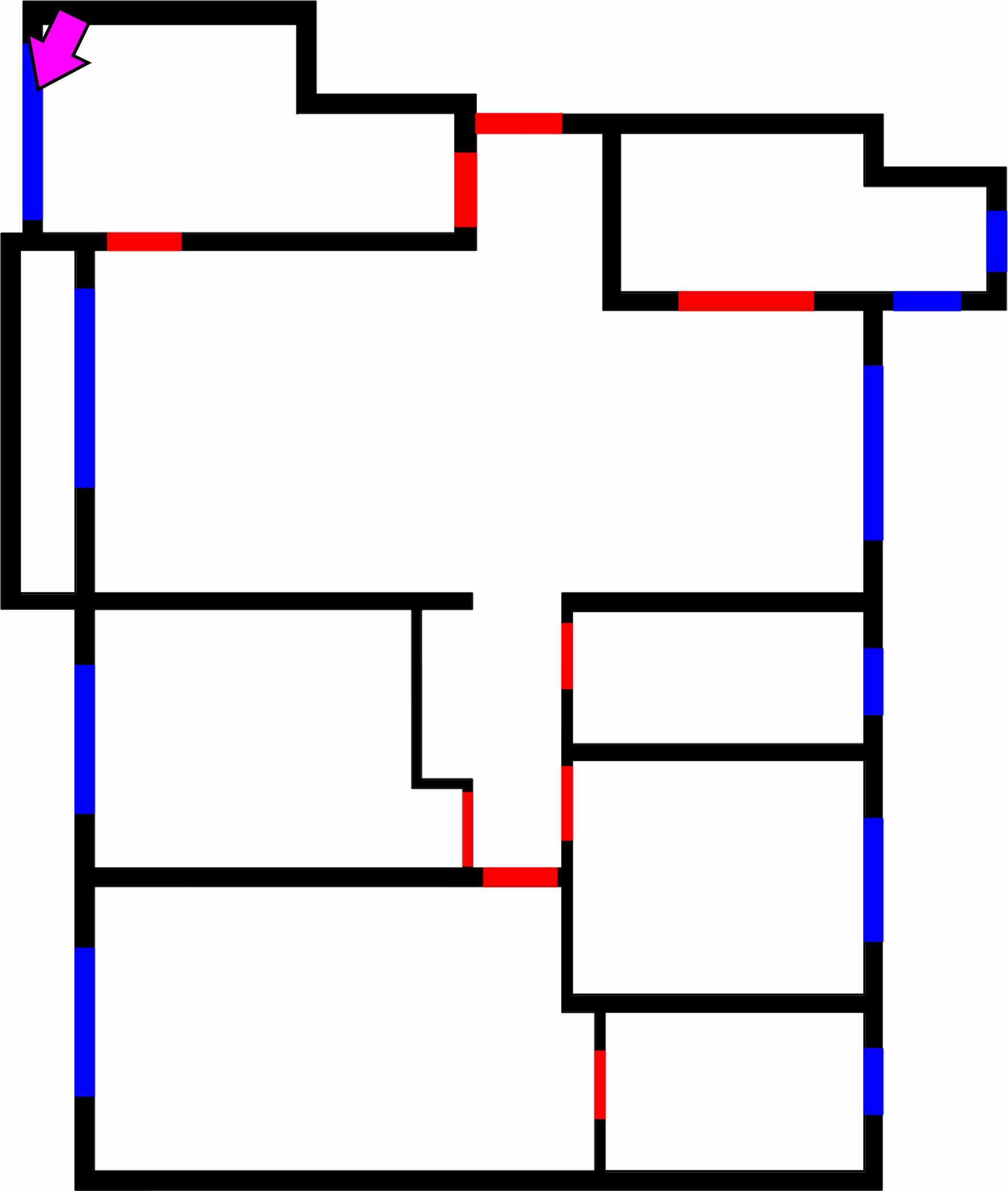}\\[0.5ex]
  \vphantom{$(7.08m, 179^\circ)$}} &
\parbox[c]{\linewidth}{%
  \centering
  \includegraphics[width=\linewidth]{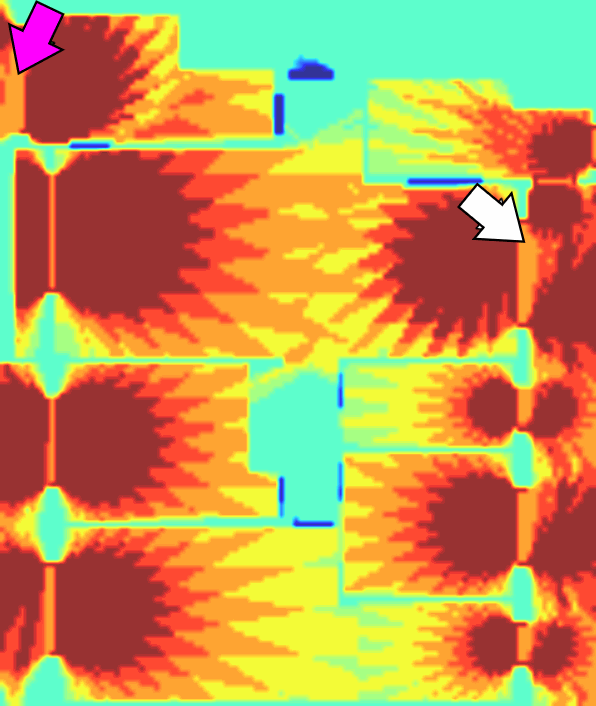}\\[0.5ex]
  $(7.08m, 179^\circ)$} &
\parbox[c]{\linewidth}{%
  \centering
  \includegraphics[width=\linewidth]{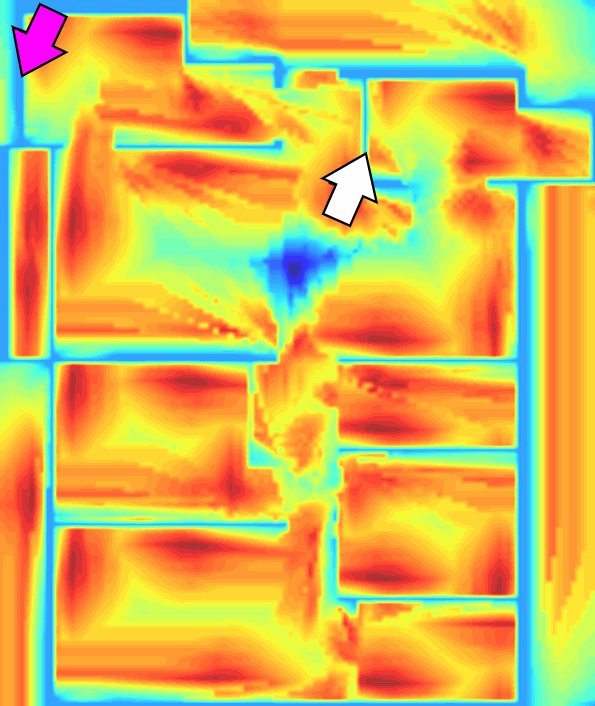}\\[0.5ex]
  $(9.49m, 71^\circ)$} &
\parbox[c]{\linewidth}{%
  \centering
  \includegraphics[width=\linewidth]{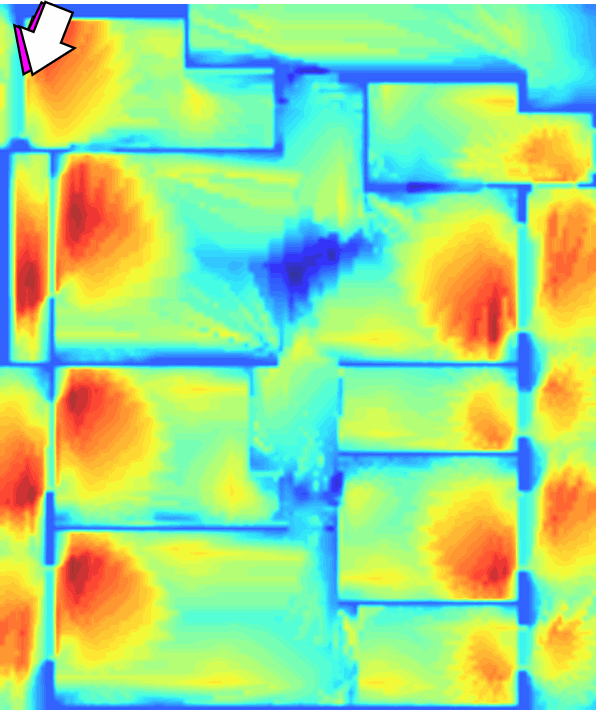}\\[0.5ex]
  $(0.15m, 1^\circ)$} \\
\parbox[c]{\linewidth}{%
  \centering
  \includegraphics[width=\linewidth]{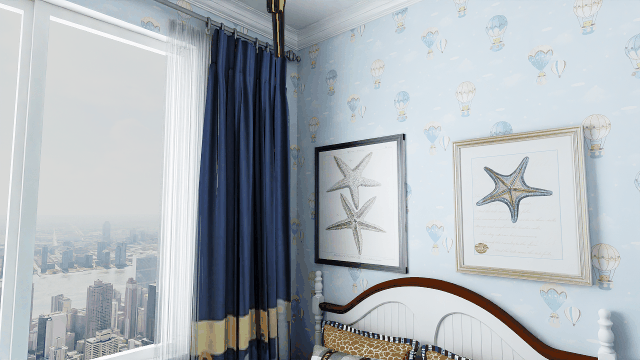}\\[0.5ex]} &
\parbox[c]{\linewidth}{%
  \centering
  \includegraphics[width=\linewidth]{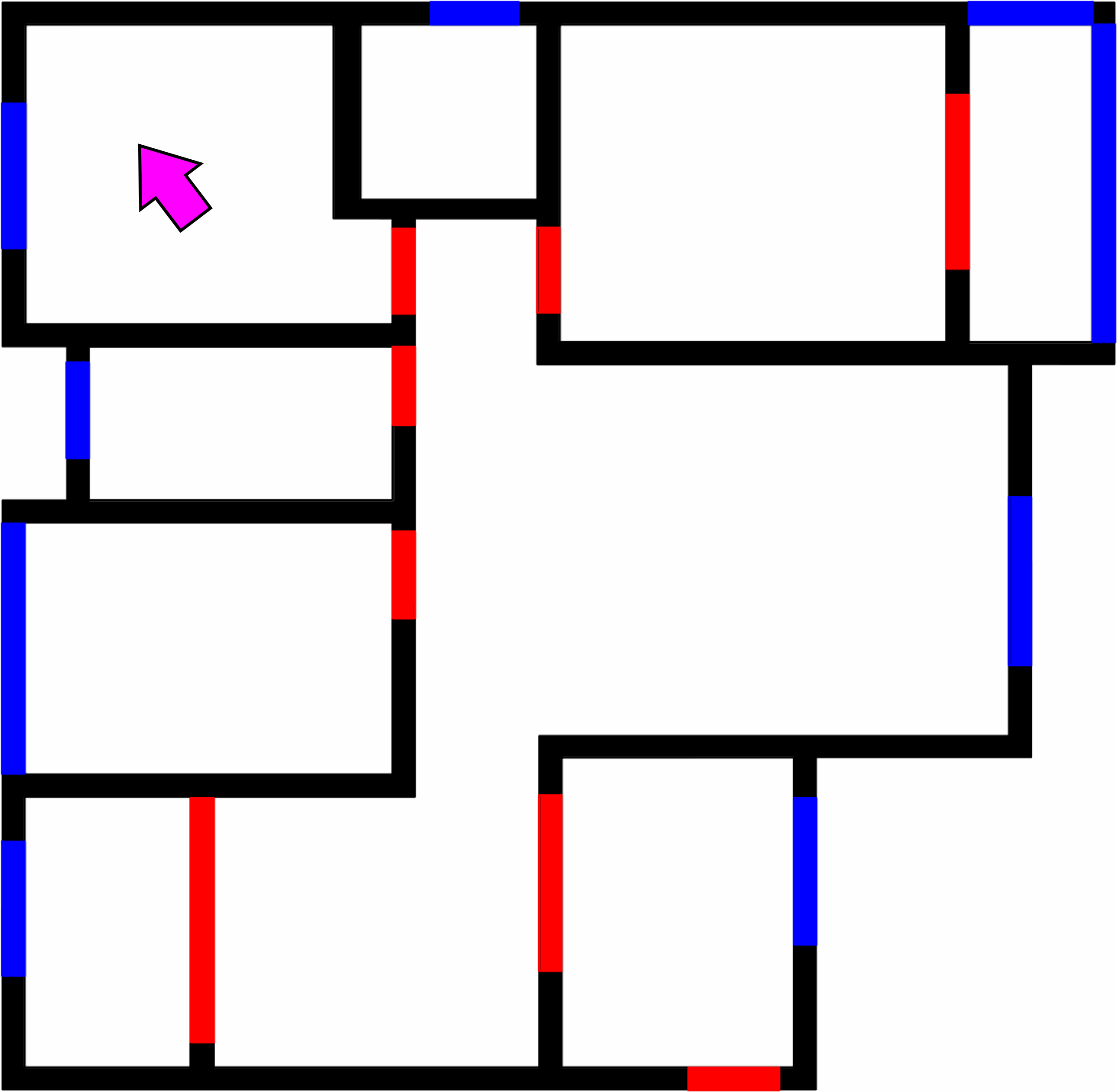}\\[0.5ex]
  \vphantom{$(7.85m, 168^\circ)$}} &
\parbox[c]{\linewidth}{%
  \centering
  \includegraphics[width=\linewidth]{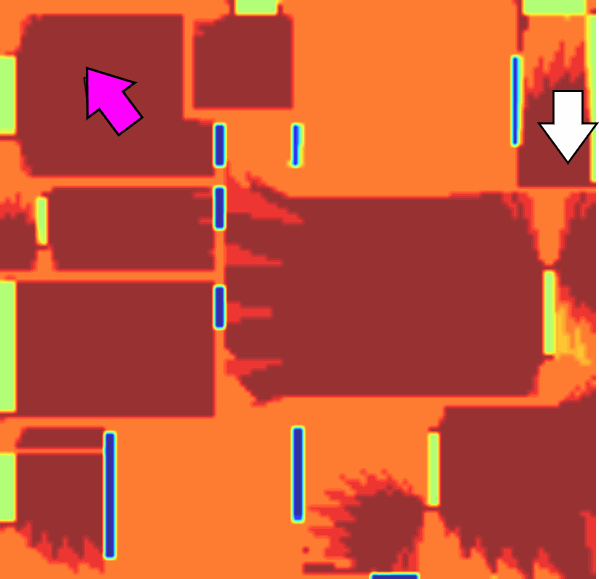}\\[0.5ex]
  $(7.85m, 168^\circ)$} &
\parbox[c]{\linewidth}{%
  \centering
  \includegraphics[width=\linewidth]{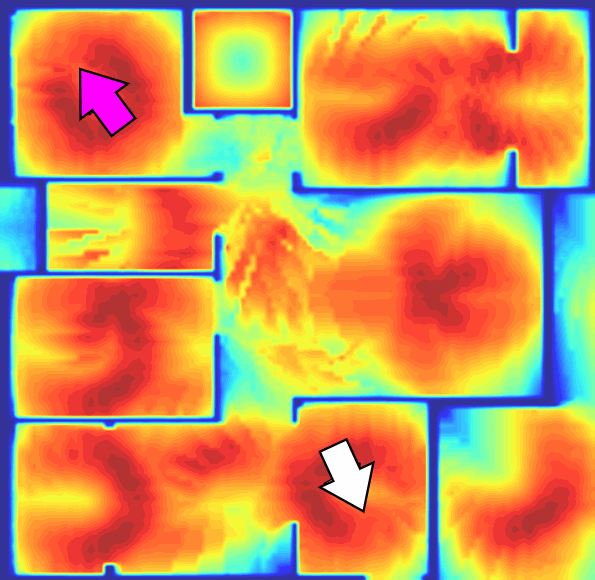}\\[0.5ex]
  $(2.59m, 8^\circ)$} &
\parbox[c]{\linewidth}{%
  \centering
  \includegraphics[width=\linewidth]{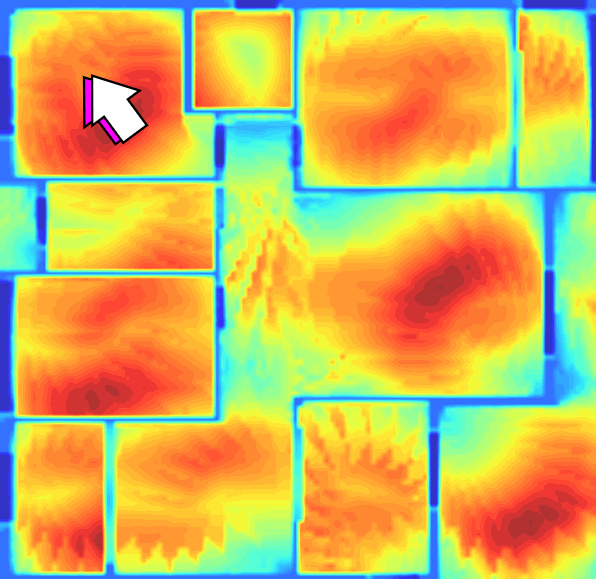}\\[0.5ex]
  $(0.13m, 2^\circ)$} \\
\hline\hline
\\
\end{tabular}
    }
    \vspace{-14pt}
    \caption{Qualitative Comparison of Depth-Only, Semantic-Only, and Fused Structural-Semantic Probability Volumes. Below each map we report the localization error in meters and degrees. Warmer colors correspond to regions with higher probabilities.}
    \label{fig:ablation_fig}
\end{figure}

We analyze the impact of different combinations of predicted depth and semantic features on the S3D dataset. 
Figure~\ref{fig:recall_plot} summarizes the recall performance for various depth and semantic weight configurations used to compute the \emph{structural-semantic probability volume}. In Figure \ref{fig:ablation_fig} we visualize examples from two extreme scenarios (depth only and semantic only) and from the configuration adopted in our work ($[w_s,w_d] = [0.4,0.6]$), selected according to the validation set.

As can be observed from Figure~\ref{fig:ablation_fig} (and also reflected in Figure~\ref{fig:recall_plot}), relying solely on semantic rays tends to produce a more diffused probability volume with multiple ambiguous candidate locations. This occurs because most images contain a single semantic object of a standard size, allowing an accurate ray pattern to be fitted in more than one location. Similarly, depth cues alone also suffer from ambiguity, particularly in repetitive environments or when the image captures three or fewer walls, which leads to uncertainty in the localization estimate. However, when these semantic cues are combined with depth rays, the resulting probability volume becomes significantly more concentrated. This integration effectively filters out spurious candidates and sharpens the localization estimate. 

Additionally, the supplementary material offers a detailed analysis of how varying the Top-K candidates influences localization refinement, along with additional experiments that motivate various design choices, such as using hard thresholds in multiple steps of our pipeline.

\subsubsection*{Runtime Analysis}

We report runtime performance breakdown in Table~\ref{tab:aggregated_metrics}, using the same parameters employed in Table \ref{tab:comparison_s3d_and_zind} with varying K values. 
As shown in the table, the per-image inference time increases with the number of candidate refinements, \(K\), reflecting the additional computations required during refinement. All experiments were conducted on a single CPU without multithreading to avoid introducing bias. Future work could explore parallelizing the refinement stage by computing all ground-truth rays simultaneously for further speed-up. Importantly, even at \(K=5\)---the setting we adopt in our work, the inference time remains reasonable, striking a practical balance between computational cost and improvements in localization accuracy. We further observe that the prediction and localization steps require similar amounts of time, while the refinement step grows monotonically as \(K\) increases.


\begin{table}[t]
  \centering
  \small                          
  \setlength{\tabcolsep}{2pt}     
  \renewcommand{\arraystretch}{1.2}

  \scalebox{0.90}[1.0]{%
    \begin{tabular}{ccccc}
      \toprule
      {\normalsize\bfseries $K$}
      & {\normalsize\bfseries Prediction}
      & {\normalsize\bfseries Loc}
      & {\normalsize\bfseries Refin}
      & {\normalsize\bfseries Total} \\
      \midrule
      1 & $0.038\pm0.119$ & $0.174\pm0.033$ & $0.141\pm0.073$ & $0.364\pm0.142$ \\
      3 & $0.033\pm0.093$ & $0.154\pm0.026$ & $0.356\pm0.119$ & $0.554\pm0.157$ \\
      5 & $0.034\pm0.103$ & $0.155\pm0.029$ & $0.577\pm0.185$ & $0.778\pm0.218$ \\
      \bottomrule
    \end{tabular}%
  }

  \caption{Performance breakdown over different Top-K values. Each entry is mean\,$\pm$\,std (s). \textit{Prediction} denotes the ray predictions, \textit{Loc} refers to the localization process, and \textit{Refin} represents the refinement stage, during which candidate locations are evaluated by computing the ground truth rays.}
  \label{tab:aggregated_metrics}
\end{table}

\section{Conclusion}




In this work, we presented a semantic-aware localization framework that extends floorplan-based camera localization by fusing semantic labels with geometric depth cues. Our approach leverages a novel semantic ray prediction network alongside an established depth estimation method to generate a semantic-structural probability volume, which significantly improves localization accuracy, especially in environments with repetitive or ambiguous structural patterns.

Our extensive experiments on the S3D and ZInD datasets demonstrate that integrating semantic cues effectively resolves depth-based ambiguities and consistently outperforms state-of-the-art methods such as F3Loc and LASER. Ablation studies confirm that a balanced combination of depth and semantic information, coupled with a coarse-to-fine localization strategy and the use of room labels, yields optimal performance. 

Looking forward, extending our framework to incorporate additional semantic labels and other modalities, such as textual information, promises to further enhance localization robustness in challenging indoor settings. In general, our approach represents an important step towards accurate and reliable indoor localization systems by effectively leveraging semantic and geometric cues.

{
\small
\bibliographystyle{ieee_fullname}
\bibliography{references}
}
\clearpage
\appendix

\section{Additional Details}
\label{sec:details}
In this section, we provide detailed information on the network architecture, training procedure, evaluation pipeline, baselines, dataset handling, and parameter settings used in our experiments.

\subsection{Network Architecture and Design Choices}
\label{sec:architecture}
Our model adopts a ResNet50 backbone pretrained on ImageNet to extract features from the input RGB image. The extracted feature map (of dimension 2048) is then reduced to 128 channels via a convolution followed by batch normalization and ReLU activation (implemented in our custom \texttt{ConvBnReLU} module). These features are further projected to a 48-dimensional space using a linear layer.

To preserve spatial information, a positional encoding is computed from normalized $(x,y)$ coordinates using a small MLP with a \texttt{Tanh} activation. Two sets of learnable query tokens are introduced:
\begin{itemize}
    \item A single CLS token for predicting a global room-type label.
    \item 40 ray tokens for predicting semantic rays.
\end{itemize}
Both sets of tokens attend to the flattened spatial features using a single-head cross attention module. The ray tokens are additionally processed by a self-attention block (with residual connections and a feed-forward network) followed by an MLP to produce per-ray logits over semantic classes. The room token is processed similarly to yield room type logits.

\subsection{Training Settings and Hyperparameters}
\label{sec:training_details}
As mentioned in the main paper, our semantic network is implemented within a PyTorch Lightning module to perform multi-task predictions, simultaneously producing 40 semantic ray outputs (one per ray) and one global room-type label. During training, the predicted ray outputs (with shape \((N, 40, \text{num\_ray\_classes})\)) are supervised via cross-entropy loss against the ground-truth semantic labels (shape \((N, 40)\)), while the global room-type prediction (with shape \((N, \text{num\_room\_types})\)) is similarly trained using cross-entropy loss. The overall loss is defined as the sum of these two components. We optimize the network using the Adam optimizer with a learning rate of \(1\times10^{-3}\) and a batch size of 16.

\subsection{Dataset Descriptions}
\label{datasets_supp}

Additional dataset processing details are provided here for clarity.



\paragraph{\textbf{S3D}} We use the fully furnished, perspective dataset of Structured3D (S3D) with the official splits and processing protocol.
\paragraph{\textbf{ZInD}} For ZInD, we follow the official splits and prior works to generate a fixed-size dataset by cropping each panorama to a single 80° FoV, 0° yaw perspective image.


\subsection{Baseline Methods}
We compare our method against several baselines to assess its performance under a consistent evaluation protocol.

\paragraph{\textbf{F3Loc}} For the F3Loc baseline, we use the publicly available code from the official repository and made a some modifications to the way we calculate rays and identify walls for the ZInD dataset, but. For the S3D dataset, we report the official paper results as we operate on the exact same data split and processing protocol. For the ZInD dataset, we evaluate F3Loc by running its training and inference using the provided code and configuration.

\paragraph{\textbf{LASER}} For the LASER baseline, we use the official implementation available from the authors. Since the provided code runs on both datasets, we execute LASER as-is. For S3D, we follow F3Loc by evaluating on the official fully furnished perspective dataset. For ZInD, we run the official training and evaluation code while adjusting the configuration to crop the panoramas to an 80° FoV and to disable random view augmentations, as detailed in Section~\ref{datasets_supp}.

\subsection{Additional Implementation Details}
\subsubsection{Semantic Interpolation via Majority Voting}
As described in the main section, we introduce a majority voting algorithm to interpolate the predicted \(l\) semantic rays into a smaller subset. As shown in our ablation study, this interpolation alone yields a 4.2\% improvement in 1m recall. The detailed algorithm is provided in Algorithm~\ref{alg:semantic_interp_majority}.
\begin{algorithm}[t]
\caption{Semantic Ray Interpolation with Majority Voting}
\label{alg:semantic_interp_majority}
\begin{algorithmic}[1]
\Require \\
\begin{enumerate}
    \item A semantic ray vector \(r\) of length \(N\).
    \item Field-of-view \(\text{fov} = 80^\circ\).
    \item Desired number of rays \(N_d\).
    \item Desired angular gap \(\Delta\theta\).
    \item Window size \(w\) for majority voting.
\end{enumerate}
\State Compute the angle between original rays: \(\Delta\alpha \).
\State Compute the center index: \(c \leftarrow \lfloor N/2 \rfloor\).
\State Initialize an empty semantic ray vector \(r_{\text{interp}}\).
\For{\(i = 0\) to \(N_d - 1\)}
    \State Compute the desired angle relative to the center:
    \[
    \theta_i \leftarrow \left(i - \lfloor N_d/2 \rfloor\right) \times \Delta\theta.
    \]
    \State Compute the index offset:
    \[
    o \leftarrow \frac{\theta_i}{\Delta\alpha}.
    \]
    \State Determine the target index:
    \[
    \text{idx} \leftarrow \text{round}(c + o).
    \]
    \State Collect neighbor labels: 
    \[
    \text{neighbors} \leftarrow \{\, r[j] \mid j = \text{idx}-w,\,\dots,\,\text{idx}+w \}.
    \]
    \State Determine the majority label \(l^*\).
    \State Append \(l^*\) to \(r_{\text{interp}}\).
\EndFor
\Return \(r_{\text{interp}}\).
\end{algorithmic}
\end{algorithm}

\subsubsection{Ray Similarity Measurement}
To assess the alignment between the predicted rays and the candidate rays in our refinement procedure, we compute a similarity score that combines both depth and semantic discrepancies. Specifically, we calculate the L1 distance between the predicted depth rays and the candidate depth rays to capture the geometric error, and we compute a semantic error as the mean mismatch between the predicted semantic labels and the candidate semantic labels. These two error metrics are then combined using a weighted sum:
\[
\text{score} = \alpha \cdot \text{depth\_error} + (1-\alpha) \cdot \text{semantic\_error},
\]
where the depth error is computed as the average absolute difference between corresponding depth values, and the semantic error is quantified as the average binary mismatch between semantic labels. In all our experiments, we set \(\alpha\) equal to \(w_d\), the weight assigned to the depth probability volume in our fusion equation.


\subsection{System Configuration}
\label{sec:system_config}
All training experiments were conducted on a virtual machine with the following specifications:
\begin{itemize}
    \item \textbf{CPUs:} 12 cores (Intel Xeon E5-2690 v4 @ 2.60GHz)
    \item \textbf{GPU:} Tesla V100-PCIE GPUs (with 16GB memory each)
\end{itemize}

These hardware details ensure reproducibility and highlight the computational resources available during training.

\section{Additional Ablation Studies}
\label{sec:ablation}
In this section, we present a series of ablation studies to evaluate key components of our localization pipeline. In Section~\ref{sec:room_polygons} we analyze the impact of using external room-polygon masks. Section~\ref{sec:topk} examines the effect of varying Top-K candidate selections and refinement parameters. Finally, in Section~\ref{sec:location_extraction} we investigate the influence of the refinement threshold \(\delta_{\text{res}}\) on balancing fine and coarse localization accuracy.
\subsection{Effect of Room Polygon Usage}
\label{sec:room_polygons}

\begin{figure}[h]
    \centering
    \begin{subfigure}[t]{0.2\textwidth}
        \centering
        \includegraphics[width=\textwidth]{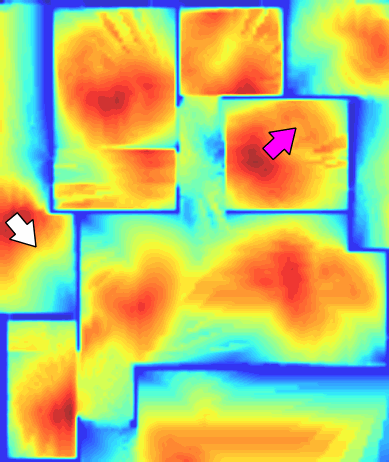}
        \caption{no mask}
        \label{fig:no_mask}
    \end{subfigure}
    \quad
    \begin{subfigure}[t]{0.2\textwidth}
        \centering
        \includegraphics[width=\textwidth]{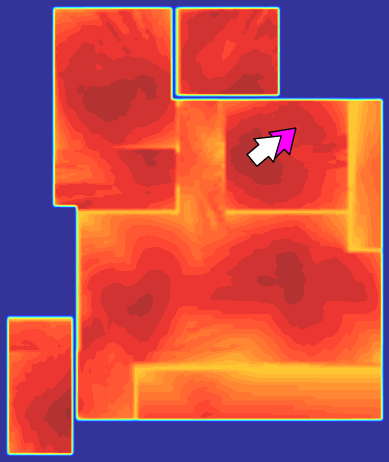}
        \caption{with mask}
        \label{fig:with_mask}
    \end{subfigure}
    \caption{Comparison of the scene without mask (a) and with mask (b).}
    \label{fig:comparison}
\end{figure}

As part of our usage of room-polygon masks, we also compare the performance of using external house-area masks versus not using them. As shown in Figure~\ref{fig:comparison}, we use a mask to exclude points from outside the house. This avoids matching windows and corners that lie beyond the interior. Notably, when the highest-probability location is masked out, the next best match is closer to the ground-truth location, yielding an improvement.

Table~\ref{tab:mask_comparison} presents a comparison of the recall obtained by our method with and without external masking, demonstrating that this procedure does not yield any substantial gains.

\begin{table}[h]
    \centering
    \begin{tabular}{c cccc}
    \toprule
    \textbf{Mask Setting} & 0.1\,m & 0.5\,m & 1\,m & 1\,m 30$^\circ$ \\
    \midrule
    with  & \textbf{5.63} & \textbf{45.67} & \textbf{59.36} & \textbf{57.82}\\
    without & 5.13  & 45.07 & 59.24 & 57.61 \\
    \bottomrule
    \end{tabular}
    \caption{Comparison of localization accuracy on S3D with and without external house-area masks.}
    \label{tab:mask_comparison}
\end{table}

\subsection{Impact of Top-K Candidate Selection on Test Set Performance}

We further analyze our coarse-to-fine approach by conducting an experiment to evaluate the effect of selecting different numbers of Top-K candidates. Table~\ref{tab:localization_extraction_ablation} details the impact of various Top-K values on the localization refinement. We observe that as \(k\) increases, the overall localization accuracy improves. In particular, the largest improvement is achieved when increasing from Top-1 to Top-2 candidates, which is sensible since over 70\% of the ground-truth locations lie within the Top-1 and Top-2 candidate set. Beyond Top-2, while further increases in \(k\) yield additional improvements, these gains are minor compared to the initial boost. This is likely due to prediction errors and noise. As \(k\) increases, additional candidates may include rays that were previously interpolated out, leading to mislocalizations when they are erroneously matched.
\newcommand{\scoreonly}[1]{\vtop{\hbox to 1.8cm{\strut #1}}}
\newcommand{\scorewith}[2]{\vtop{\hbox to 1.8cm{\strut #1\hfill\,\textcolor[HTML]{006400}{\small{#2}}}}}

\begin{table}[t]
    \centering
    \resizebox{\columnwidth}{!}{
    \begin{tabular}{cc|cccc}
        \hline
        \textbf{TopK} & \textbf{Method} & \textbf{0.1m} & \textbf{0.5m} & \textbf{1m} & \textbf{1m 30°} \\
        \hline
        \multirow{3}{*}{Top1} 
            & $\text{No refine}$ & \scoreonly{4.65}   & \scoreonly{38.35}  & \scoreonly{49.40}  & \scoreonly{48.44} \\    
            & $\text{Ours}_s$ & \scoreonly{4.73}   & \scoreonly{38.35}  & \scoreonly{49.59}  & \scoreonly{48.59}  \\
            & $\text{Ours}_r$ & \scorewith{5.29}{11.84\%\(\uparrow\)} & \scorewith{42.81}{11.63\%\(\uparrow\)} & \scorewith{55.76}{12.44\%\(\uparrow\)} & \scorewith{54.30}{11.74\%\(\uparrow\)} \\
        \hline
        \multirow{2}{*}{Top2} 
            & $\text{Ours}_s$ & \scoreonly{4.96}  & \scoreonly{41.08}  & \scoreonly{52.20}  & \scoreonly{51.39}  \\
            & $\text{Ours}_r$ & \scorewith{5.48}{10.48\%\(\uparrow\)}  & \scorewith{45.31}{10.30\%\(\uparrow\)}  & \scorewith{58.43}{11.93\%\(\uparrow\)}  & \scorewith{57.19}{11.28\%\(\uparrow\)}  \\
        \hline
        \multirow{2}{*}{Top3} 
            & $\text{Ours}_s$ & \scoreonly{5.23}  & \scoreonly{41.27}  & \scoreonly{52.96}  & \scoreonly{52.04}  \\
            & $\text{Ours}_r$ & \scorewith{5.34}{2.10\%\(\uparrow\)}  & \scorewith{45.24}{9.63\%\(\uparrow\)}  & \scorewith{58.77}{11.00\%\(\uparrow\)}  & \scorewith{57.28}{10.07\%\(\uparrow\)}  \\
        \hline
        \multirow{2}{*}{Top5} 
            & $\text{Ours}_s$ & \scoreonly{\textbf{5.42}} & \scoreonly{\textbf{41.87}} & \scoreonly{\textbf{53.52}} & \scoreonly{\textbf{52.61}} \\
            & $\text{Ours}_r$ & \scorewith{\textbf{5.70}}{5.17\%\(\uparrow\)} & \scorewith{\textbf{45.53}}{8.74\%\(\uparrow\)} & \scorewith{\textbf{58.78}}{9.83\%\(\uparrow\)} & \scorewith{\textbf{57.49}}{9.28\%\(\uparrow\)} \\
        \hline
    \end{tabular}
    }
    \vspace{-8pt}
    \caption{Ablation study on the coarse-to-fine Top-k selection in the S3D dataset, evaluating the location extraction module and the effect of room type prediction in our pipeline. Recall metrics (in \%) for our methods (\(\text{Ours}_s\) and \(\text{Ours}_r\)) are reported. For each metric, the improvement is shown to the right of the \(\text{Ours}_r\) score in \textcolor[HTML]{006400}{dark green} with an upward arrow indicating the relative improvement over \(\text{Ours}_s\).}
    \label{tab:localization_extraction_ablation}
\end{table}


\subsection{Top-K Location Distribution Analysis}
\label{sec:topk}

To better understand the effectiveness of our coarse-to-fine strategy, we conducted an in-depth study on the impact of selecting the Top-K candidate poses and on the localization accuracy.
For simplicity of this analysis, no angular augmentations were applied in this analysis. all data were collected from the S3D test dataset using the following parameters: \(\delta_{\text{res}} = 1\)\,m, \(\delta_{\text{ang}} = 0^\circ\), and \(\Delta_{\text{max}} = 0^\circ\).

\begin{figure}[htbp]
\centering
\includegraphics[width=\linewidth]{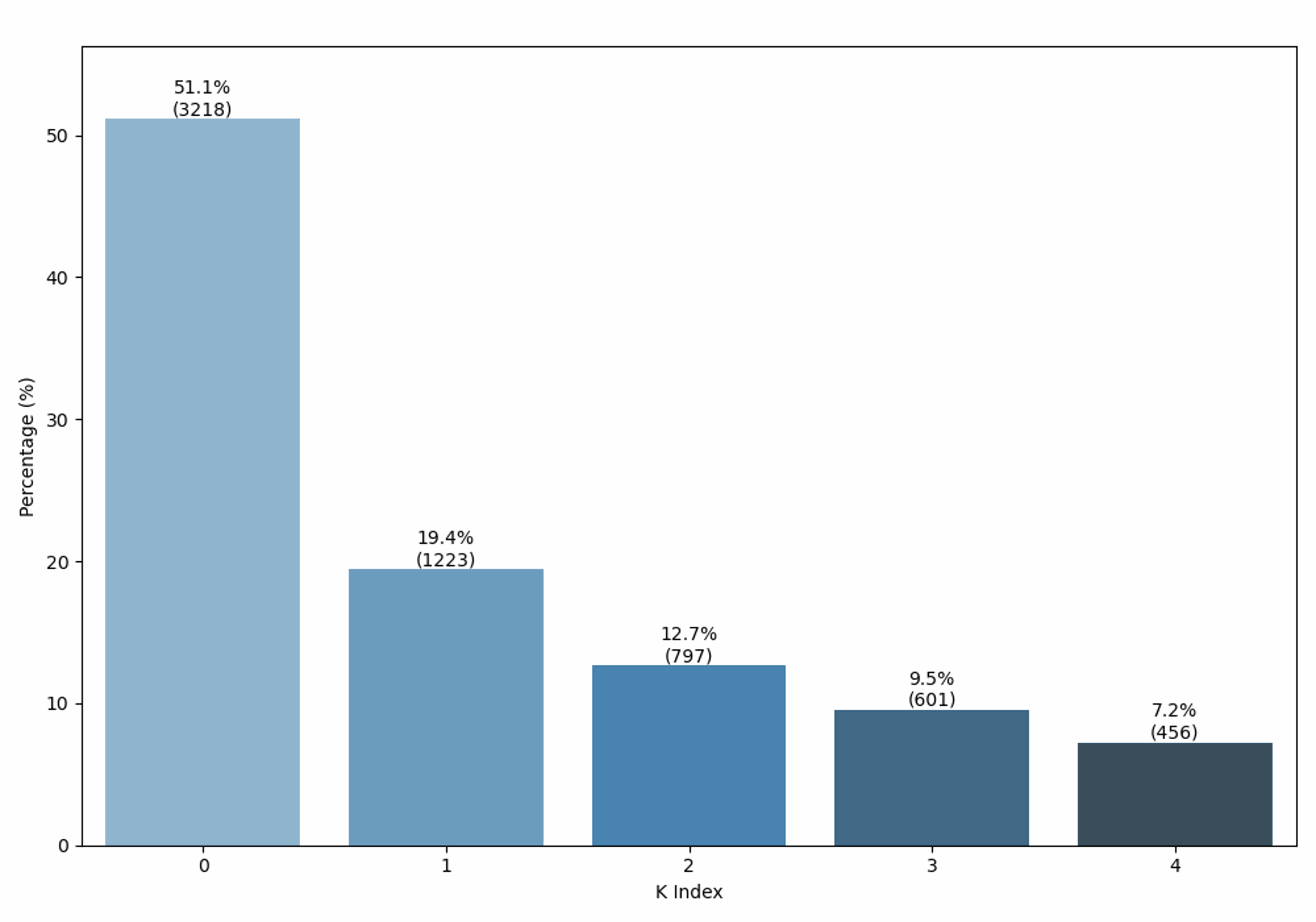}
\caption{Distribution of the best candidate index on the S3D test set. The Top 1 candidate is closest to the ground truth in 51.1\% of cases, followed by the second and third candidates.}
\label{fig:best_index_distribution}
\end{figure}

Figure~\ref{fig:best_index_distribution} presents the candidate ranking distribution. In 51.1\% of cases, the Top 1 candidate is closest to the ground truth, while the second and third candidates account for 19.4\% and 12.7\% of cases, respectively. In this analysis, we maintain a 1\,m exclusion radius around each candidate to emphasize strong mismatches. This motivates refining the Top-K candidates instead of relying solely on the Top-1 candidate during the coarse stage.

\begin{figure}[htbp]
\centering
\includegraphics[width=\linewidth]{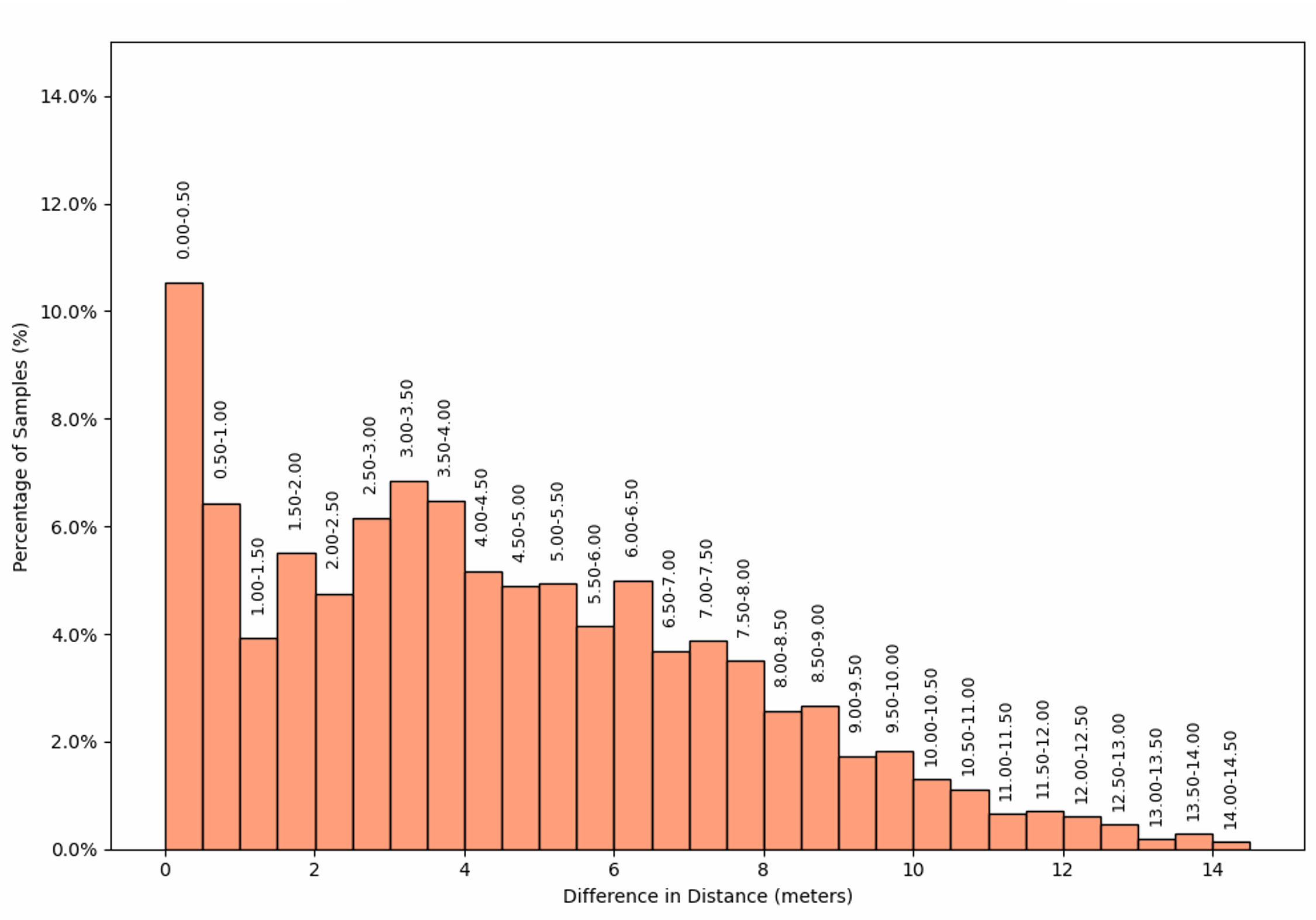}
\caption{Histogram of distance improvements for Top-K selections. Approximately 90\% of the improvements exceed 0.5\,m compared to the highest-scoring candidate (K=0).}
\label{fig:distance_histogram}
\end{figure}

Furthermore, Figure~\ref{fig:distance_histogram} shows that approximately 90\% of the localization improvements occur when a candidate is shifted by more than 0.5\,m relative to the highest-scoring candidate (K=0) in the structural-semantic probability volume. This finding reinforces the benefit of selecting the best candidate among the Top-K predictions.

\begin{figure}[htbp]
\centering
\includegraphics[width=\linewidth]{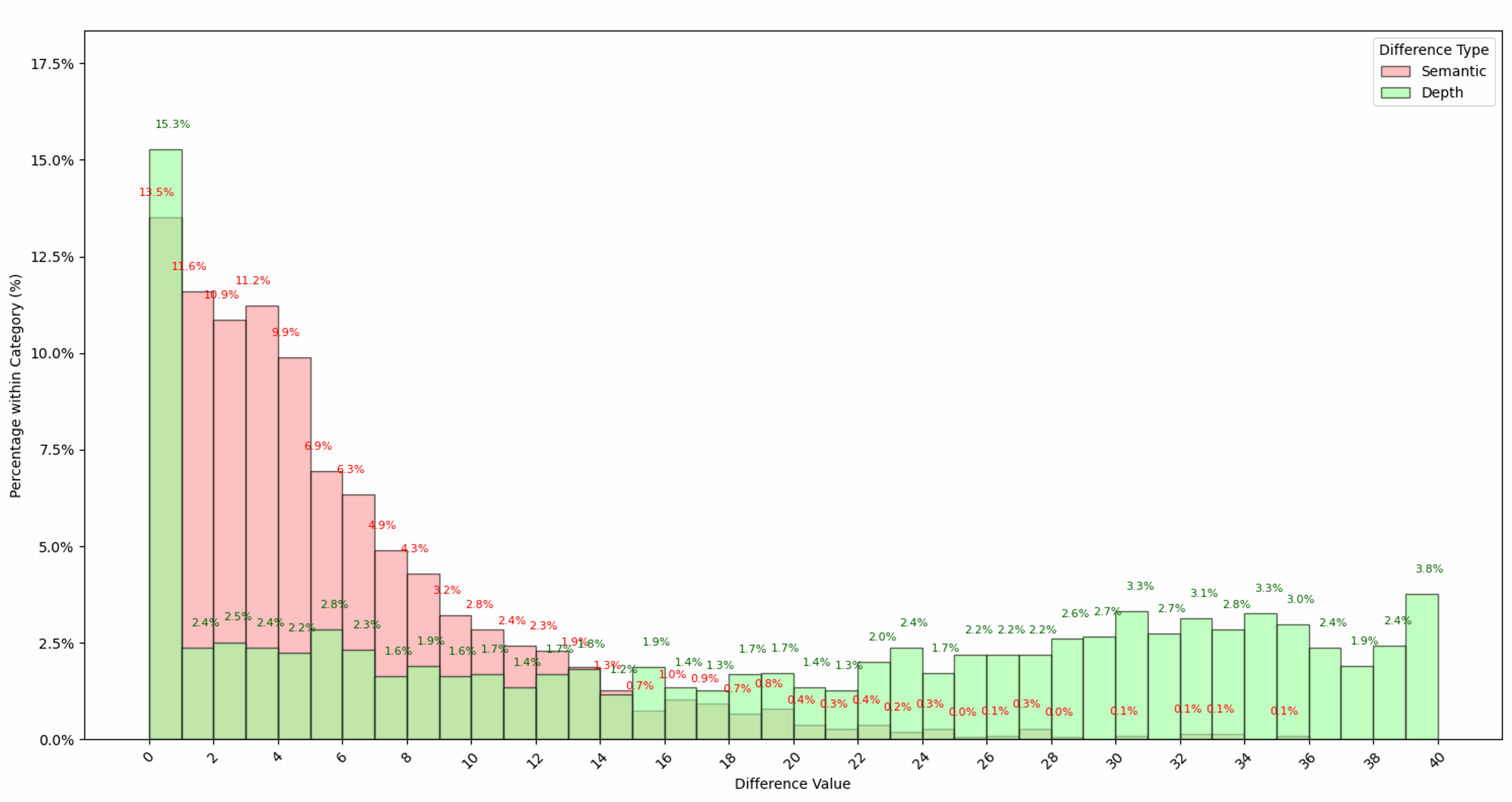}
\caption{Semantic and Depth Ray Differences. The Y-axis represents the percentage of samples, and the X-axis indicates the number of ray differences between the top candidate (K0) and the best candidate, with \textcolor{green}{depth} differences shown in green and \textcolor{red}{semantic} differences in red.}
\label{fig:semantic_depth_differences}
\end{figure}

Figure~\ref{fig:semantic_depth_differences} illustrates the discrepancies between semantic and depth ray predictions when the top candidate (K0) is not the best match. The trend of decreasing sample percentages with increasing differences in the semantic rays confirms that even small changes in semantic cues are critical for accurate localization. This effect is also evident when a semantic label resolves ambiguity between two structurally identical environments, further emphasizing the importance of integrating semantics into the localization process. Note that we consider two depth rays to be identical if they differ by less than 10\,cm.

\begin{figure}[t]
\centering
\begin{tabular}{ccc}
  \begin{minipage}[b]{0.30\linewidth}
    \centering
    \includegraphics[width=\linewidth]{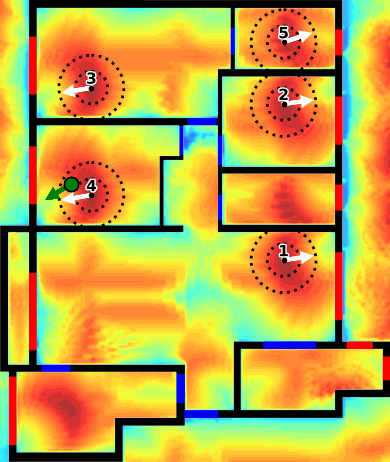}\\[0.5ex]
    \includegraphics[width=\linewidth]{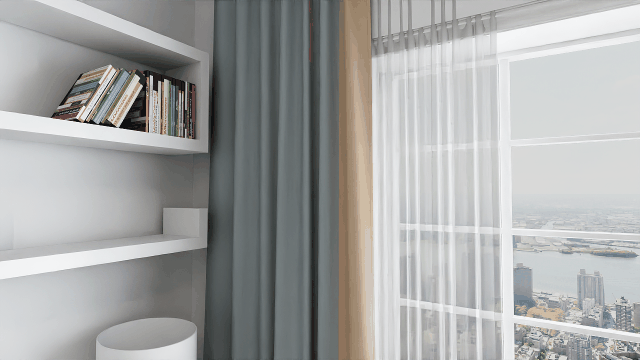}
  \end{minipage} &
  \begin{minipage}[b]{0.30\linewidth}
    \centering
    \includegraphics[width=\linewidth]{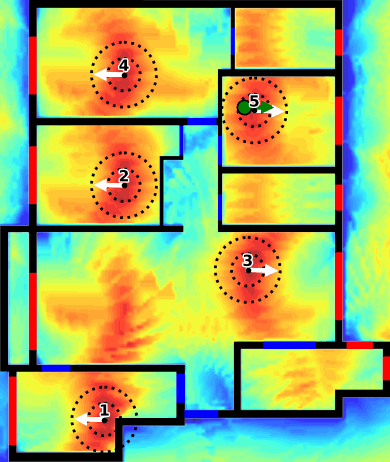}\\[0.5ex]
    \includegraphics[width=\linewidth]{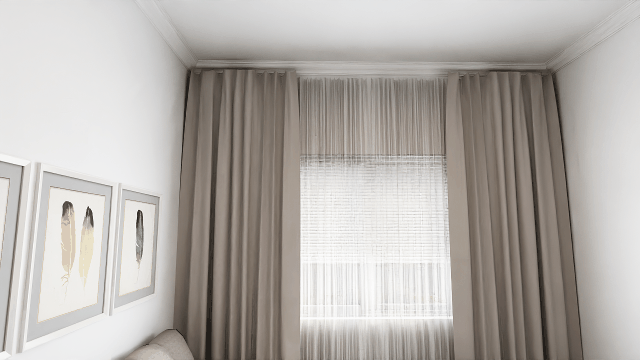}
  \end{minipage} &
  \begin{minipage}[b]{0.30\linewidth}
    \centering
    \includegraphics[width=\linewidth]{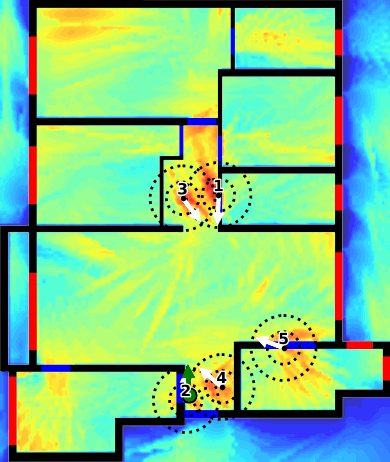}\\[0.5ex]
    \includegraphics[width=\linewidth]{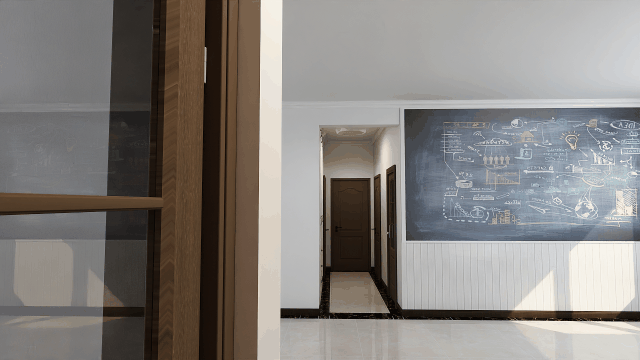}
  \end{minipage}
\end{tabular}
\caption{Illustrative examples of the impact of repetitive patterns on localization accuracy. This figure demonstrates difficult cases where, despite accurate semantic and depth predictions, floorplan localization remains challenging. The identical depth and semantic patterns may result in the top score not corresponding to the ground truth location, which motivates our analysis of Top-K recall.}
\label{fig:top_k_example}
\end{figure}

In Figure~\ref{fig:top_k_example} we illustrate the impact of different Top-K selections on localization accuracy. In many cases, especially in environments with repetitive patterns, the Top-1 candidate does not necessarily correspond to the correct prediction (as can also be seen quantitatively in Figure~\ref{fig:best_index_distribution}).

In Table~\ref{tab:recall_metrics_top_k}, we observe numerically that recall improves drastically as we compute recall within the Top-K candidates. This indeed indicates that our pipeline strongly captures the true location of the images within the top results, but it still remains a challenge to extract the correct location.
\begin{table}[h]
    \centering
    \begin{tabular}{lcccc}
        \hline
        \textbf{Top K} & \textbf{0.1m} & \textbf{0.5m} & \textbf{1m} & \textbf{1m 30°} \\
        \hline
        Top 2 & 7.45  & 57.55  & 70.75  & 69.45 \\
        Top 3 & 7.85  & 63.60  & 78.98  & 76.65 \\
        Top 5 & 8.82  & 69.18  & 85.69  & 83.18 \\
        \hline
    \end{tabular}
        \caption{Recall metrics for different \(K\) values evaluated on the S3D dataset. Recall is defined as the percentage of samples for which the ground truth location is within a specified distance threshold of at least one of the Top \(K\) candidate locations extracted from the probability volume. Higher \(K\) values lead to improved recall, as more candidate locations are considered. For this experiment, we exclude the room-aware module to specifically isolate the effect of the refinement module.}
    \label{tab:recall_metrics_top_k}
\end{table}



\subsection{Ablation on Recall With Different \(\delta_{\text{res}}\) }
\label{sec:location_extraction}
As shown in Table~\ref{tab:unified_refinement_params}, the refinement threshold \(\delta_{\text{res}}\) plays a critical role in balancing fine and coarse localization accuracy. In particular, when using a lower \(\delta_{\text{res}}\) value (0.05\,m), we observe a significant improvement at the fine accuracy threshold (0.1\,m), achieving a recall of 18.40\%. In contrast, a higher \(\delta_{\text{res}}\) value (0.5\,m) yields better performance at the coarser thresholds (0.5\,m, 1\,m, and 1\,m 30$^\circ$). This demonstrates the benefit of customizing the refinement process to meet specific application needs, thereby making it a flexible procedure.

\begin{table}[htbp]
\centering
\begin{tabular}{c cccc}
\toprule
\(\delta_{\text{res}}\) (m) & 0.1\,m & 0.5\,m & 1\,m & 1\,m 30$^\circ$ \\
\midrule
0.05 &\textbf{ 18.40} & 63.02  & 71.60  & 70.08 \\
0.2  & 12.94 & 64.37  & 73.14  & 71.57 \\
0.5  & 8.84  & \textbf{67.07}  & 77.25  & 75.33 \\
1    & 7.85  & 63.60  & \textbf{78.98}  & \textbf{76.65} \\
\bottomrule
\end{tabular}
\caption{Recall performance on the S3D dataset for candidate refinement using the Top 3 candidates. Recall is defined as the percentage of test instances for which at least one of the Top 3 refined candidate poses falls within the specified distance thresholds (0.1\,m, 0.5\,m, 1\,m) and within a 30° orientation tolerance at 1\,m, evaluated under different \(\delta_{\text{res}}\) values.}

\label{tab:unified_refinement_params}
\end{table}

\subsection{Integrating Our Refinement into F3Loc}
Table~\ref{tab:comparison_s3d_and_zind_F3loc_with_refine} quantifies the impact of our refinement module on the baseline F3Loc across both the S3D and ZInD datasets. By incorporating the refinement stage, F3Loc’s recall gains substantial improvements in every threshold (e.g., R@1 m30° on S3D rises from 21.3 to 29.6), demonstrating that our refinement module is indeed effective and substantially enhances localization performance. However, even with refinement, F3Loc+Refine still falls short of the recall achieved by our full method (both with and without room-aware predictions), which underlines that the semantics awareness of our method achieves significant gains beyond what geometric refinement alone can provide.
\begin{table}[t]
\centering
\begin{tabular}{lcccc}
\toprule
\multicolumn{5}{c}{\textbf{S3D R@}} \\
\midrule
Method & 0.1m & 0.5m & 1m & 1m 30° \\
\midrule
F3Loc & 1.5 & 14.6 &  22.4 &  21.3 \\
F3Loc + Refine & 2.74 & 23.29 &  30.74 &  29.59 \\
$\text{Ours}_s$ & 5.42 &  41.87 & 53.52 & 52.61   \\
$\text{Ours}_r$ & \textbf{5.70} & \textbf{45.53} & \textbf{58.78} & \textbf{57.49} \\
\midrule
\multicolumn{5}{c}{\textbf{ZInD R@}} \\
\midrule
Method & 0.1m & 0.5m & 1m & 1m 30° \\
\midrule
F3Loc & 0.67 & 7.90 & 15.07 & 11.46 \\
F3Loc + Refine & 1.21 & 10.46 & 16.94 & 14.21 \\
$\text{Ours}_s$  &2.98 & 24.00 & 33.96 & 29.30 \\
$\text{Ours}_r$ & \textbf{3.31} & \textbf{26.60} & \textbf{38.01} & \textbf{31.86} \\
\bottomrule
\end{tabular}
\vspace{-8pt}
\caption{Recall performance on the S3D and ZInD datasets. The table reports recall at thresholds of 0.1\,m, 0.5\,m, 1\,m, and 1\,m with a 30° orientation tolerance for the baseline F3Loc with and without our refinement module.}
\label{tab:comparison_s3d_and_zind_F3loc_with_refine}
\end{table}
\section{Additional Experiments and Analysis}
\label{sec:experimental}

\subsection{Probability Volume Fusing Weights}
In our approach, the structural-semantic probability volume is obtained by fusing the depth and semantic probability volumes:
\[
P_c = w_s \cdot P_s + w_d \cdot P_d,
\]
where \(w_d\) and \(w_s\) denote the weights assigned to depth and semantic cues, respectively. We determine the optimal weight configuration by evaluating recall metrics on the validation sets. Below, we report our experiments on the S3D and ZInD datasets.

As in the main paper, all experiments use a floorplan resolution of 0.1\,m and an angular granularity of \(10^\circ\). Specifically, we predict 40 rays per image and interpolate these to 9 rays during the coarse stage of localization. For the Location Extraction module, we set \(\delta_{\text{res}}=0.05\)\,m, \(\delta_{\text{ang}}=5^\circ\), and \(\Delta_{\text{max}}=10^\circ\), and report results using Top \(K = 5\) candidates.



\subsubsection{Performance Breakdown on the S3D Dataset}
Table~\ref{tab:combinations} presents a consolidated view of recall performance for various weight configurations on the S3D validation set. Based on these results, we selected \(w_d = 0.6\) and \(w_s = 0.4\) as our final configuration, as it yielded the best overall performance over the validation split.

\begin{table}[htbp]
\centering
\begin{tabular}{cc cccc}
\toprule
\multicolumn{2}{c}{Weights} & 0.1\,m & 0.5\,m & 1\,m & 1\,m 30$^\circ$ \\
{\footnotesize \(w_d\)} & {\footnotesize \(w_s\)} & & & & \\
\midrule
1.0&0   & 2.83  & 22.31 & 30.27 & 29.05 \\
0.9&0.1 & 4.79  & 34.71 & 44.33 & 43.56 \\
0.8&0.2 & 5.19  & 38.04 & 48.82 & 48.03 \\
0.7&0.3 & \textbf{5.20}  & 38.68 & 49.83 & 49.02 \\
0.6&0.4 & 4.93  & \textbf{39.22} & \textbf{50.16} & \textbf{49.48} \\
0.5&0.5 & 5.17  & 38.31 & 49.44 & 48.64 \\
0.4&0.6 & 4.96  & 37.43 & 48.68 & 47.89 \\
0.3&0.7 & 4.52  & 36.29 & 47.56 & 46.46 \\
0.2&0.8 & 4.21  & 35.01 & 45.66 & 44.55 \\
0.1&0.9 & 4.29  & 34.40 & 44.49 & 43.45 \\
0&1.0   & 0.11  & 3.60  & 8.93  & 7.27 \\
\bottomrule
\end{tabular}
\caption{Recall metrics on the S3D validation set obtained with our model without room aware and refinement.}
\label{tab:combinations}
\end{table}



\subsubsection{Performance Breakdown on the ZInD Dataset}
Table~\ref{tab:zind_model} shows the recall performance on the ZInD validation set for different weight configurations. For this dataset, the configuration \(w_d = 0.4\) and \(w_s = 0.6\) achieved the best overall performance.

\begin{table}[htbp]
\centering
\begin{tabular}{cc cccc}
\toprule
\multicolumn{2}{c}{Weights} & 0.1\,m & 0.5\,m & 1\,m & 1\,m 30$^\circ$ \\
{\footnotesize \(w_d\)} & {\footnotesize \(w_s\)} & & & & \\
\midrule
1.0&0   & 0.83   & 8.95   & 14.45 & 11.85 \\
0.9&0.1 & 1.13   & 13.14  & 20.53 & 18.07 \\
0.8&0.2 & 1.28   & 15.21  & 23.57 & 20.96 \\
0.7&0.3 & 1.53   & 16.61  & 25.69 & 22.90 \\
0.6&0.4 & \textbf{1.56} & 16.88  & 26.07 & 23.58 \\
0.5&0.5 & 1.51   & 16.74  & 26.37 & 23.31 \\
0.4&0.6 & 1.38   & \textbf{16.90} & \textbf{26.86} & \textbf{23.87} \\
0.3&0.7 & 1.31   & 16.38  & 26.39 & 23.67 \\
0.1&0.9 & 1.22   & 16.16  & 25.81 & 22.97 \\
0&1.0   & 0.04   & 1.83   & 5.25  & 3.04  \\
\bottomrule
\end{tabular}
\caption{Recall metrics on the ZInD validation set obtained with our model without room aware and refinement.}
\label{tab:zind_model}
\end{table}
\subsection{Room Type Classification Results}
In this section, we evaluate the performance of our room type prediction branch on two datasets: S3D (\ref{S3D Dataset room prediction}) and ZInD (\ref{zind Dataset room prediction}). Accurate room type classification not only provides semantic context for localization but also reduces the effective search space for image matching.

\subsubsection{Room Type - S3D}
\label{S3D Dataset room prediction}
\begin{figure}[htbp]
    \centering
    \includegraphics[width=\linewidth]{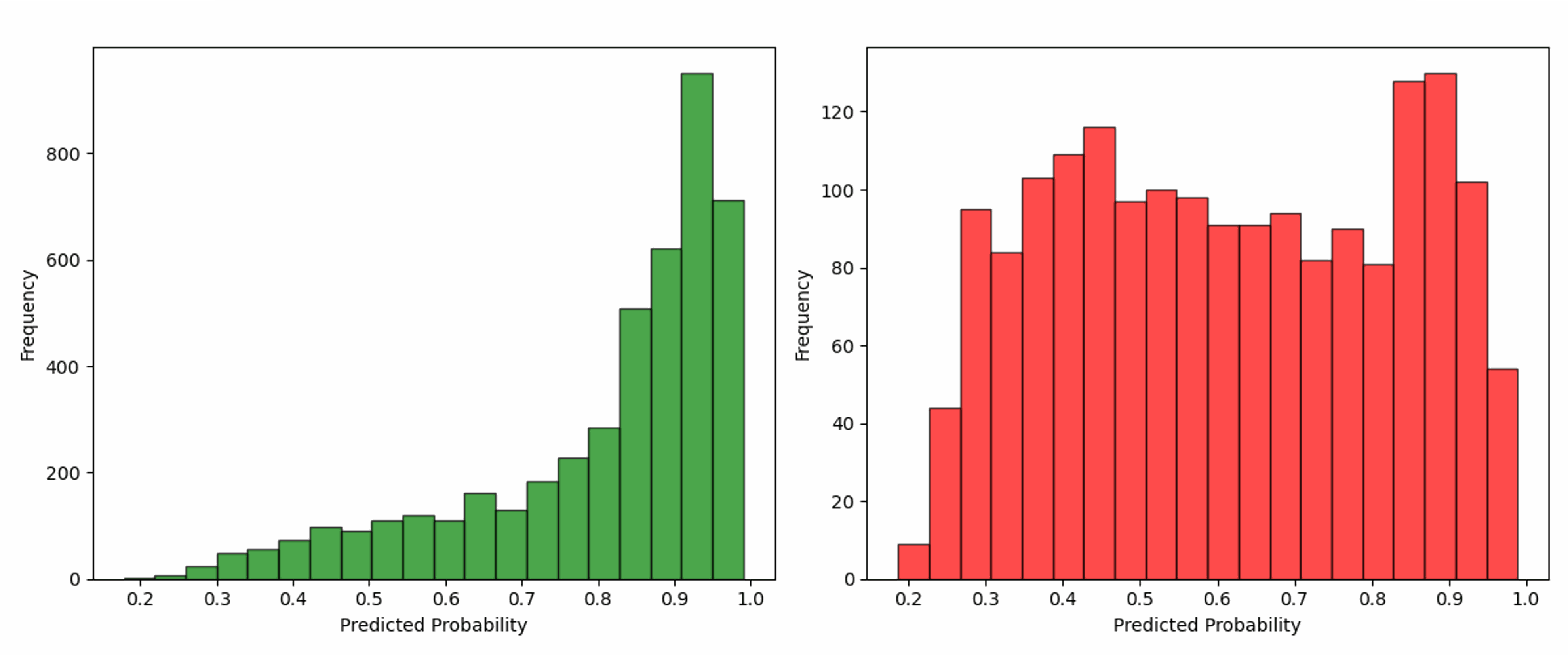}
    \caption{Room type prediction branch confidence scores over the S3D dataset. Correct predictions (\textcolor{green}{green}, left side) show high confidence, while incorrect predictions (\textcolor{red}{red}, right side) are more uniformly distributed.}
    \label{fig:room_confidence}
\end{figure}

On the S3D dataset, which consists of fully furnished environments, our model achieves a room type prediction accuracy of 72.1\%. A major source of misclassifications stems from uninformative images and rooms lacking furniture, which are common in the dataset. As shown in Figure~\ref{fig:room_confidence}, correct predictions generally exhibit high confidence scores (greater than 0.8), whereas misclassifications tend to display a more uniform confidence distribution across incorrect labels. Based on these observations, we set our threshold \(T_{\text{room}} = 0.8\): any prediction with a confidence score lower than 0.8 is rejected. This strategy limits misclassifications and effectively narrows the search space, resulting in an average improvement of 6.2\% across the 0.5m, 1m, and 1m 30$^\circ$ thresholds, as seen from the gap between $\text{Ours}_s$ and $\text{Ours}_r$. Notably, on the 1m 30$^\circ$ metric, the improvement is 3.74 percentage points.

\begin{figure}[htbp]
    \centering
    \includegraphics[width=\linewidth]{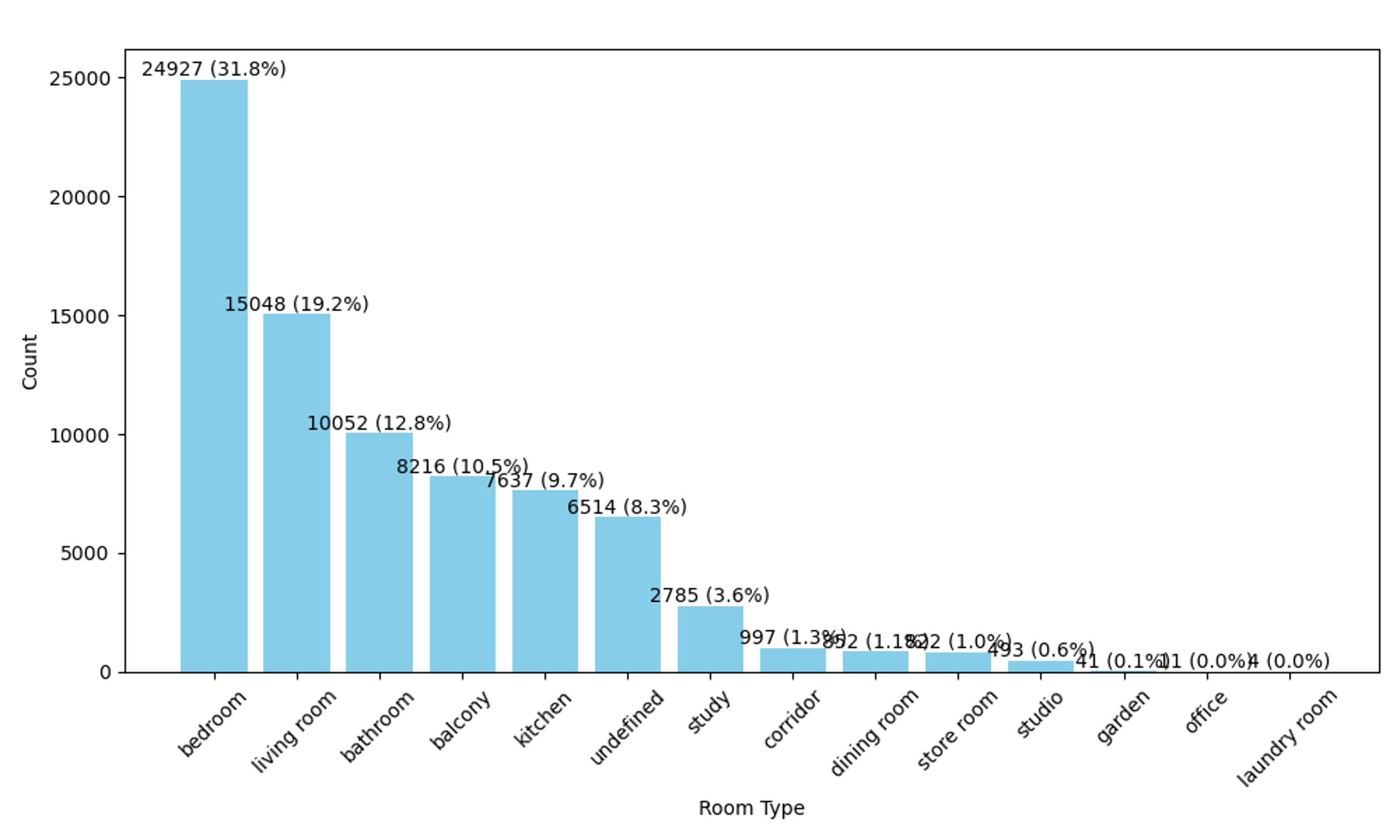}
    \caption{Overall room type distribution in the S3D dataset. Each column indicates the total number of rooms with the corresponding label and their percentage out of all rooms.}
    \label{fig:room_distribution}
\end{figure}

Figure~\ref{fig:room_distribution} illustrates the overall room type distribution in the S3D dataset. Notably, bedrooms dominate the dataset, with an average of three per floorplan. Although this narrows the search space, it does not isolate a single room type. Furthermore, our analysis reveals that the areas corresponding to room labels account for only 27.6\% of the total apartment area. This means that, on average, if true room labels were available, the effective area to be searched would be reduced to just 27.6\% of the full apartment, significantly narrowing the search space for image localization.

\subsubsection{Room Type - ZInD}
\label{zind Dataset room prediction}
For the ZInD dataset, the prediction accuracy drops significantly to 45\%. This lower accuracy can be attributed to the unfurnished nature of the dataset, which results in many ambiguous room images, and to the large number (over 250) and inconsistency of room labels. To address these issues, we grouped similar labels (e.g., “bedroom -1”, “primary bedroom”, “main bedroom”) into a single category. After grouping, we selected the top 15 room labels and classified all remaining labels as \textit{undefined} (thereby excluding sparse categories). Although the gain from incorporating room predictions on the 1m 30° metric in ZInD is 2.11 percentage points, lower than that observed in S3D, it still constitutes a significant enhancement in narrowing the search space for image localization.

\subsection{Effects of Refinement Parameter Choices}
\label{sec:runtime}
In Table \ref{tab:refinemnet_parameters_table} we present an experiment on the S3D validation set, comparing baseline methods with refinement results across various configurations. From this table, we selected the best score and used its corresponding parameters for evaluation on our test set.
\begin{table}[ht]
\centering
\resizebox{\columnwidth}{!}{%
\small
\begin{tabular}{lccccccc}
\hline
Method & dist & alpha & Top-K & R@0.1m & R@0.5m & R@1m & R@1\,m 30$^\circ$ \\
\hline
Baseline & 0.1 & 0.1 & 3 & \textbf{0.055} & 0.417 & 0.545 & 0.533 \\
Refine   & 0.1 & 0.1 & 3 & 0.049 & \textbf{0.432} & \textbf{0.566} & \textbf{0.553} \\
\hline
Baseline & 0.1 & 0.1 & 5 & \textbf{0.054} & 0.416 & 0.547 & 0.535 \\
Refine   & 0.1 & 0.1 & 5 & 0.047 & \textbf{0.426} & \textbf{0.564} & \textbf{0.553} \\
\hline
Baseline & 0.1 & 0.3 & 3 & \textbf{0.054} & 0.420 & 0.548 & 0.537 \\
Refine   & 0.1 & 0.3 & 3 & 0.049 & \textbf{0.437} & \textbf{0.570} & \textbf{0.557} \\
\hline
Baseline & 0.1 & 0.3 & 5 & 0.050 & 0.409 & 0.539 & 0.528 \\
Refine   & \textbf{0.1} & \textbf{0.3} & \textbf{5} & \textbf{0.052} & \textbf{0.428} & \textbf{0.563} & \textbf{0.551} \\
\hline
Baseline & 0.1 & 0.5 & 3 & 0.050 & 0.413 & 0.539 & 0.528 \\
Refine   & 0.1 & 0.5 & 3 & \textbf{0.051} & \textbf{0.430} & \textbf{0.563} & \textbf{0.551} \\
\hline
Baseline & 0.1 & 0.5 & 5 & \textbf{0.053} & 0.417 & 0.544 & 0.532 \\
Refine   & 0.1 & 0.5 & 5 & 0.052 & \textbf{0.435} & \textbf{0.564} & \textbf{0.553} \\
\hline
Baseline & 0.5 & 0.1 & 3 & \textbf{0.054} & \textbf{0.413} & 0.541 & \textbf{0.530} \\
Refine   & 0.5 & 0.1 & 3 & 0.046 & 0.377 & \textbf{0.542} & 0.528 \\
\hline
Baseline & 0.5 & 0.1 & 5 & \textbf{0.055} & \textbf{0.413} & \textbf{0.538} & \textbf{0.526} \\
Refine   & 0.5 & 0.1 & 5 & 0.042 & 0.350 & 0.527 & 0.513 \\
\hline
Baseline & 0.5 & 0.3 & 3 & \textbf{0.054} & \textbf{0.420} & 0.547 & 0.536 \\
Refine   & 0.5 & 0.3 & 3 & 0.053 & 0.392 & \textbf{0.552} & \textbf{0.539} \\
\hline
Baseline & 0.5 & 0.3 & 5 & \textbf{0.055} & \textbf{0.417} & \textbf{0.543} & \textbf{0.531} \\
Refine   & 0.5 & 0.3 & 5 & 0.046 & 0.372 & 0.535 & 0.522 \\
\hline
Baseline & 0.5 & 0.5 & 3 & \textbf{0.053} & \textbf{0.414} & 0.546 & 0.533 \\
Refine   & 0.5 & 0.5 & 3 & 0.050 & 0.388 & \textbf{0.546} & 0.533 \\
\hline
Baseline & 0.5 & 0.5 & 5 & \textbf{0.051} & \textbf{0.412} & \textbf{0.540} & \textbf{0.529} \\
Refine   & 0.5 & 0.5 & 5 & 0.048 & 0.370 & 0.536 & 0.523 \\
\hline
Baseline & 1.0 & 0.1 & 3 & \textbf{0.054} & \textbf{0.420} & \textbf{0.552} & \textbf{0.540} \\
Refine   & 1.0 & 0.1 & 3 & 0.050 & 0.358 & 0.496 & 0.484 \\
\hline
Baseline & 1.0 & 0.1 & 5 & \textbf{0.052} & \textbf{0.413} & \textbf{0.541} & \textbf{0.530} \\
Refine   & 1.0 & 0.1 & 5 & 0.042 & 0.321 & 0.455 & 0.444 \\
\hline
Baseline & 1.0 & 0.3 & 3 & \textbf{0.054} & \textbf{0.414} & \textbf{0.542} & \textbf{0.531} \\
Refine   & 1.0 & 0.3 & 3 & 0.053 & 0.382 & 0.516 & 0.504 \\
\hline
Baseline & 1.0 & 0.3 & 5 & 0.052 & \textbf{0.412} & \textbf{0.543} & \textbf{0.531} \\
Refine   & 1.0 & 0.3 & 5 & \textbf{0.053} & 0.369 & 0.504 & 0.492 \\
\hline
\end{tabular}%
}
\caption{Refinement parameter experiment on the S3D validation set.}
\label{tab:refinemnet_parameters_table}
\end{table}
\subsection{Additional Comparison with LASER}
Table~\ref{tab:compare_to_laser} compares our approach with the LASER baseline. To ensure a fair evaluation, we train our model under the same protocol as LASER, applying random yaw perturbations to the panoramas during training. We then evaluate both methods on the test set—using the same random yaw sampling—and report the mean recall over five independent runs. Our method significantly outperforms LASER at the 1\,m and 1\,m\,30$^\circ$ thresholds; in particular, we achieve a 64\% absolute improvement on the 1\,m\,30$^\circ$ metric, which is the most critical measure for our application. LASER, however, attains higher recall on the fine localization metrics (0.1\,m), suggesting that given a large training set, their model can achieve finer-grained accuracy.
We observe that the scores for the dataset when randomly cropping panoramas are lower than those for the perspective sets.  
Two factors contribute to this gap: (i)~under random-yaw training, a larger fraction of panorama crops contain uninformative wall-only views, making localization harder. And (ii) ~in the S3D dataset the resolution of a cropped panorama view is much lower than that of an image covering the same field of view in the perspective dataset—\textit{e.g.}, approximately $228\times512$\,px versus $1280\times720$\,px.  
Both the reduced visual content and the lower image quality adversely affect model performance on the panorama random yaw crop.
With these results, we consider the LASER baseline to be faithfully reproduced. 

\label{sec:runtime}
\begin{table}[htbp]
\centering
\small
\begin{tabular}{ccccc}
\toprule
Method & 0.1\,m & 0.5\,m & 1\,m & 1\,m 30$^\circ$ \\
\midrule
LASER & \textbf{6.48}  & 25.75  & 31.05 & 22.57 \\
$\text{Ours}_s$ & 3.12  & 23.84  & 32.34  & 29.52 \\
$\text{Ours}_r$ & 4.33  & \textbf{31.12 } & \textbf{42.49 } & \textbf{37.13} \\
\bottomrule
\end{tabular}
\caption{Recall metrics on the S3D dataset, with a random yaw in the training stage. Results are reported on the random angle of yaw of each panorama in the test set and averaged over N = 5 times.}
\label{tab:compare_to_laser}
\end{table}
\subsection{Comparison against Soft Constraints}
Our approach uses hard thresholds both for semantic ray classification—where we assign each ray the class with the highest probability—and for room‐type selection—where we apply a binary mask for the room with the maximum confidence. To validate this hard‐threshold strategy against a soft‐constraint alternative, we conduct two experiments: 
(1) \textbf{Semantic Ray Classification} compares hard vs.\ soft ray assignments, and  
(2) \textbf{Room‐Type Selection} compares hard vs.\ soft room‐type classifications.  
Results are reported in Table~\ref{tab:hard_vs_soft_exp1} and Table~\ref{tab:hard_vs_soft_exp2}

\subsubsection{Semantic Ray Classification}
For semantic ray classification, instead of selecting the highest-probability class for each ray (hard assignment), we retained the logits and computed a probability map by measuring cross-entropy with the ground truth (soft assignment). This soft approach caused a dramatic decrease in all recall metrics (e.g., R@1\,m\,30$^\circ$ dropped to 25.88\% on S3D), demonstrating that hard assignments are crucial for aggregating semantic information in our network.

\begin{table}[htbp]
\centering
\small
\begin{tabular}{lcccc}
\toprule
Method & 0.1\,m & 0.5\,m & 1\,m & 1\,m\,30$^\circ$ \\
\midrule
Hard $\text{Ours}_s$     &  5.42 & 41.87 & 53.52 & 52.61 \\
Hard $\text{Ours}_r$    & \textbf{5.70} & \textbf{45.53} & \textbf{58.78} & \textbf{57.49} \\
Soft $\text{Ours}_s$     &  1.94 & 16.74 & 26.22 & 22.66 \\
Soft $\text{Ours}_r$     &  2.24 & 19.55 & 31.43 & 25.88 \\
\bottomrule
\end{tabular}
\caption{Recall metrics on the S3D dataset for semantic ray classification under hard vs.~soft assignments (Experiment~1). The top result in each column is \textbf{bolded}.}
\label{tab:hard_vs_soft_exp1}
\end{table}

\subsubsection{Room-Type Selection}
For room-type selection, we compared a hard classification approach—where each room polygon receives a binary mask from the maximum-probability prediction—to a soft classification approach, in which each room polygon is weighted by its predicted probability. Although the gap is modest, hard classification still outperforms the soft approach (e.g., 1.23\% gap in R@1\,m\,30$^\circ$ on S3D).

\begin{table}[htbp]
\centering
\small
\begin{tabular}{lcccc}
\toprule
Method & 0.1\,m & 0.5\,m & 1\,m & 1\,m\,30$^\circ$ \\
\midrule
Hard $\text{Ours}_s$      &  5.70 & 45.53 & 58.78 & 57.49 \\
Soft $\text{Ours}_r$     &  4.55 & 43.57 & 58.51 & 56.27 \\
\bottomrule
\end{tabular}
\caption{Recall metrics on the S3D dataset for room-type selection under hard vs.~soft classification (Experiment~2). The top result in each column is \textbf{bolded}.}
\label{tab:hard_vs_soft_exp2}
\end{table}

\section{Additional Visualizations}
\label{sec:qualitative}

\subsection{Qualitative Visualizations}
\label{sec:visualizations}
\begin{figure*}[htbp]
    \centering
    \centering
\renewcommand{\arraystretch}{1.2}
\begin{tabular}{%
  >{\centering\arraybackslash}m{0.25\textwidth}%
  >{\centering\arraybackslash}m{0.15\textwidth}%
  >{\centering\arraybackslash}m{0.15\textwidth}%
  >{\centering\arraybackslash}m{0.15\textwidth}%
  >{\centering\arraybackslash}m{0.15\textwidth}%
}
\hline
\textbf{Input Image} & \textbf{Floorplan} & \textbf{Ours} & \textbf{Ground Truth Rays} & \textbf{Predicted Rays} \\
\hline
\noalign{\vskip 3pt}
\parbox[c]{\linewidth}{%
  \centering
  \includegraphics[width=\linewidth]{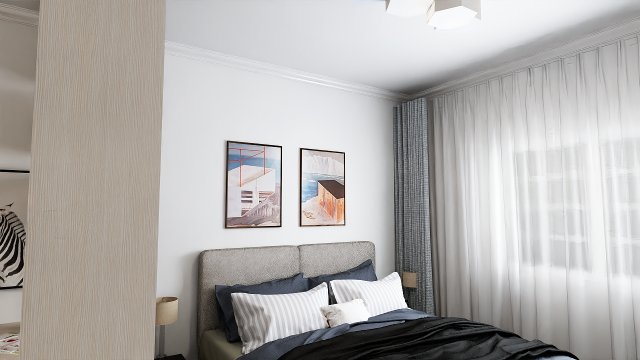}} &
\parbox[c]{\linewidth}{%
  \centering
  \includegraphics[width=\linewidth]{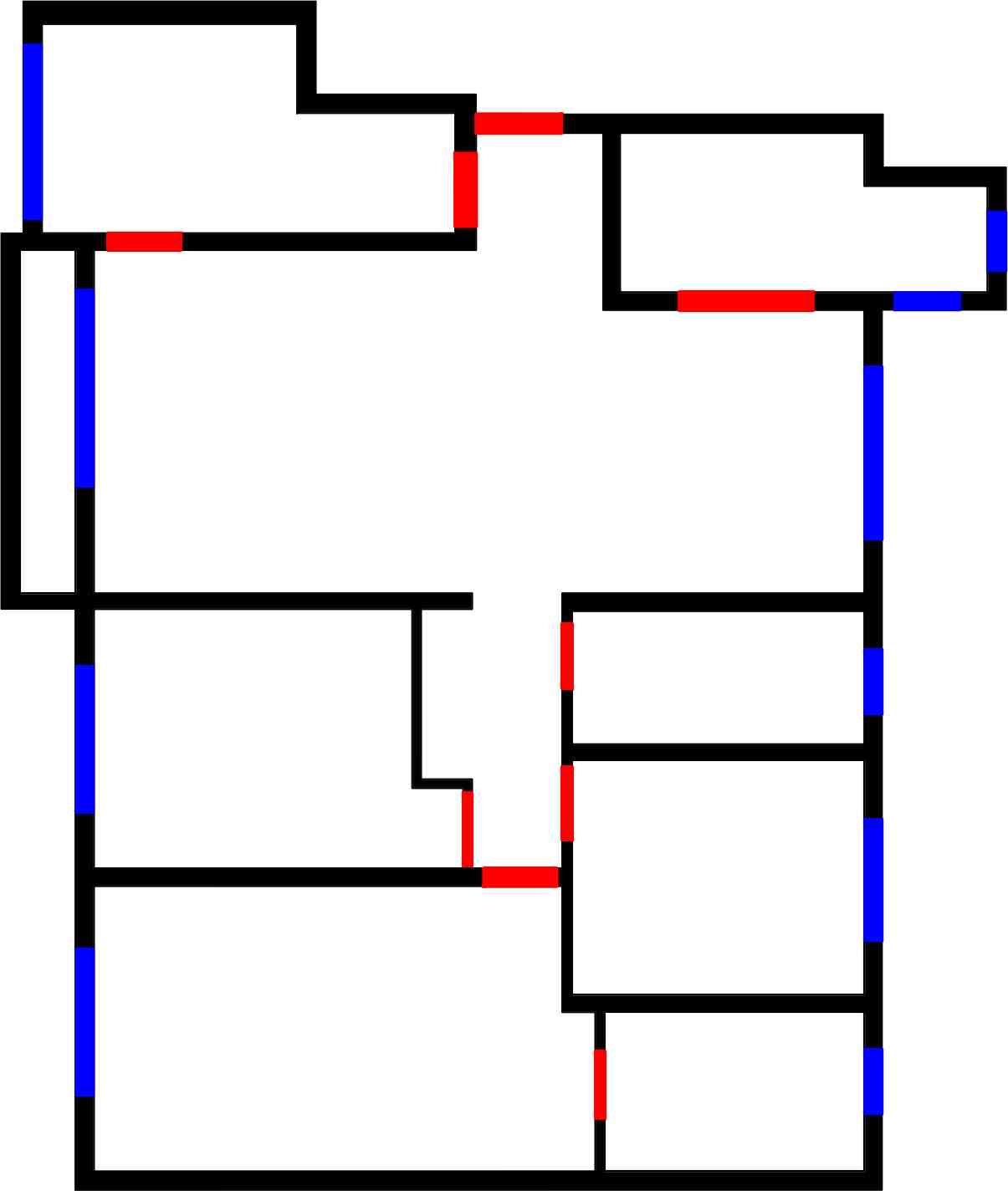}} &
\parbox[c]{\linewidth}{%
  \centering
  \includegraphics[width=\linewidth]{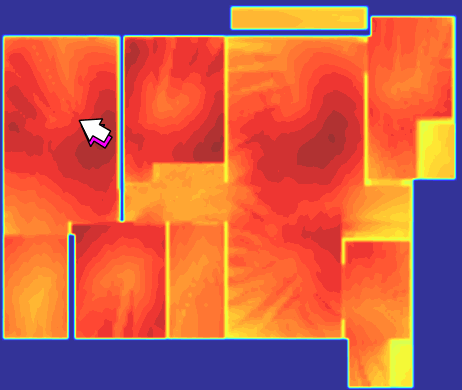}} &
\parbox[c]{\linewidth}{%
  \centering
  \includegraphics[width=\linewidth]{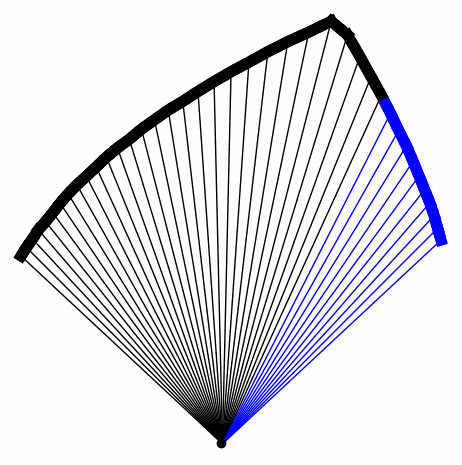}} &
\parbox[c]{\linewidth}{%
  \centering
  \includegraphics[width=\linewidth]{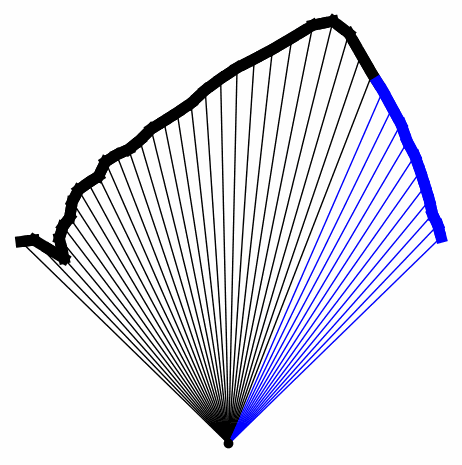}} \\
\noalign{\vskip 3pt}
\parbox[c]{\linewidth}{%
  \centering
  \includegraphics[width=\linewidth]{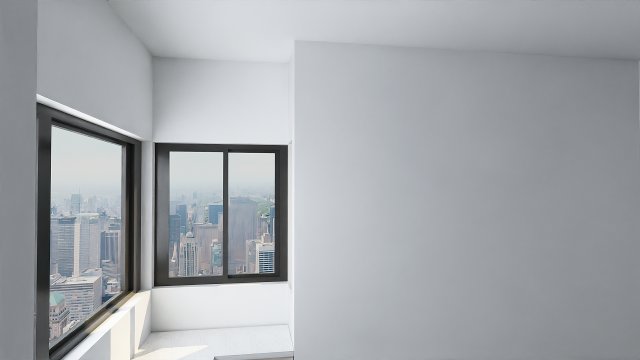}} &
\parbox[c]{\linewidth}{%
  \centering
  \includegraphics[width=\linewidth]{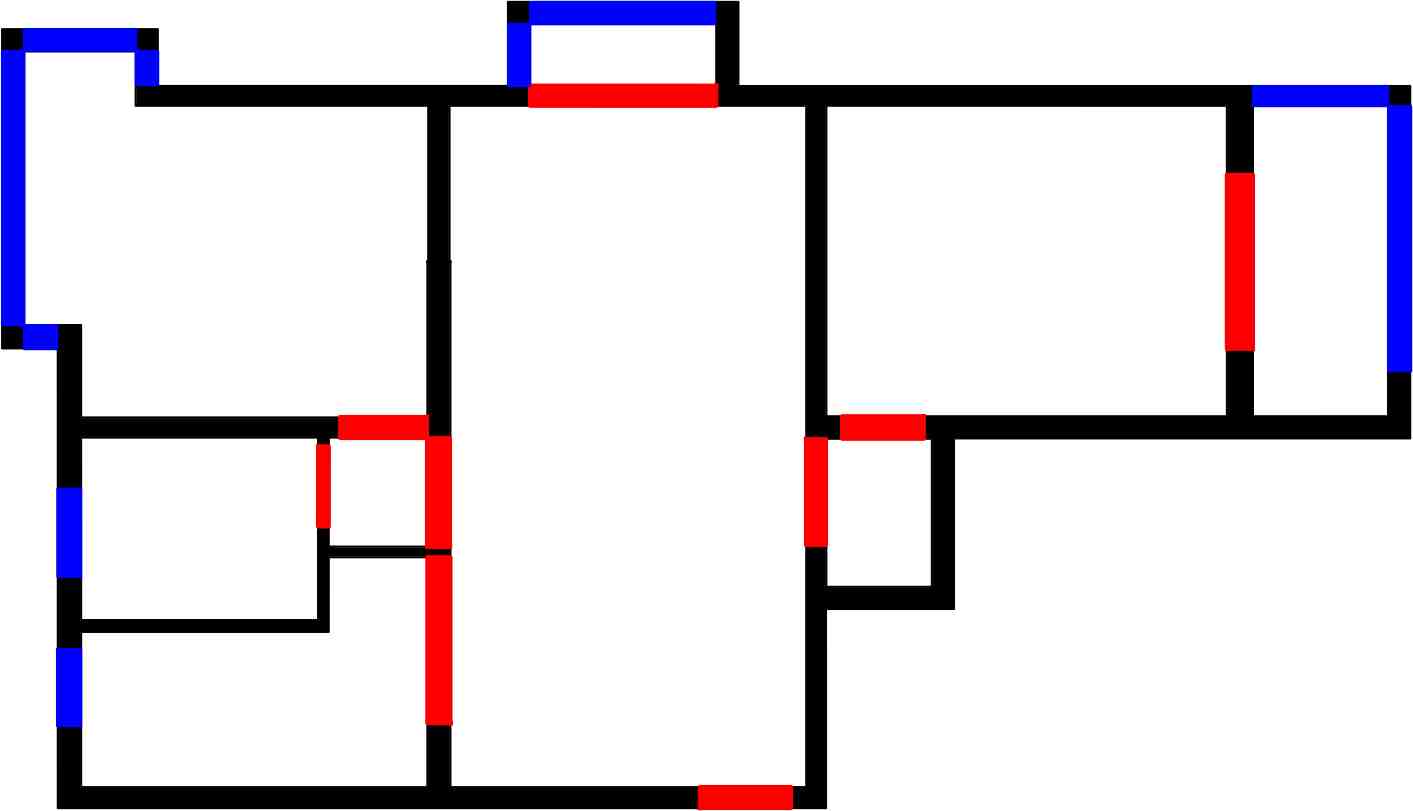}} &
\parbox[c]{\linewidth}{%
  \centering
  \includegraphics[width=\linewidth]{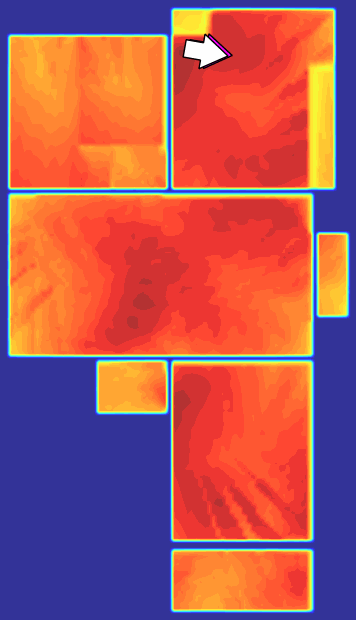}} &
\parbox[c]{\linewidth}{%
  \centering
  \includegraphics[width=\linewidth]{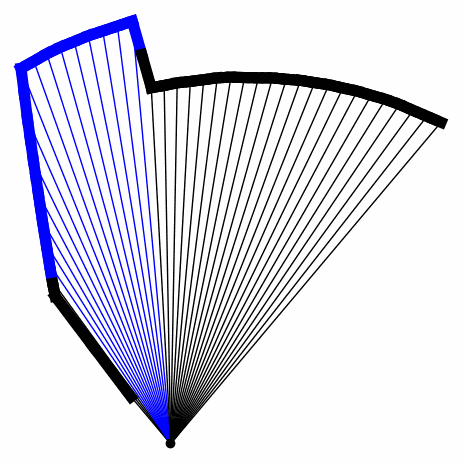}} &
\parbox[c]{\linewidth}{%
  \centering
  \includegraphics[width=\linewidth]{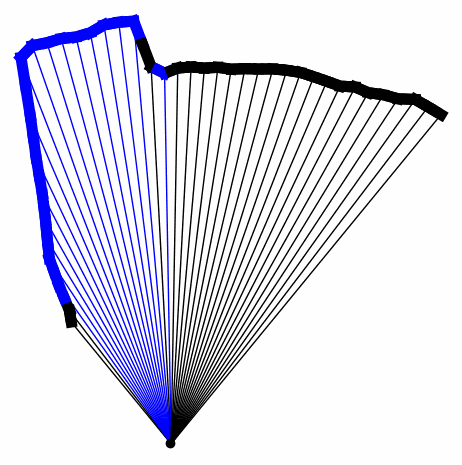}} \\
\noalign{\vskip 3pt}
\parbox[c]{\linewidth}{%
  \centering
  \includegraphics[width=\linewidth]{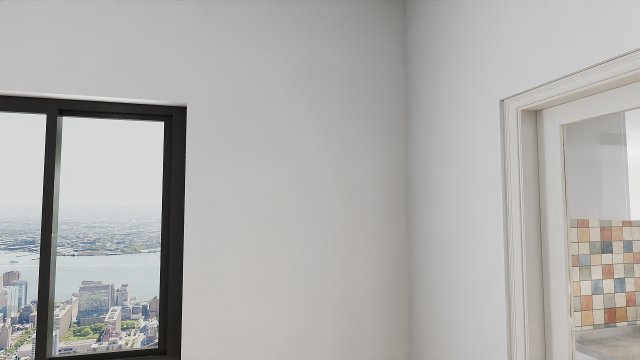}} &
\parbox[c]{\linewidth}{%
  \centering
  \includegraphics[width=\linewidth]{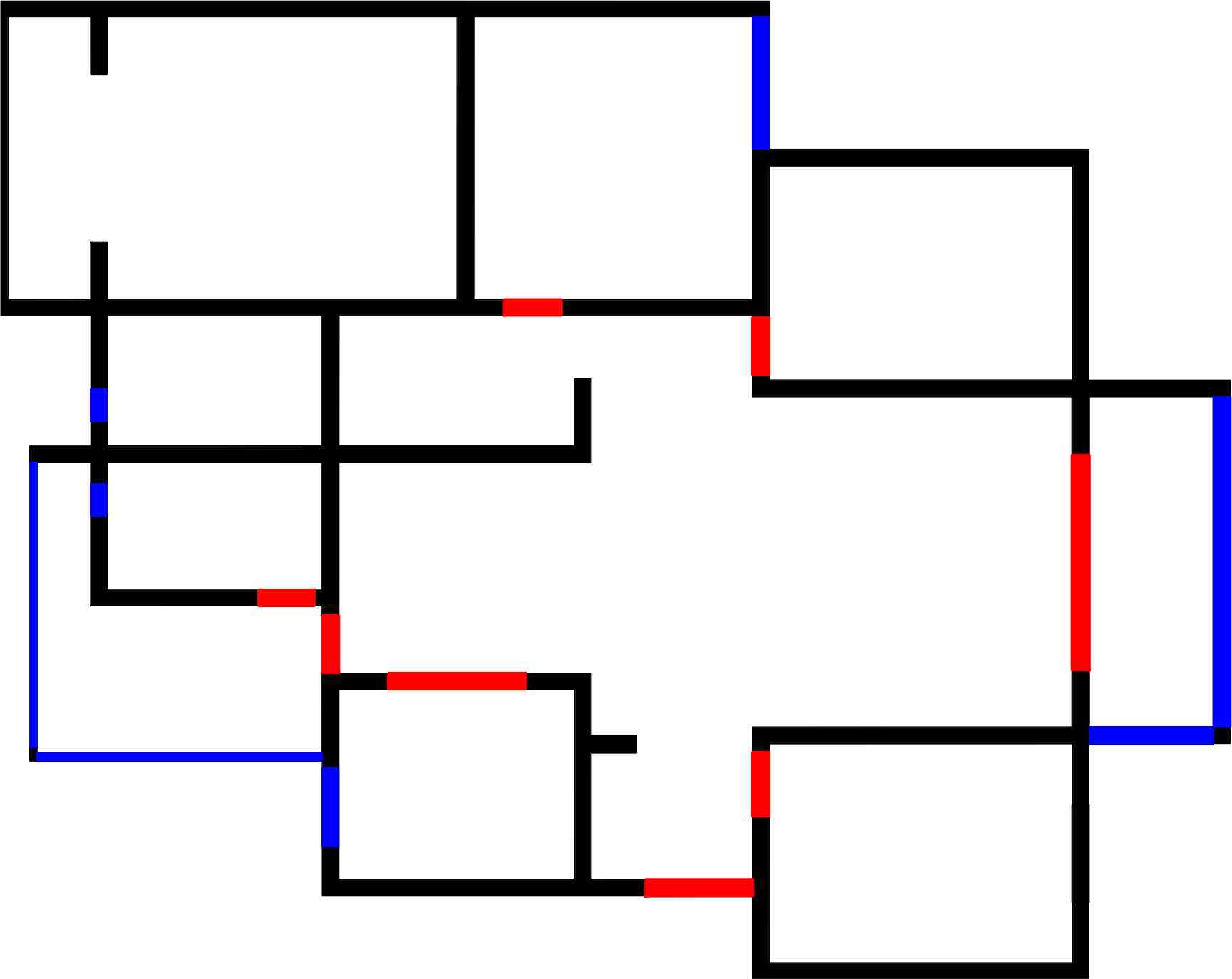}} &
\parbox[c]{\linewidth}{%
  \centering
  \includegraphics[width=\linewidth]{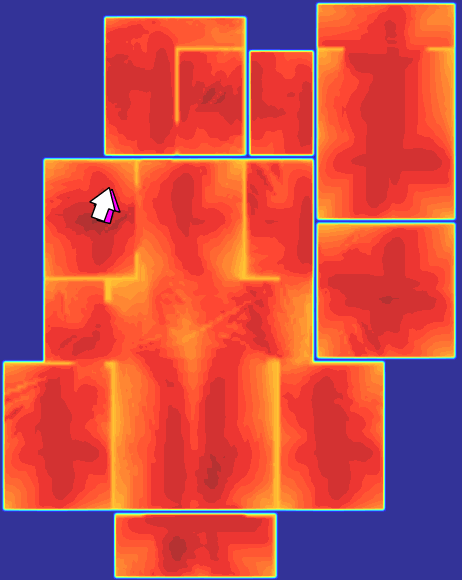}} &
\parbox[c]{\linewidth}{%
  \centering
  \includegraphics[width=\linewidth]{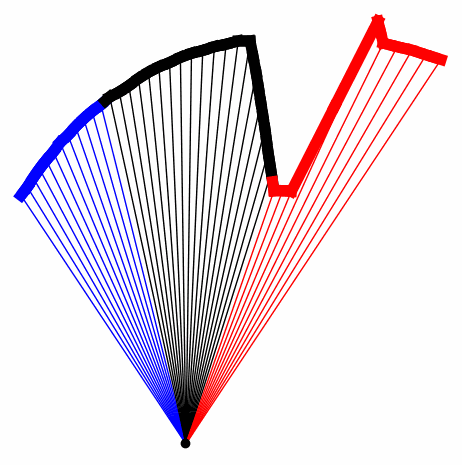}} &
\parbox[c]{\linewidth}{%
  \centering
  \includegraphics[width=\linewidth]{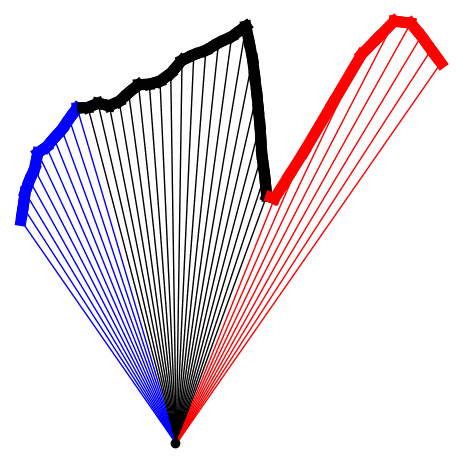}} \\
\noalign{\vskip 3pt}
\parbox[c]{\linewidth}{%
  \centering
  \includegraphics[width=\linewidth]{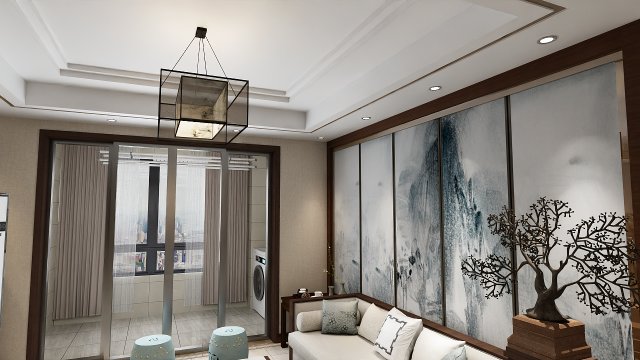}} &
\parbox[c]{\linewidth}{%
  \centering
  \includegraphics[width=\linewidth]{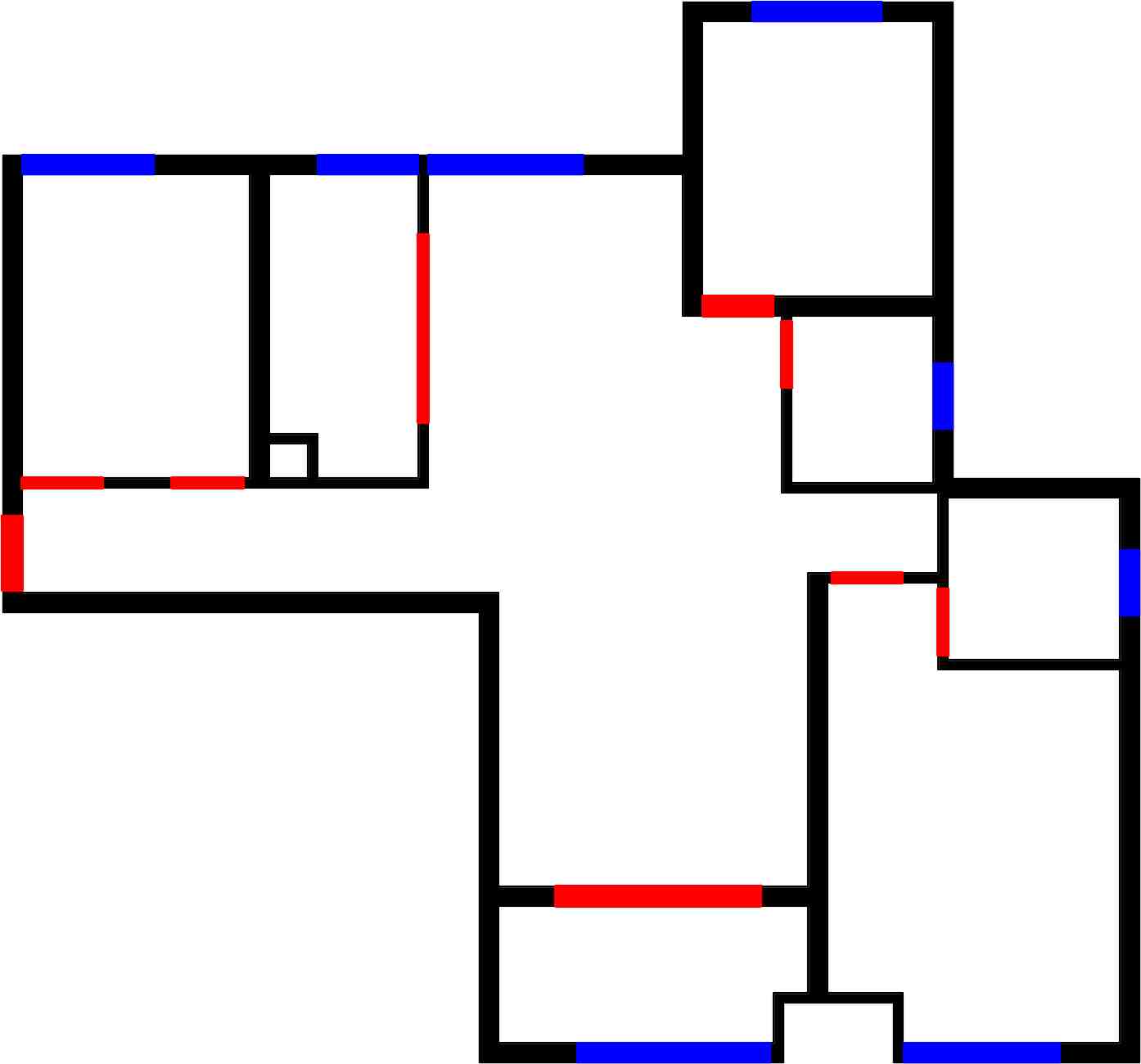}} &
\parbox[c]{\linewidth}{%
  \centering
  \includegraphics[width=\linewidth]{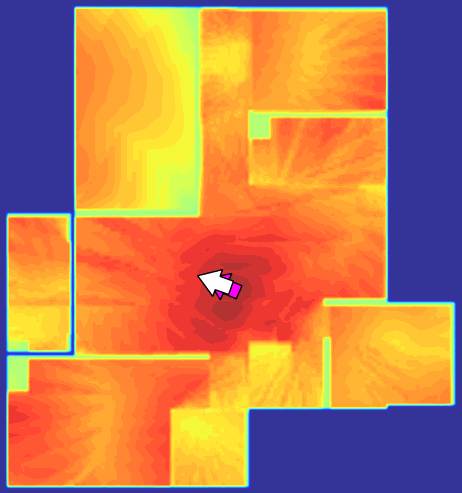}} &
\parbox[c]{\linewidth}{%
  \centering
  \includegraphics[width=\linewidth]{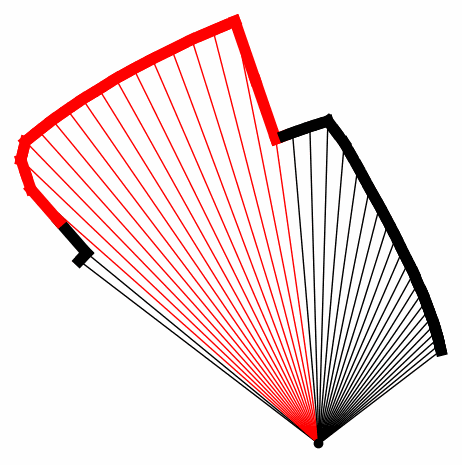}} &
\parbox[c]{\linewidth}{%
  \centering
  \includegraphics[width=\linewidth]{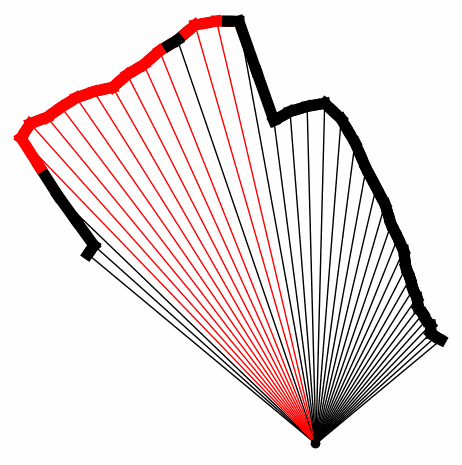}} \\
\noalign{\vskip 3pt}
\parbox[c]{\linewidth}{%
  \centering
  \includegraphics[width=\linewidth]{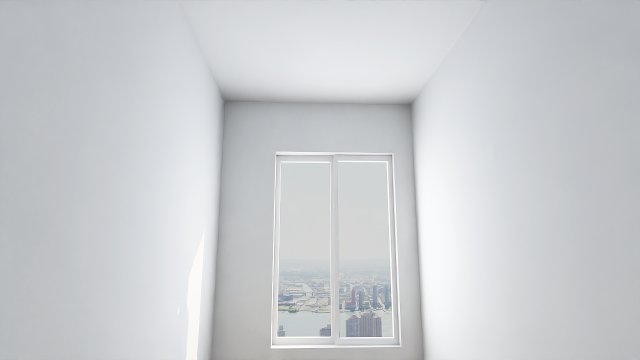}} &
\parbox[c]{\linewidth}{%
  \centering
  \includegraphics[width=\linewidth]{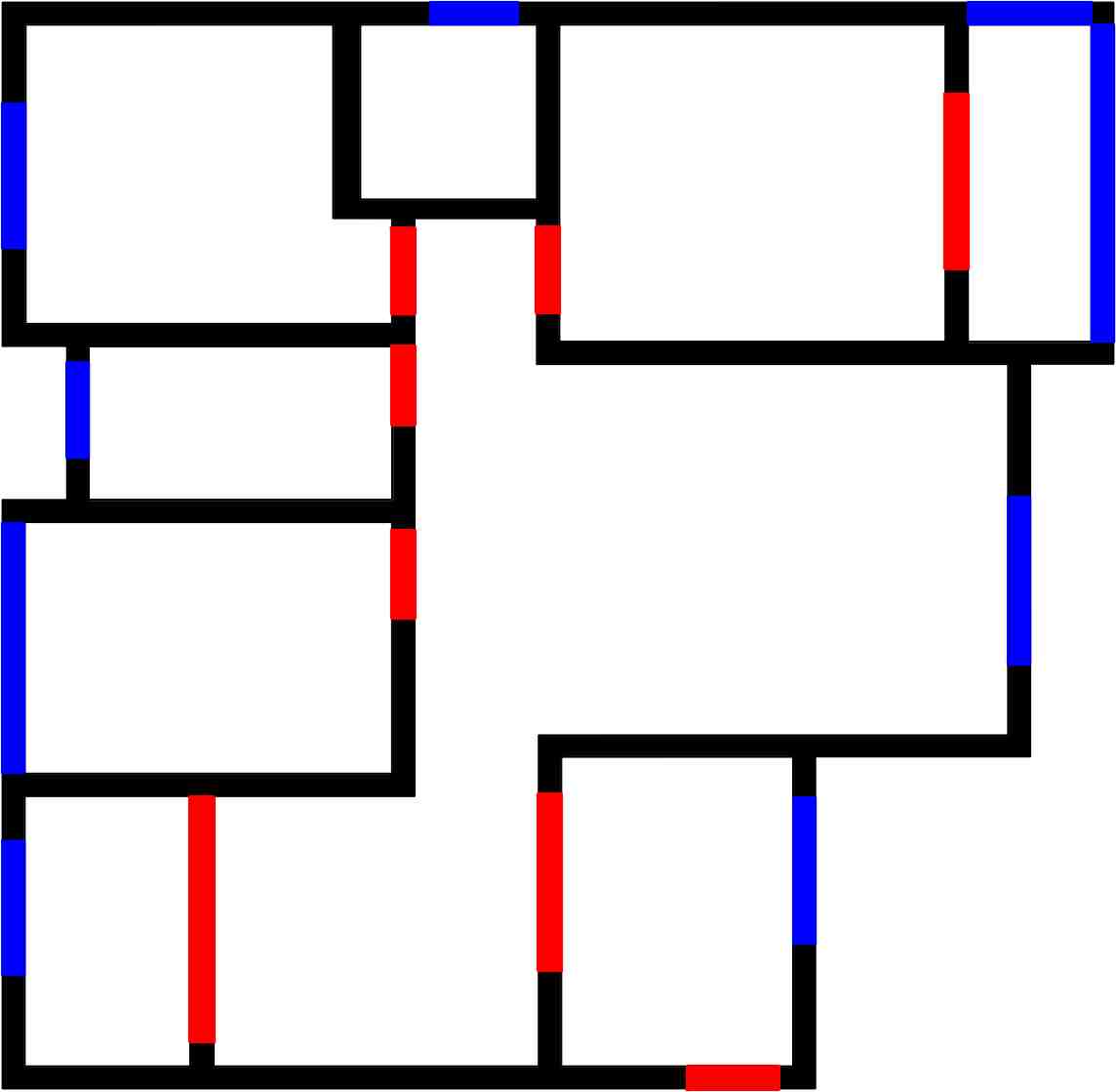}} &
\parbox[c]{\linewidth}{%
  \centering
  \includegraphics[width=\linewidth]{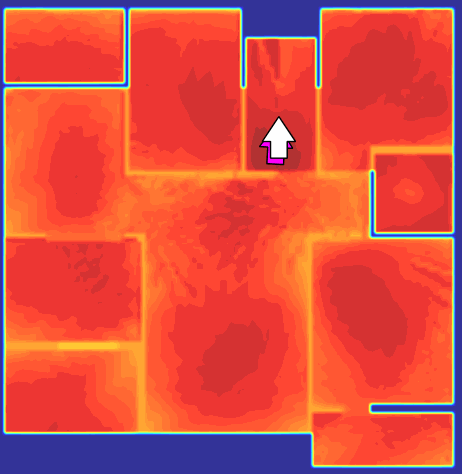}} &
\parbox[c]{\linewidth}{%
  \centering
  \includegraphics[width=\linewidth]{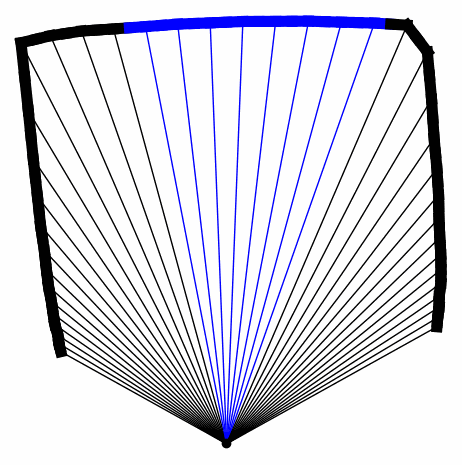}} &
\parbox[c]{\linewidth}{%
  \centering
  \includegraphics[width=\linewidth]{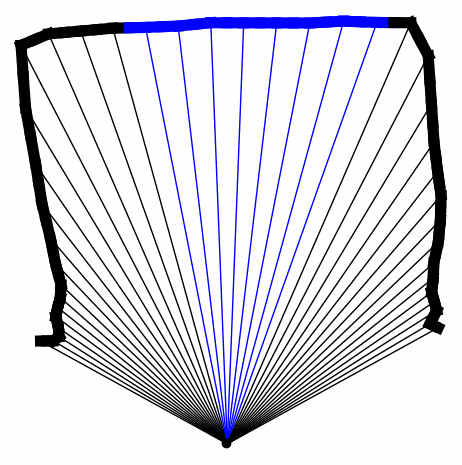}} \\
\parbox[c]{\linewidth}{%
  \centering
  \includegraphics[width=\linewidth]{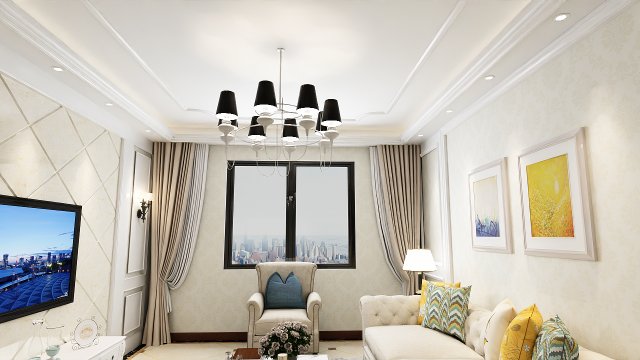}} &
\parbox[c]{\linewidth}{%
  \centering
  \includegraphics[width=\linewidth]{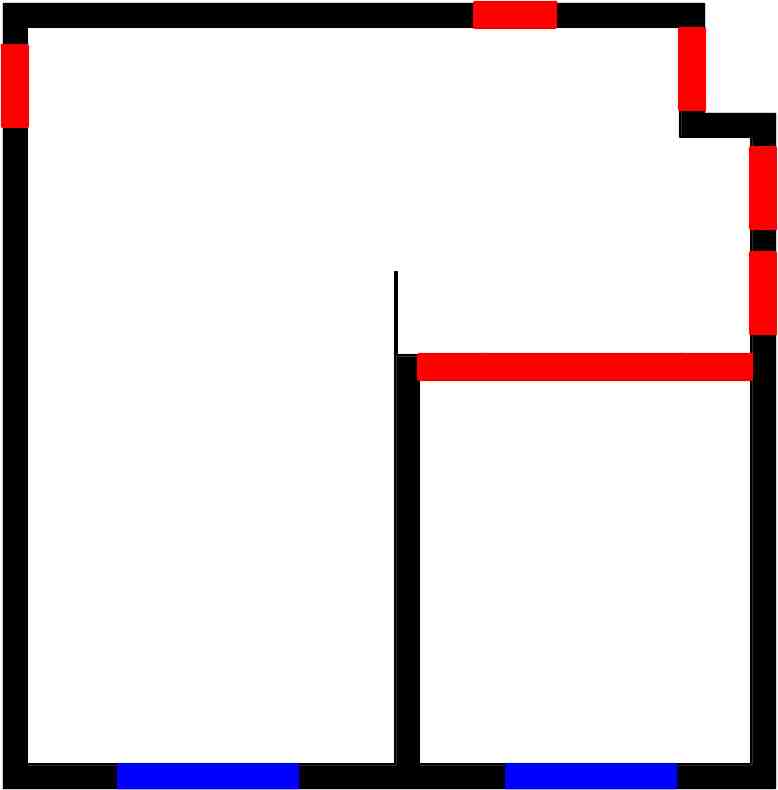}} &
\parbox[c]{\linewidth}{%
  \centering
  \includegraphics[width=\linewidth]{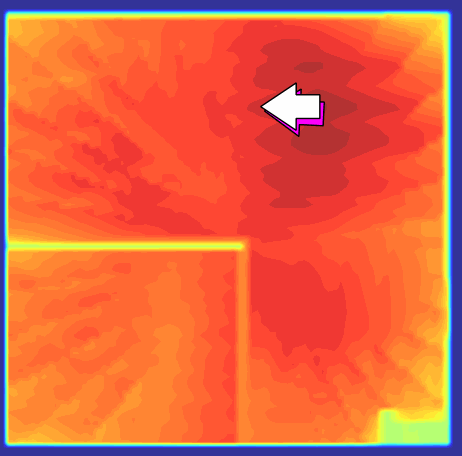}} &
\parbox[c]{\linewidth}{%
  \centering
  \includegraphics[width=\linewidth]{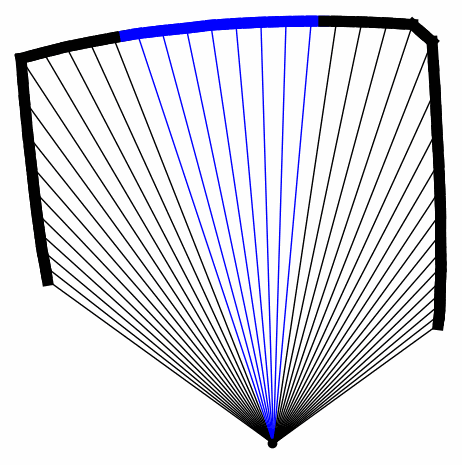}} &
\parbox[c]{\linewidth}{%
  \centering
  \includegraphics[width=\linewidth]{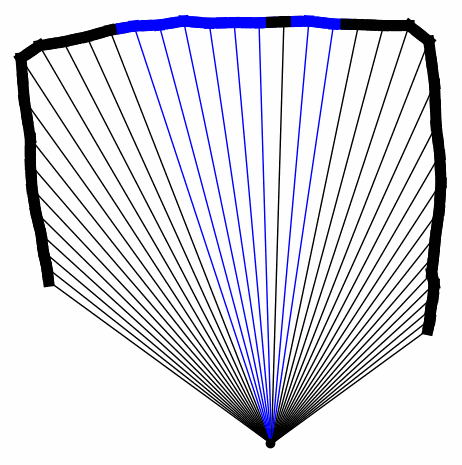}} \\
\noalign{\vskip 3pt}
\hline
\end{tabular}

    \caption{
    \textbf{Additional Qualitative Results (S3D dataset):}
    Warmer colors correspond to higher probabilities, while \textcolor{magenta}{magenta} indicates the ground-truth location and \textcolor{black}{white} denotes our predicted layout. Rays are: \textcolor{black}{wall}, \textcolor[HTML]{0000FF}{window}, and \textcolor[HTML]{FF0000}{door}.
    }
    \label{fig:good_examples_s3d}
\end{figure*}

\begin{figure*}[htbp]
    \centering
    \centering
\renewcommand{\arraystretch}{1.2}
\begin{tabular}{%
  >{\centering\arraybackslash}m{0.25\textwidth}%
  >{\centering\arraybackslash}m{0.15\textwidth}%
  >{\centering\arraybackslash}m{0.15\textwidth}%
  >{\centering\arraybackslash}m{0.15\textwidth}%
  >{\centering\arraybackslash}m{0.15\textwidth}%
}
\hline
\textbf{Input Image} & \textbf{Floorplan} & \textbf{Ours} & \textbf{Ground Truth Rays} & \textbf{Predicted Rays} \\
\hline
\noalign{\vskip 3pt}
\parbox[c]{\linewidth}{%
  \centering
  \includegraphics[width=\linewidth]{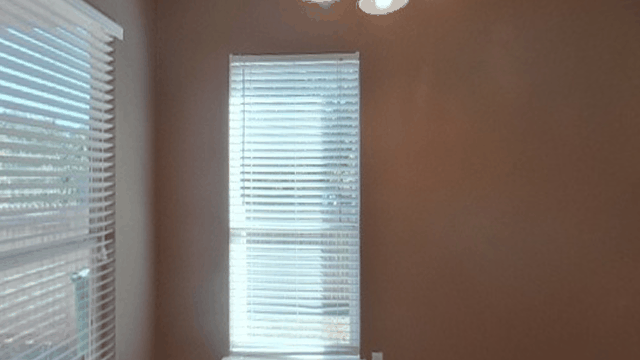}} &
\parbox[c]{\linewidth}{%
  \centering
  \includegraphics[width=\linewidth]{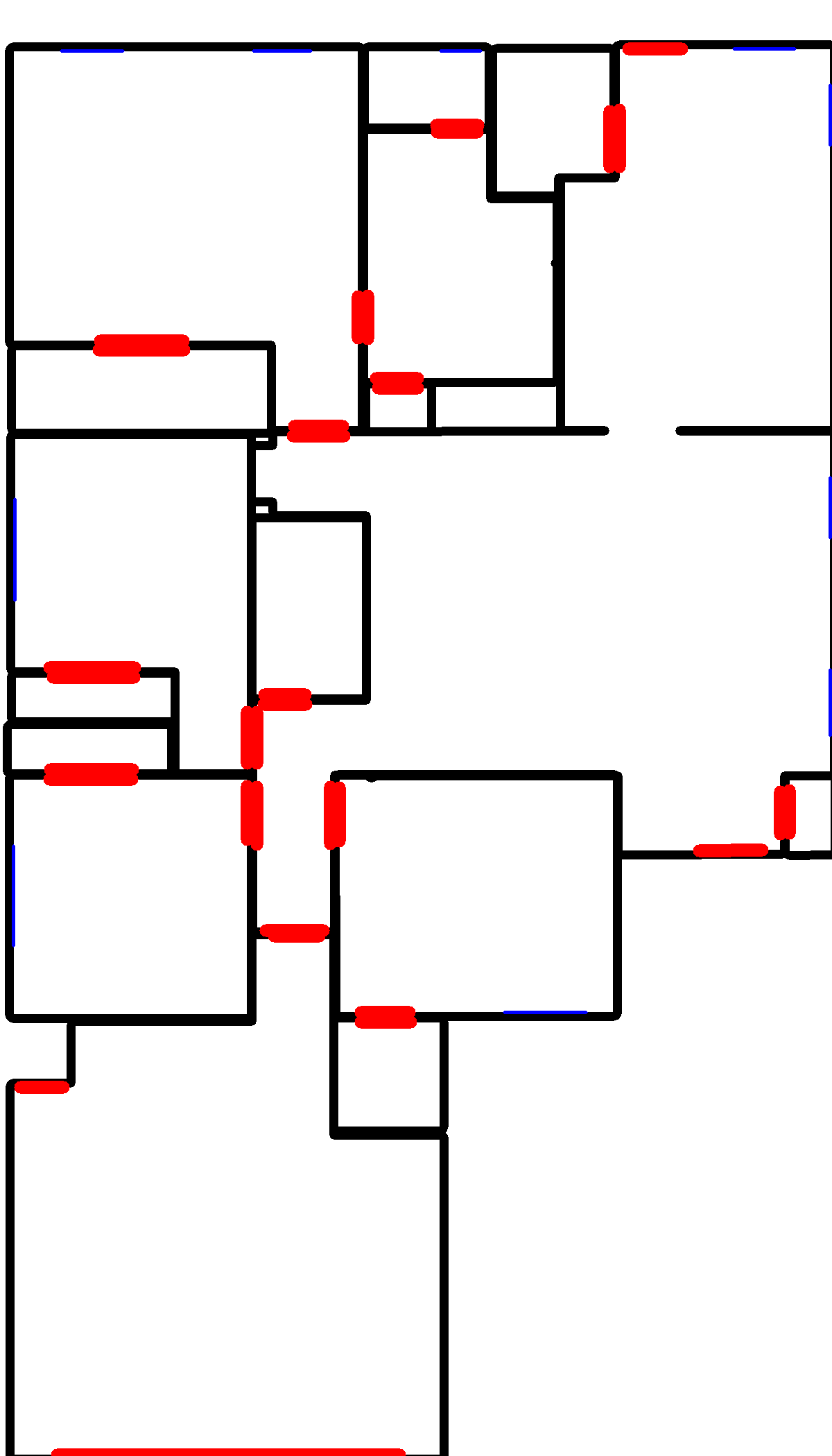}} &
\parbox[c]{\linewidth}{%
  \centering
  \includegraphics[width=\linewidth]{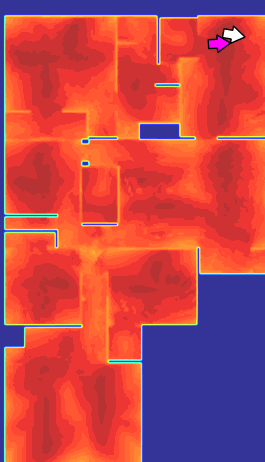}} &
\parbox[c]{\linewidth}{%
  \centering
  \includegraphics[width=\linewidth]{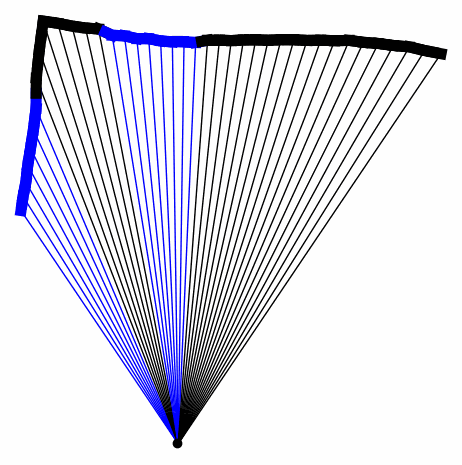}} &
\parbox[c]{\linewidth}{%
  \centering
  \includegraphics[width=\linewidth]{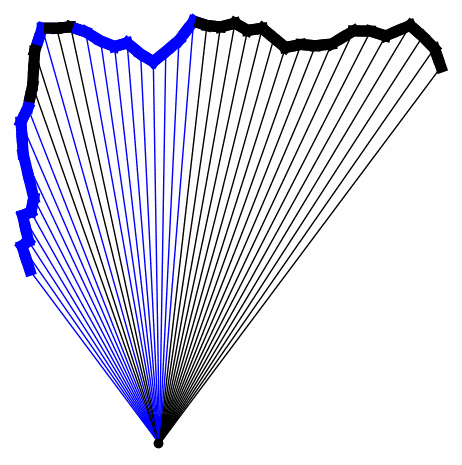}} \\
\noalign{\vskip 3pt}
\parbox[c]{\linewidth}{%
  \centering
  \includegraphics[width=\linewidth]{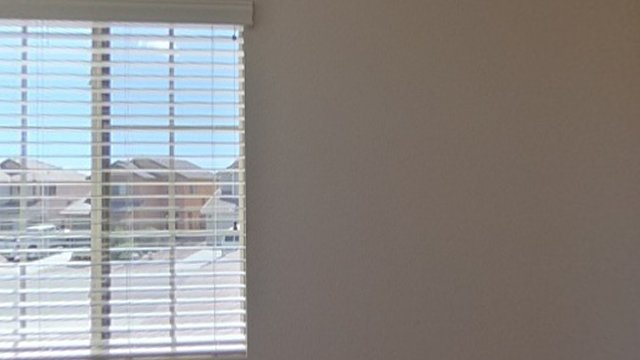}} &
\parbox[c]{\linewidth}{%
  \centering
  \includegraphics[width=\linewidth]{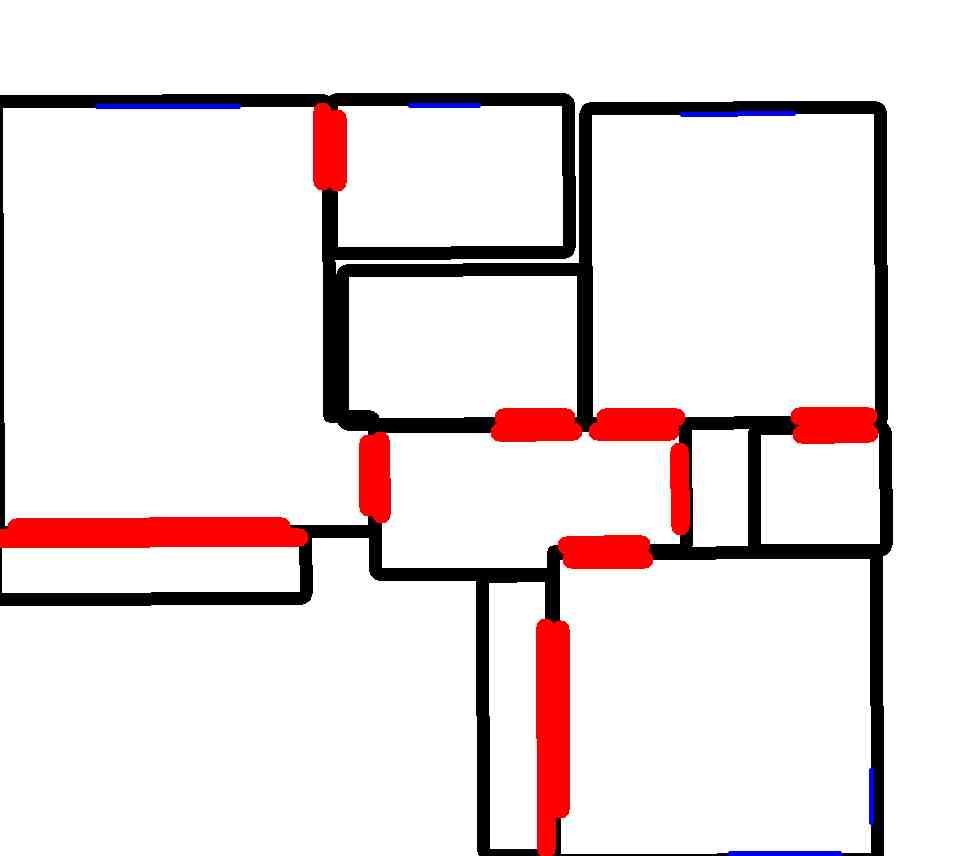}} &
\parbox[c]{\linewidth}{%
  \centering
  \includegraphics[width=\linewidth]{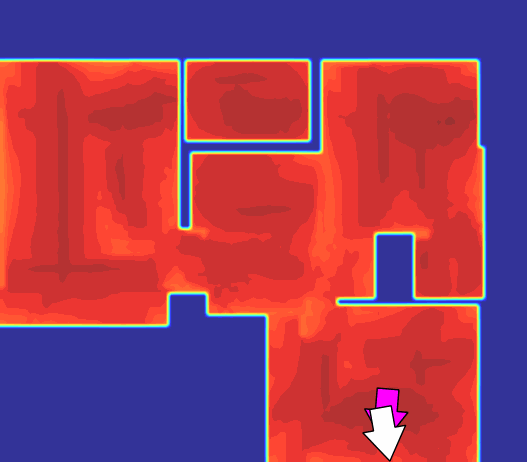}} &
\parbox[c]{\linewidth}{%
  \centering
  \includegraphics[width=\linewidth]{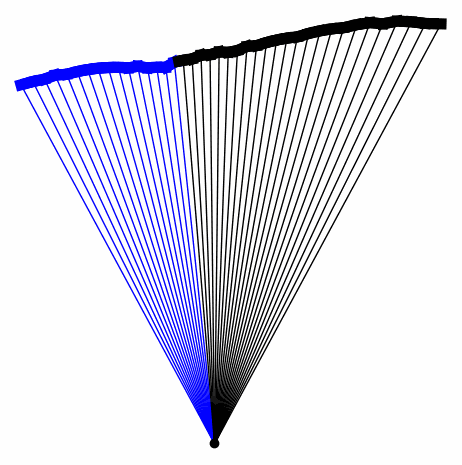}} &
\parbox[c]{\linewidth}{%
  \centering
  \includegraphics[width=\linewidth]{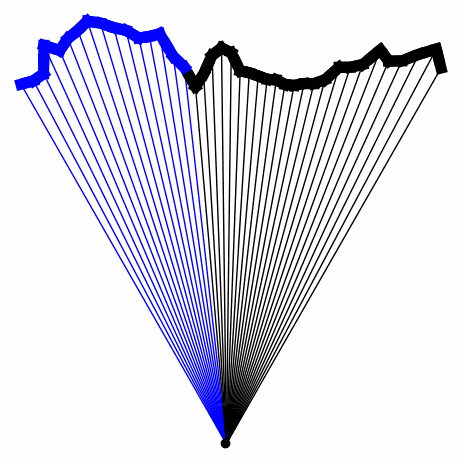}} \\
\noalign{\vskip 3pt}
\parbox[c]{\linewidth}{%
  \centering
  \includegraphics[width=\linewidth]{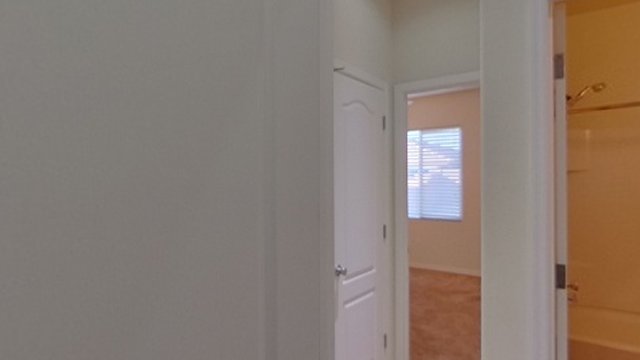}} &
\parbox[c]{\linewidth}{%
  \centering
  \includegraphics[width=\linewidth]{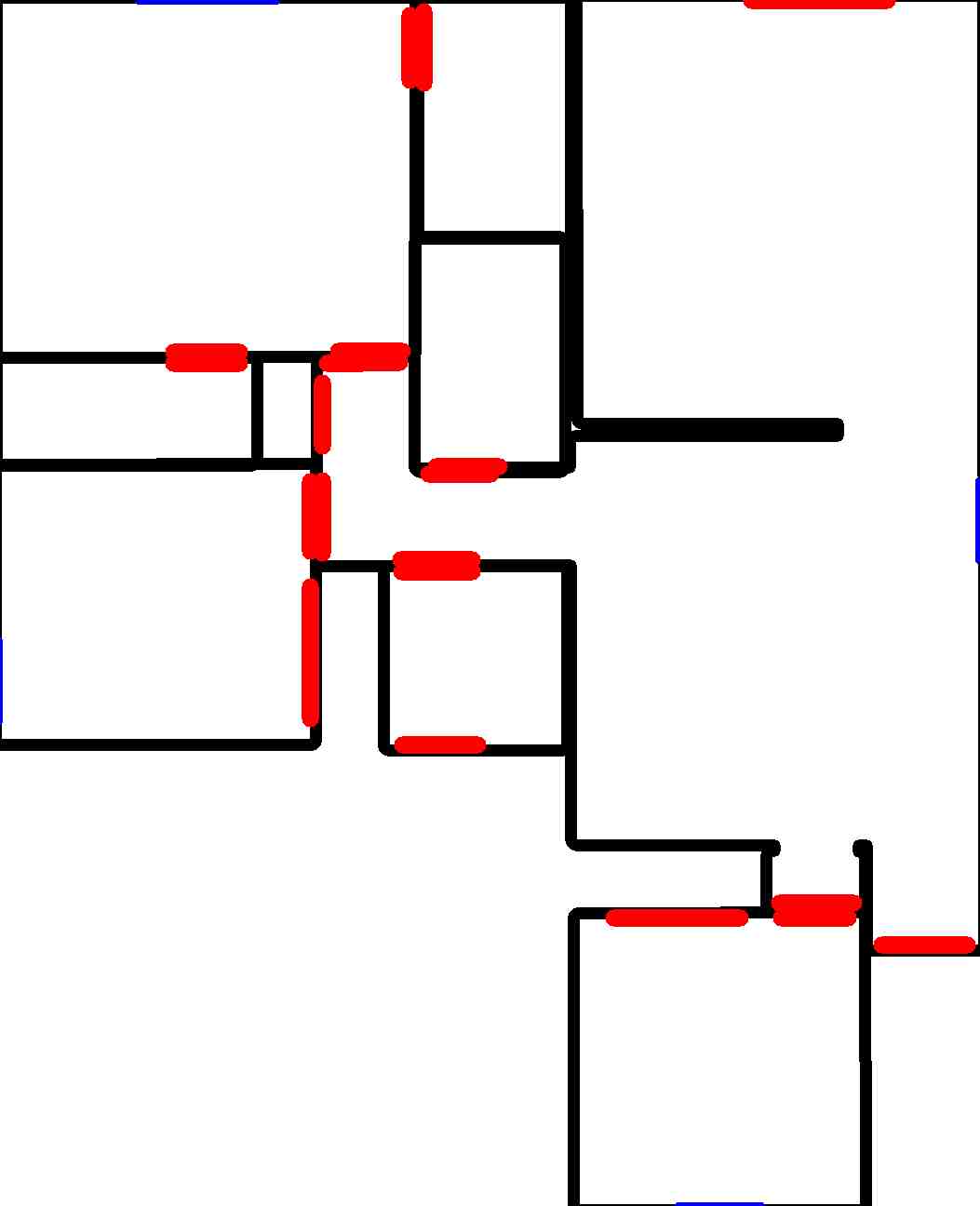}} &
\parbox[c]{\linewidth}{%
  \centering
  \includegraphics[width=\linewidth]{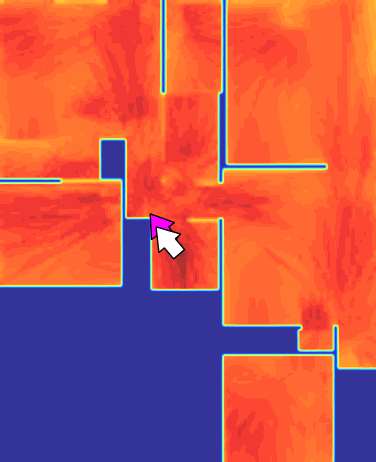}} &
\parbox[c]{\linewidth}{%
  \centering
  \includegraphics[width=\linewidth]{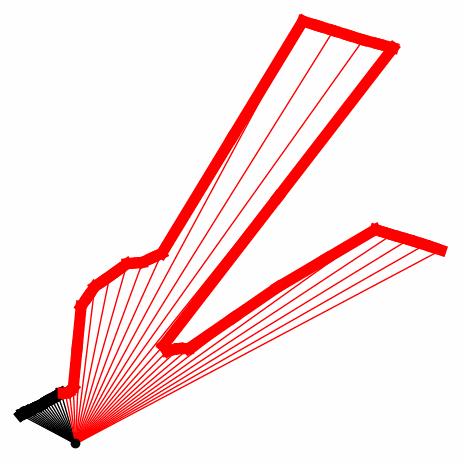}} &
\parbox[c]{\linewidth}{%
  \centering
  \includegraphics[width=\linewidth]{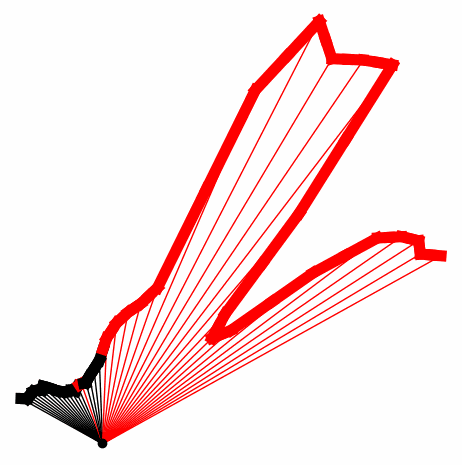}} \\
\noalign{\vskip 3pt}
\parbox[c]{\linewidth}{%
  \centering
  \includegraphics[width=\linewidth]{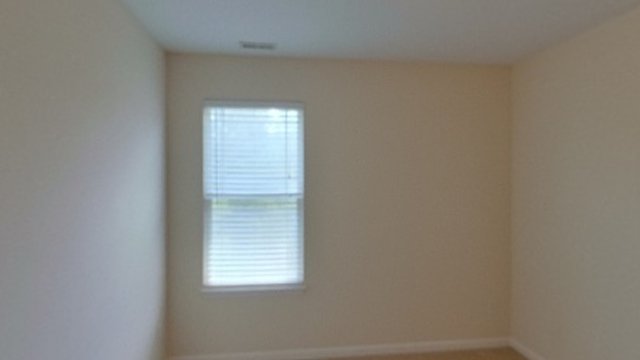}} &
\parbox[c]{\linewidth}{%
  \centering
  \includegraphics[width=\linewidth]{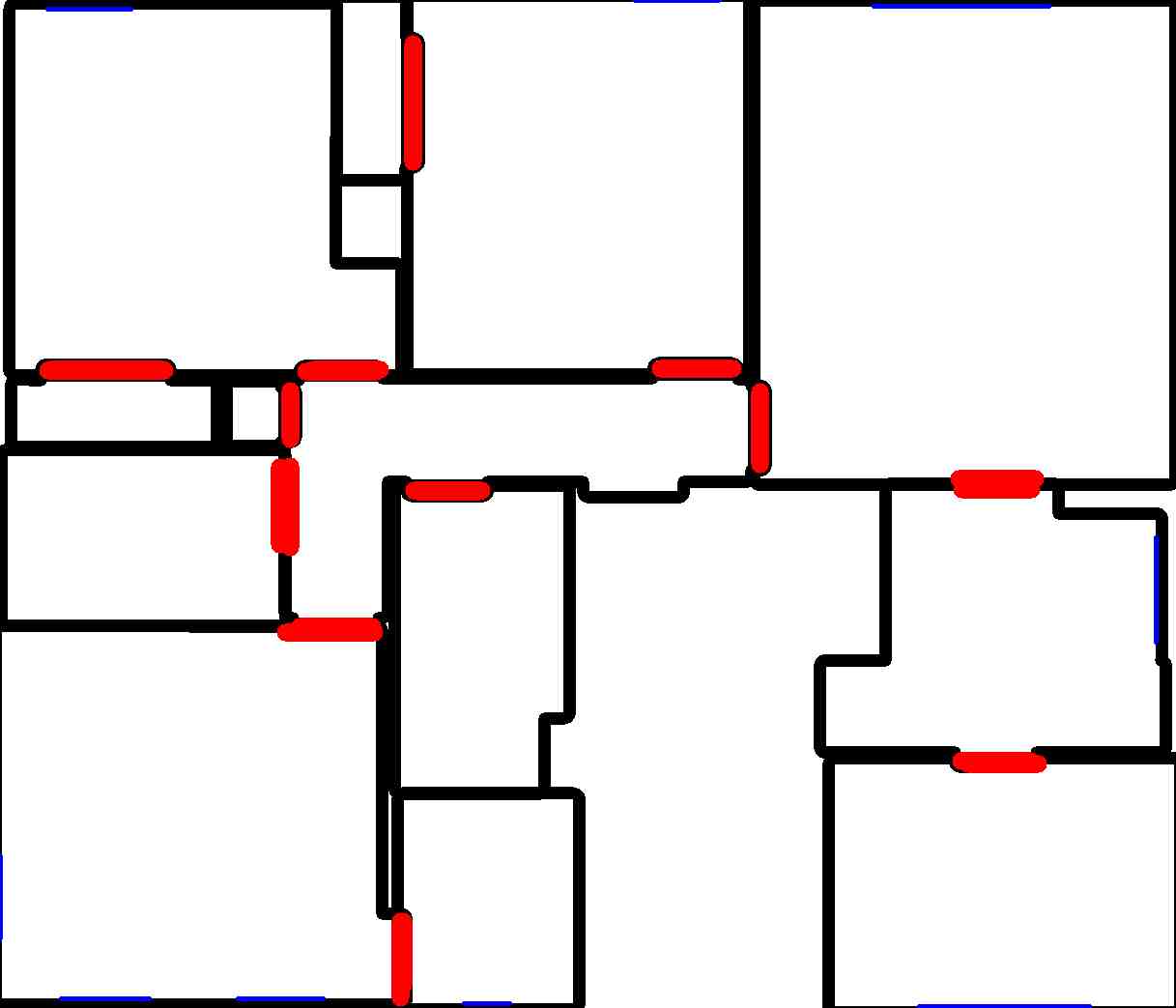}} &
\parbox[c]{\linewidth}{%
  \centering
  \includegraphics[width=\linewidth]{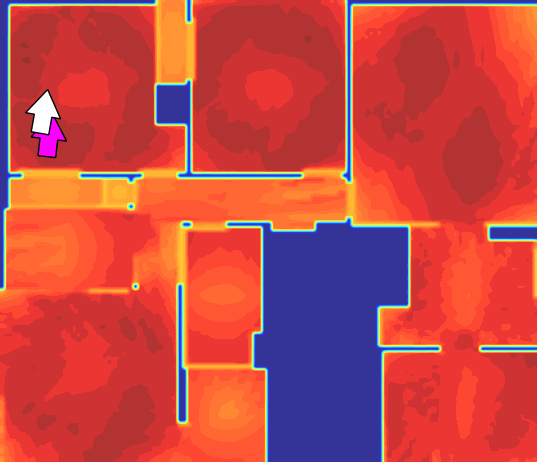}} &
\parbox[c]{\linewidth}{%
  \centering
  \includegraphics[width=\linewidth]{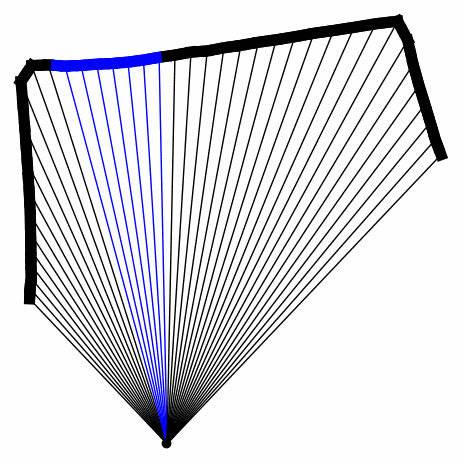}} &
\parbox[c]{\linewidth}{%
  \centering
  \includegraphics[width=\linewidth]{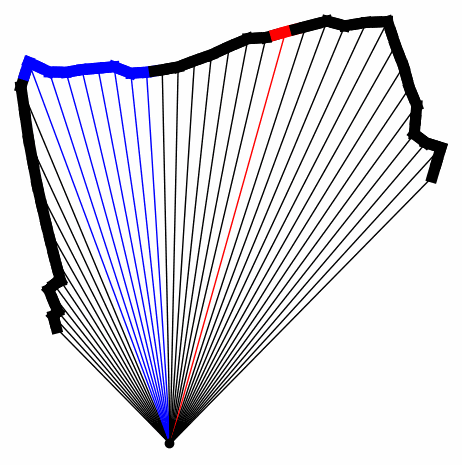}} \\
\noalign{\vskip 3pt}
\parbox[c]{\linewidth}{%
  \centering
  \includegraphics[width=\linewidth]{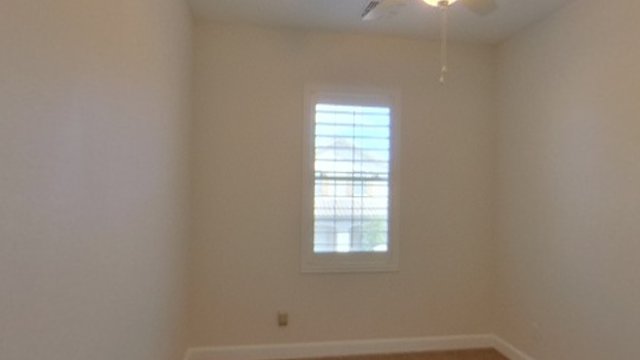}} &
\parbox[c]{\linewidth}{%
  \centering
  \includegraphics[width=\linewidth]{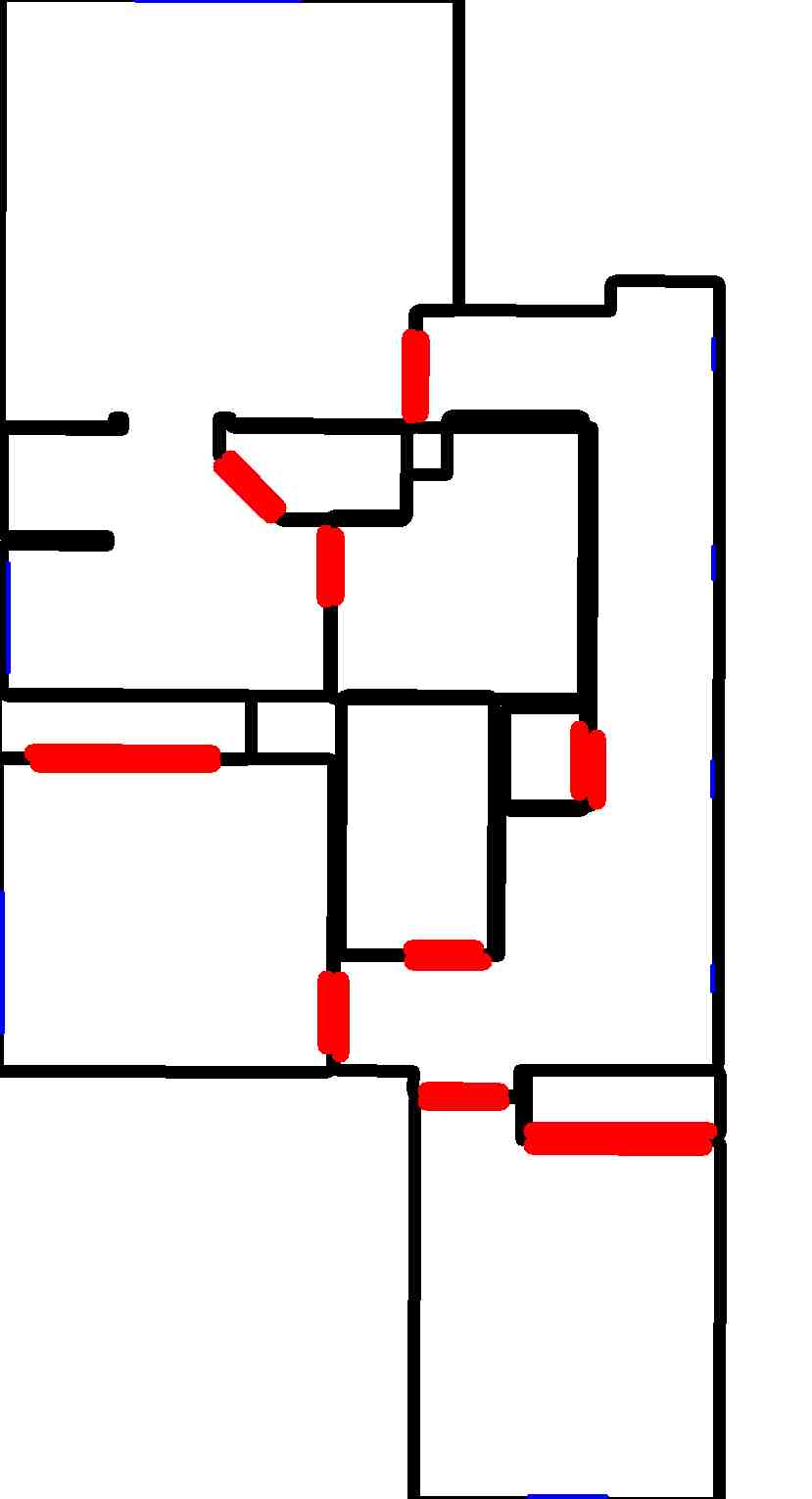}} &
\parbox[c]{\linewidth}{%
  \centering
  \includegraphics[width=\linewidth]{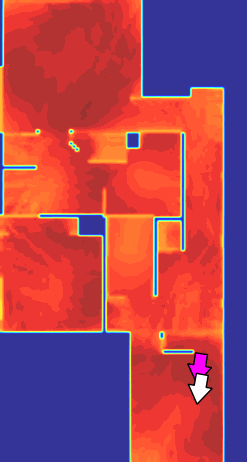}} &
\parbox[c]{\linewidth}{%
  \centering
  \includegraphics[width=\linewidth]{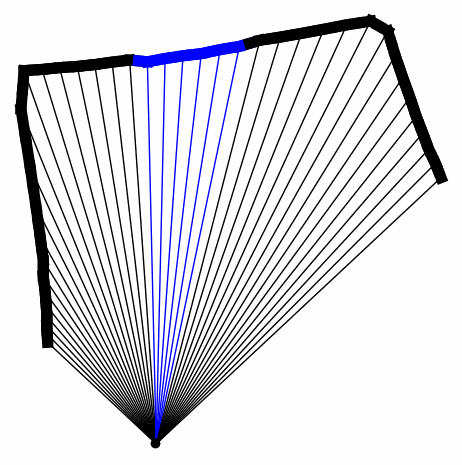}} &
\parbox[c]{\linewidth}{%
  \centering
  \includegraphics[width=\linewidth]{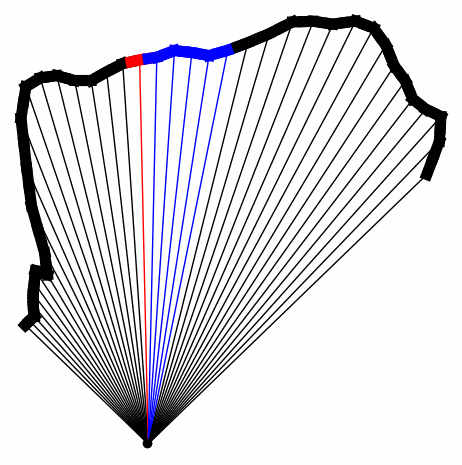}} \\
\noalign{\vskip 3pt}
\parbox[c]{\linewidth}{%
  \centering
  \includegraphics[width=\linewidth]{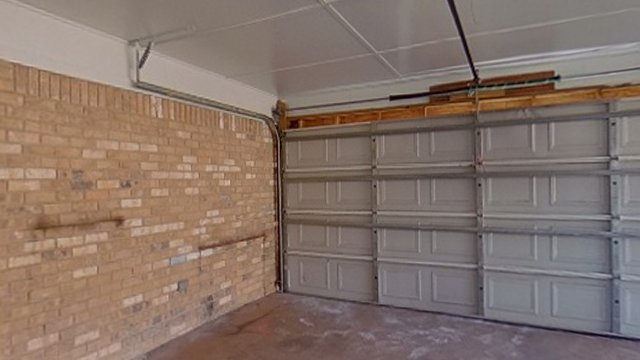}} &
\parbox[c]{\linewidth}{%
  \centering
  \includegraphics[width=\linewidth]{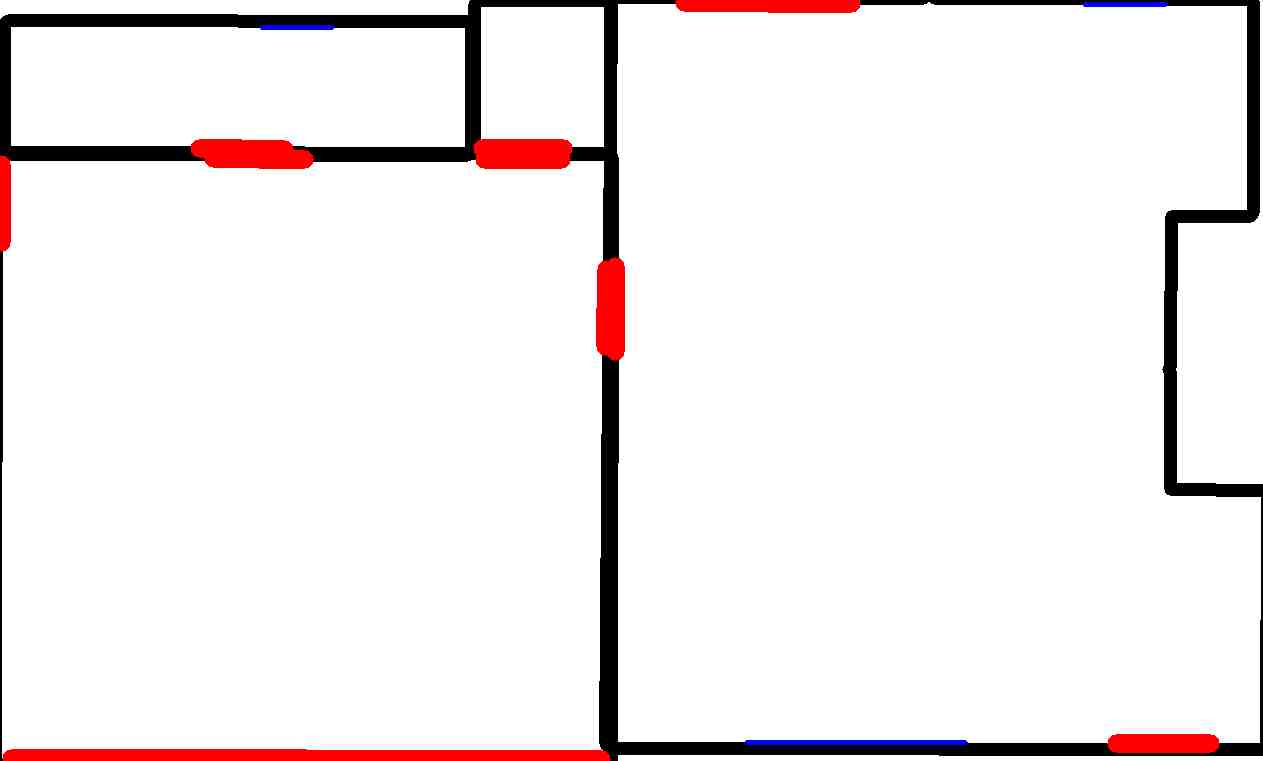}} &
\parbox[c]{\linewidth}{%
  \centering
  \includegraphics[width=\linewidth]{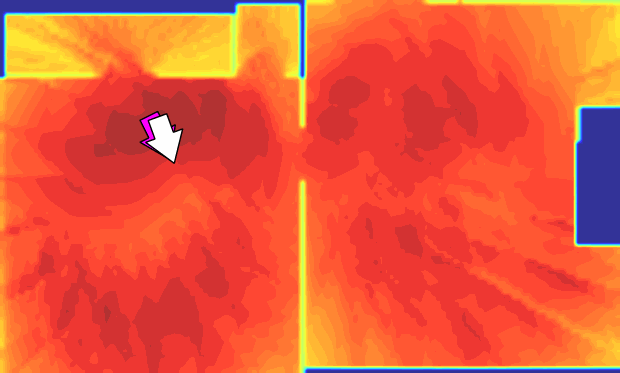}} &
\parbox[c]{\linewidth}{%
  \centering
  \includegraphics[width=\linewidth]{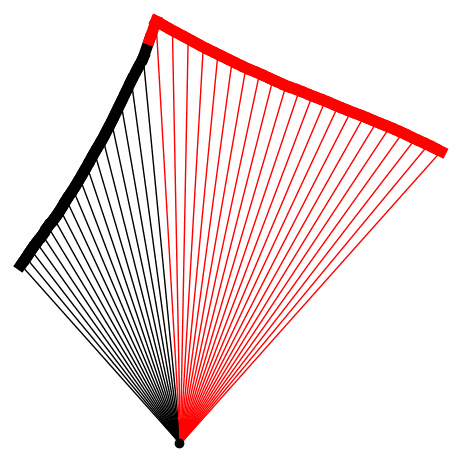}} &
\parbox[c]{\linewidth}{%
  \centering
  \includegraphics[width=\linewidth]{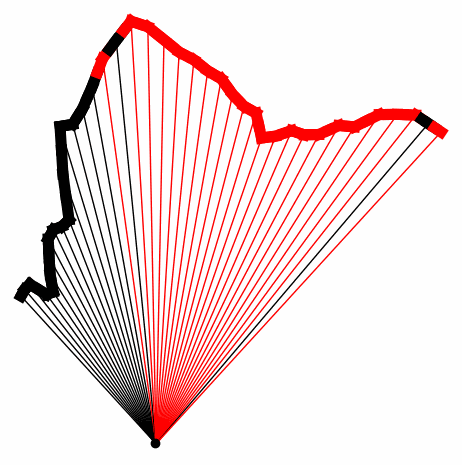}} \\
\hline
\end{tabular}

    \caption{
    \textbf{Additional Qualitative Results (ZInD dataset):}
    Warmer colors correspond to higher probabilities, while \textcolor{magenta}{magenta} indicates the ground-truth location and \textcolor{black}{white} denotes our predicted layout. Rays are: \textcolor{black}{wall}, \textcolor[HTML]{0000FF}{window}, and \textcolor[HTML]{FF0000}{door}.
    }
    \label{fig:good_examples_zind}
\end{figure*}

In this section, we show more visual examples from our predictions on both datasets. Figure~\ref{fig:good_examples_s3d} presents several successful examples from the S3D dataset, illustrating how combining precise semantic information with structural data can yield accurate localizations. 
We added an interpolated line, colored by each ray’s semantic label, connecting the ray endpoints to make interpolation easier.
Figure~\ref{fig:good_examples_zind} further demonstrates our predictions on the ZInD dataset. In both figures, warmer colors correspond to higher probabilities, with \textcolor{magenta}{magenta} indicating the ground-truth location and \textcolor{black}{white} denoting our predicted layout. The strong similarity between the ground-truth rays and the predicted rays underlines the effectiveness of our method.

\subsection{Visualization of Baseline Comparisons}
\label{sec:baseline_comparison}
\begin{figure*}[htbp]
    \centering
    \centering
\renewcommand{\arraystretch}{1.2}
\begin{tabular}{%
  >{\centering\arraybackslash}m{0.25\textwidth}%
  >{\centering\arraybackslash}m{0.15\textwidth}%
  >{\centering\arraybackslash}m{0.15\textwidth}%
  >{\centering\arraybackslash}m{0.15\textwidth}%
  >{\centering\arraybackslash}m{0.15\textwidth}%
}
\hline
\textbf{Input Image} & \textbf{Floorplan} & \textbf{Ours} & \textbf{F3Loc} & \textbf{LASER} \\
\hline
\noalign{\vskip 3pt}
\parbox[c]{\linewidth}{%
  \centering
  \includegraphics[width=\linewidth]{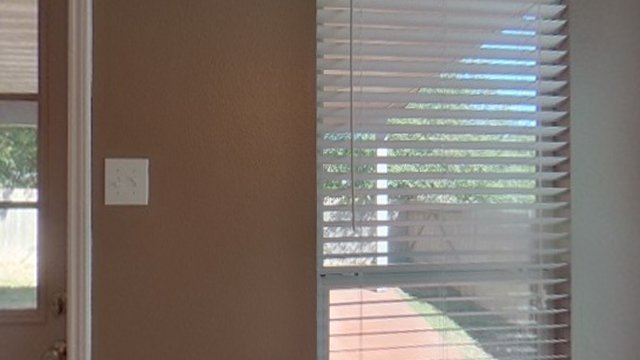}} &
\parbox[c]{\linewidth}{%
  \centering
  \includegraphics[width=\linewidth]{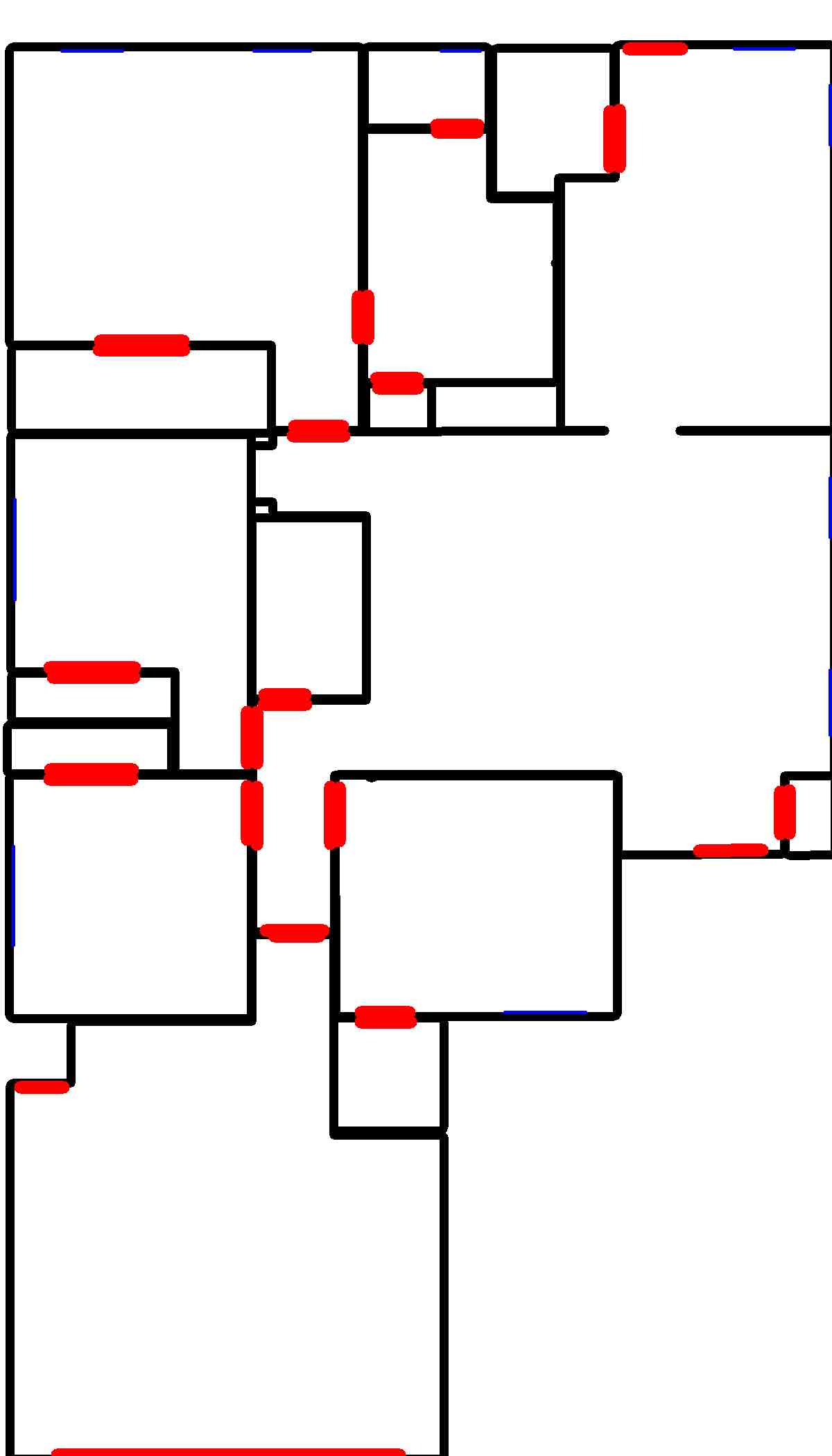}} &
\parbox[c]{\linewidth}{%
  \centering
  \includegraphics[width=\linewidth]{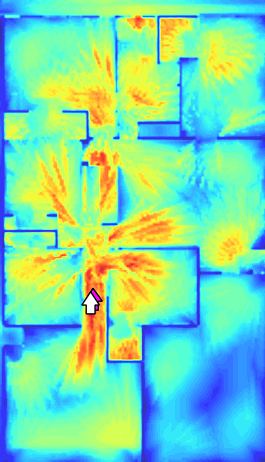}} &
\parbox[c]{\linewidth}{%
  \centering
  \includegraphics[width=\linewidth]{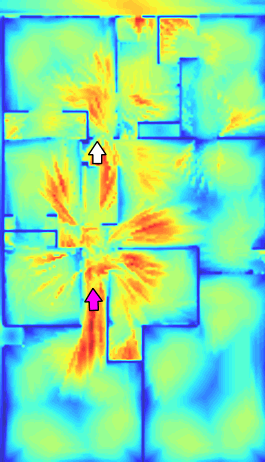}} &
\parbox[c]{\linewidth}{%
  \centering
  \includegraphics[width=\linewidth]{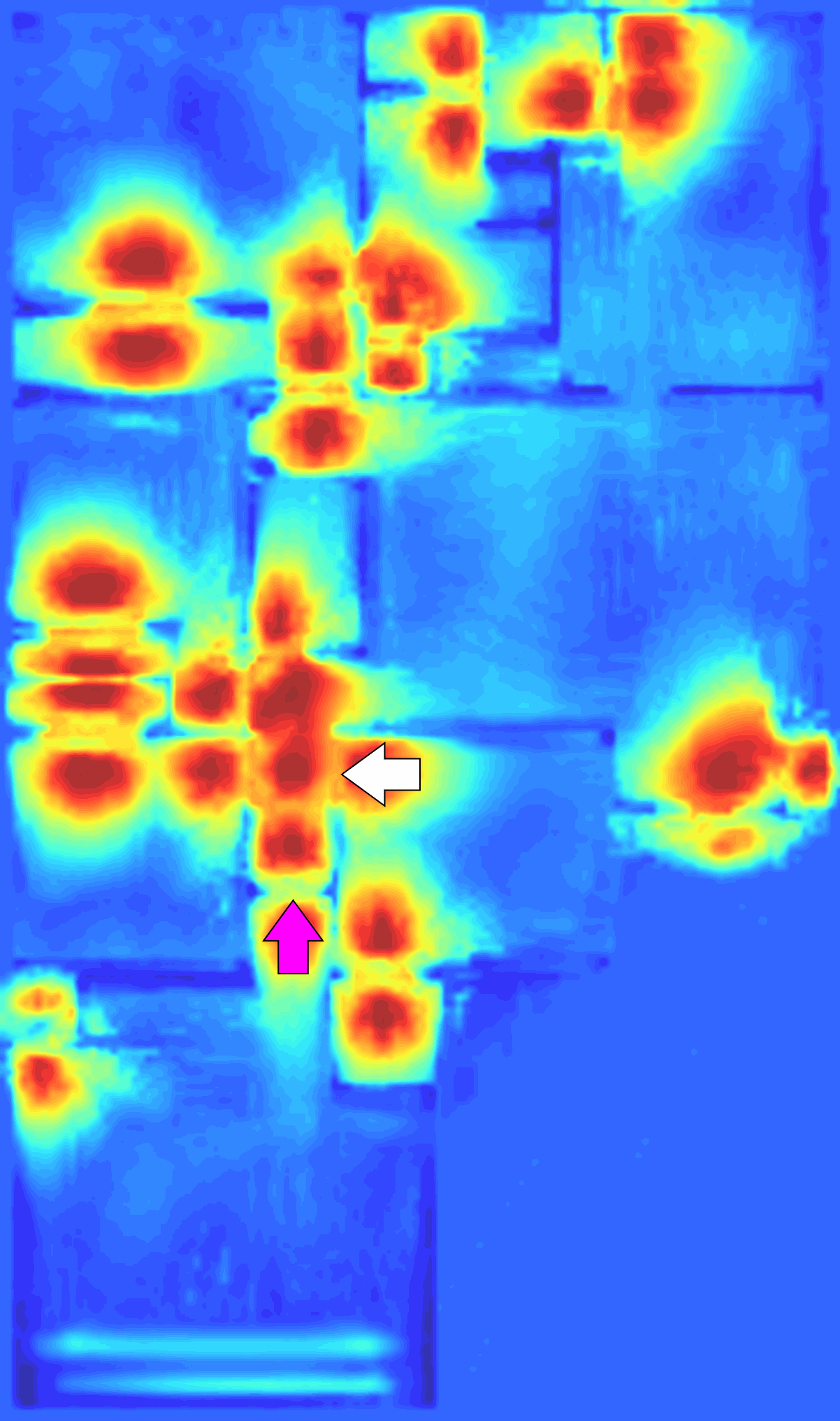}} \\
\noalign{\vskip 3pt}
\hline
\noalign{\vskip 3pt}
\parbox[c]{\linewidth}{%
  \centering
  \includegraphics[width=\linewidth]{supp/plots/ablation/method_comapre_results_zind/1050_f2_i.jpg}} &
\parbox[c]{\linewidth}{%
  \centering
  \includegraphics[width=\linewidth]{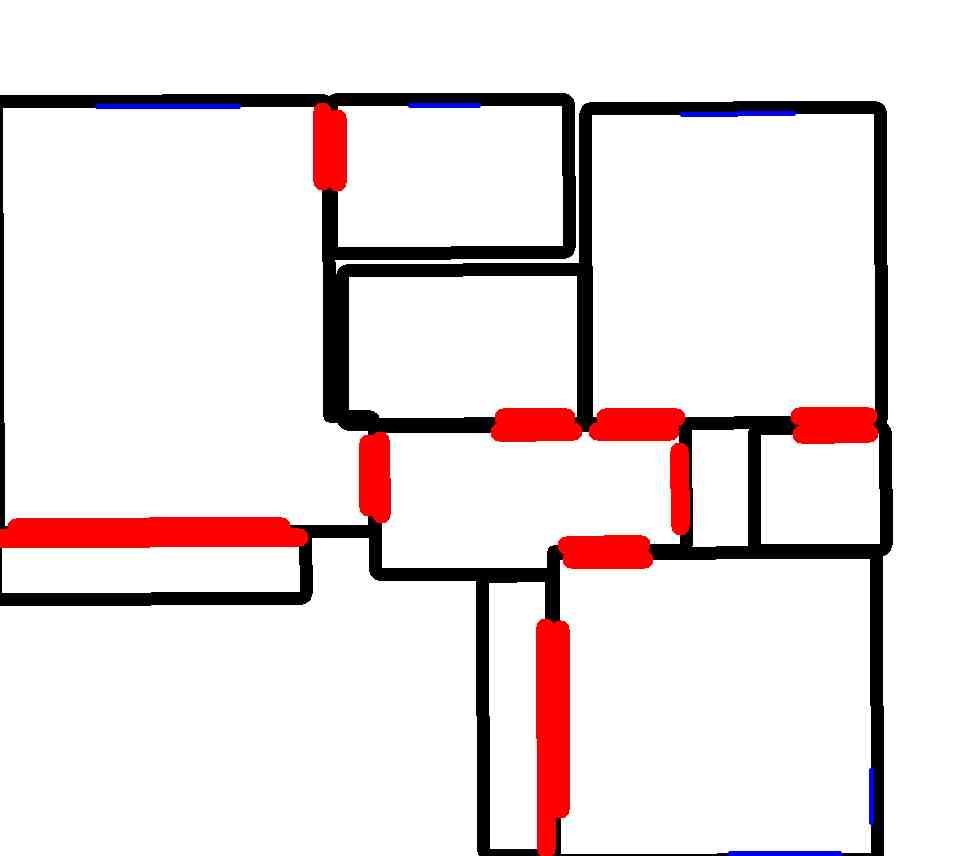}} &
\parbox[c]{\linewidth}{%
  \centering
  \includegraphics[width=\linewidth]{supp/plots/ablation/method_comapre_results_zind/scene_1050_floor_02_7-with_refine.png}} &
\parbox[c]{\linewidth}{%
  \centering
  \includegraphics[width=\linewidth]{supp/plots/ablation/method_comapre_results_zind/scene_1050_floor_02_7-depth.png}} &
\parbox[c]{\linewidth}{%
  \centering
  \includegraphics[width=\linewidth]{supp/plots/ablation/method_comapre_results_zind/Laser/scene_1050_floor_02_7-with_refine.png}} \\
\noalign{\vskip 3pt}
\hline
\noalign{\vskip 3pt}
\parbox[c]{\linewidth}{%
  \centering
  \includegraphics[width=\linewidth]{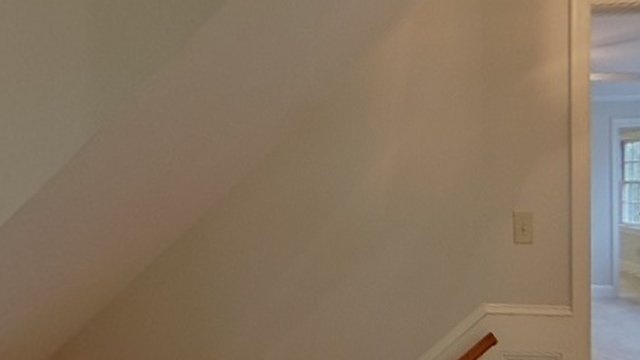}} &
\parbox[c]{\linewidth}{%
  \centering
  \includegraphics[width=\linewidth]{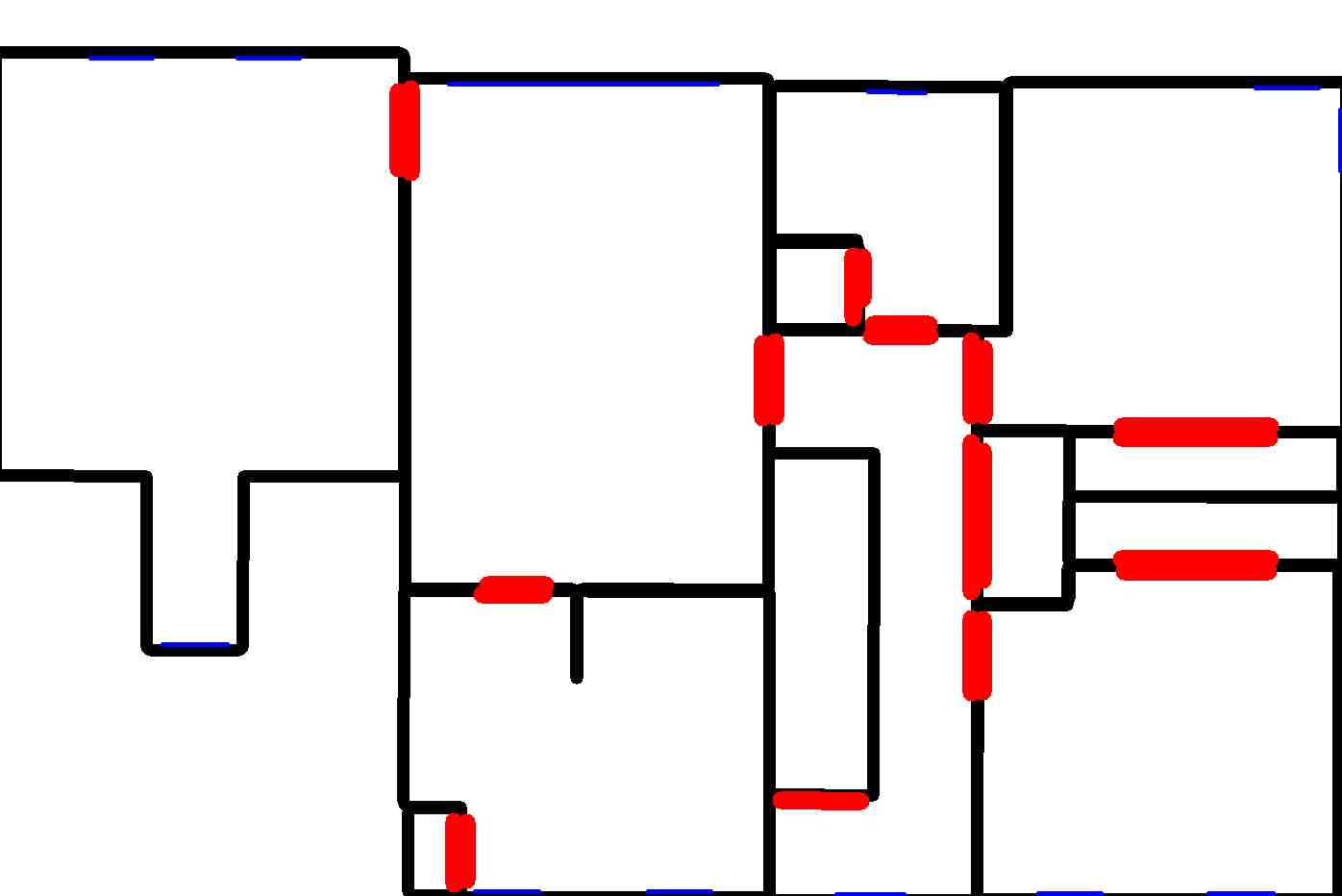}} &
\parbox[c]{\linewidth}{%
  \centering
  \includegraphics[width=\linewidth]{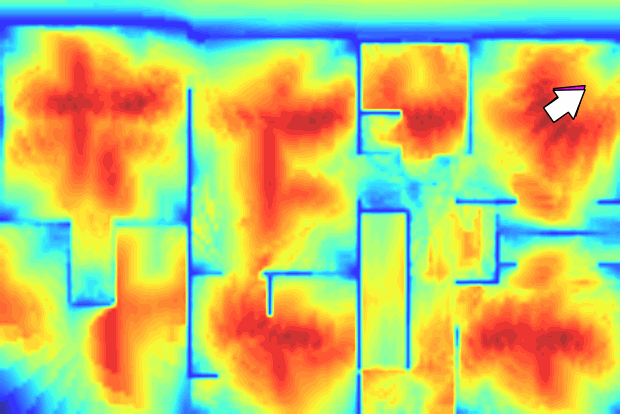}} &
\parbox[c]{\linewidth}{%
  \centering
  \includegraphics[width=\linewidth]{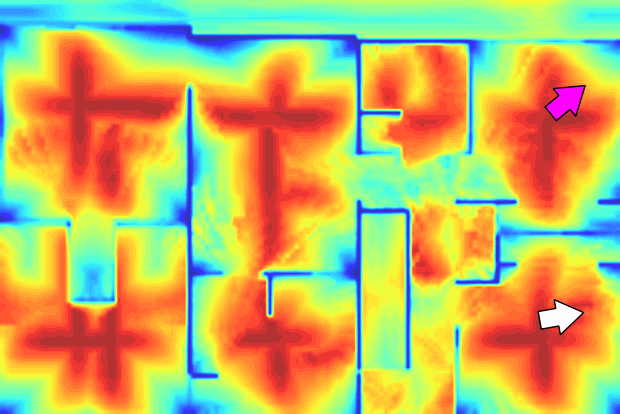}} &
\parbox[c]{\linewidth}{%
  \centering
  \includegraphics[width=\linewidth]{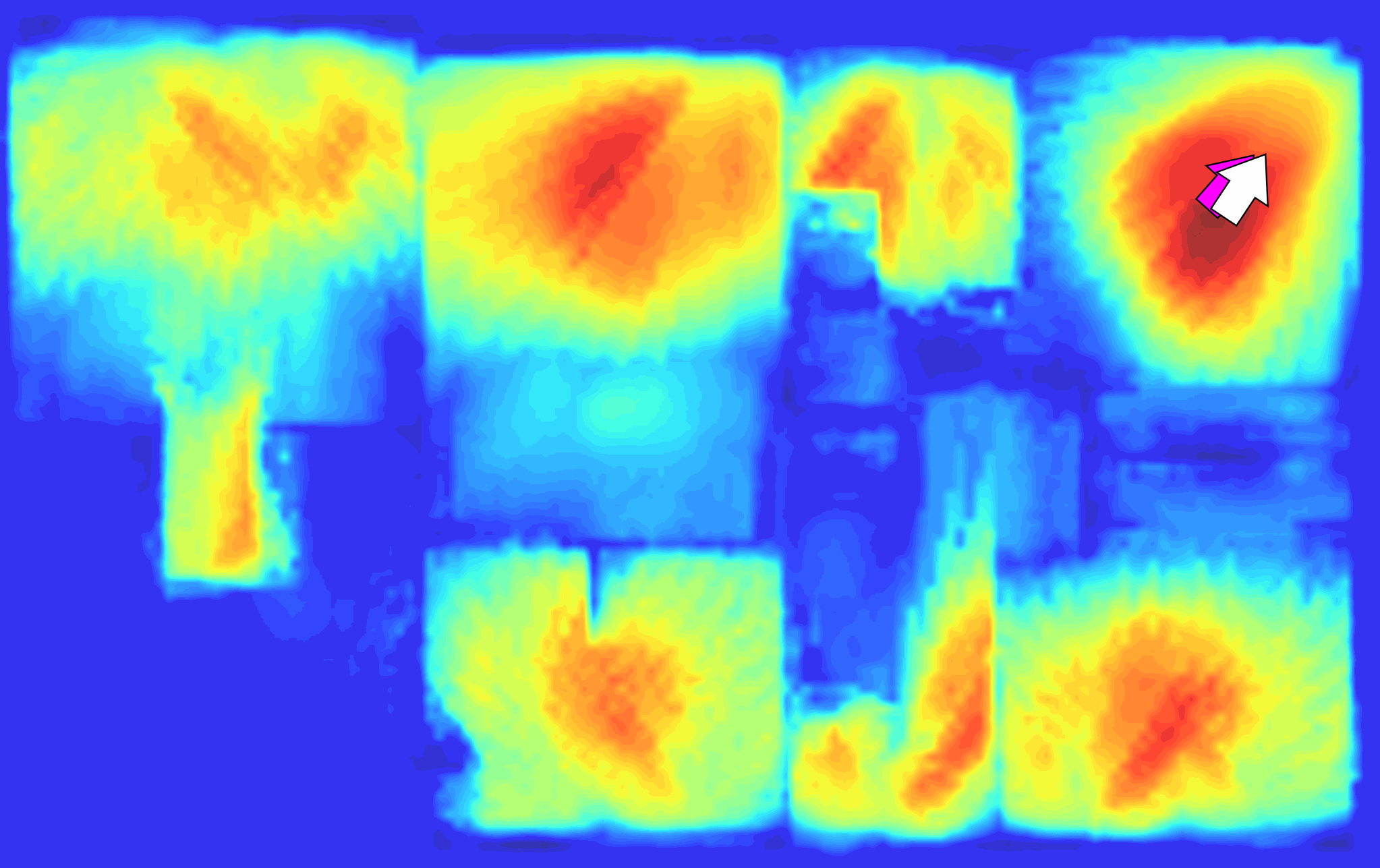}} \\
\noalign{\vskip 3pt}
\hline
\noalign{\vskip 3pt}
\parbox[c]{\linewidth}{%
  \centering
  \includegraphics[width=\linewidth]{supp/plots/ablation/method_comapre_results_zind/1001_f1_i.jpg}} &
\parbox[c]{\linewidth}{%
  \centering
  \includegraphics[width=\linewidth]{supp/plots/ablation/method_comapre_results_zind/1001_f1.jpg}} &
\parbox[c]{\linewidth}{%
  \centering
  \includegraphics[width=\linewidth]{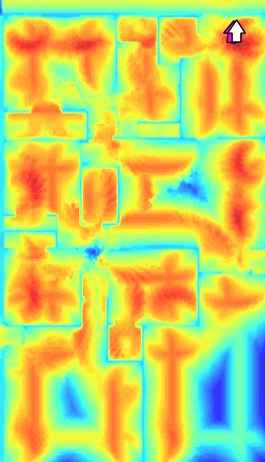}} &
\parbox[c]{\linewidth}{%
  \centering
  \includegraphics[width=\linewidth]{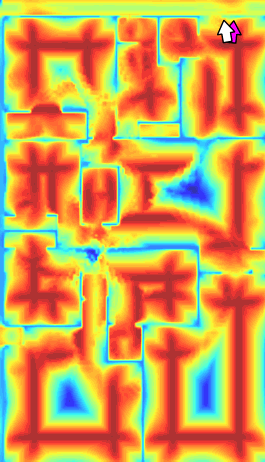}} &
\parbox[c]{\linewidth}{%
  \centering
  \includegraphics[width=\linewidth]{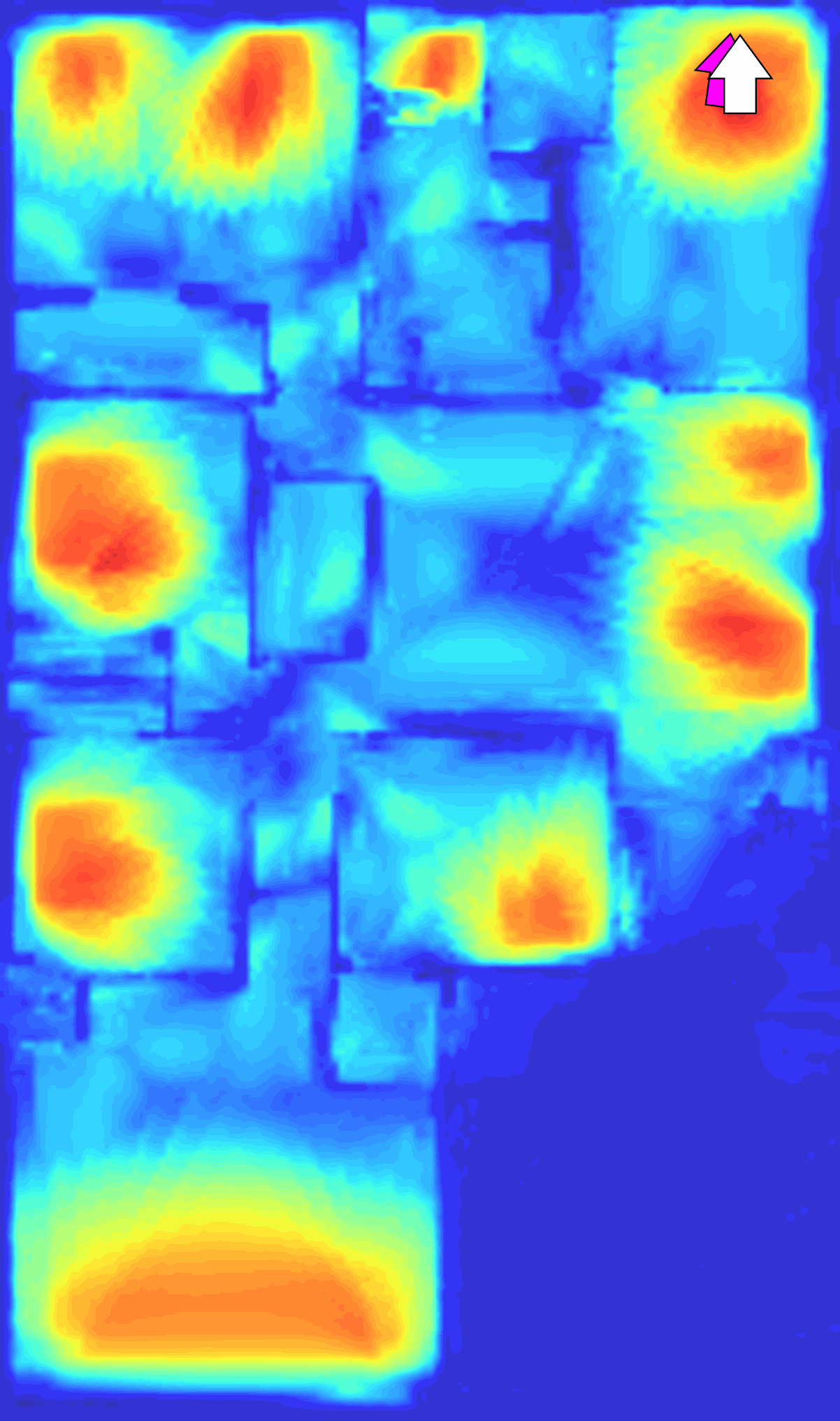}} \\
\noalign{\vskip 3pt}
\hline
\noalign{\vskip 3pt}
\parbox[c]{\linewidth}{%
  \centering
  \includegraphics[width=\linewidth]{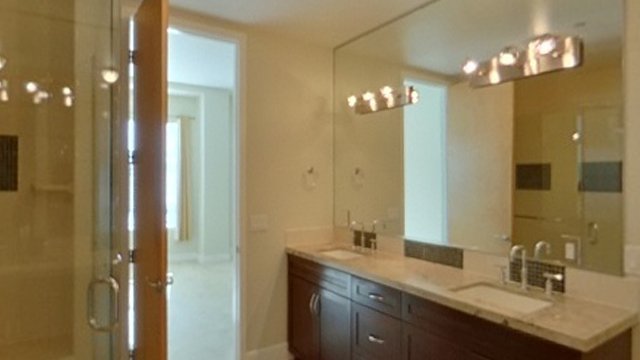}} &
\parbox[c]{\linewidth}{%
  \centering
  \includegraphics[width=\linewidth]{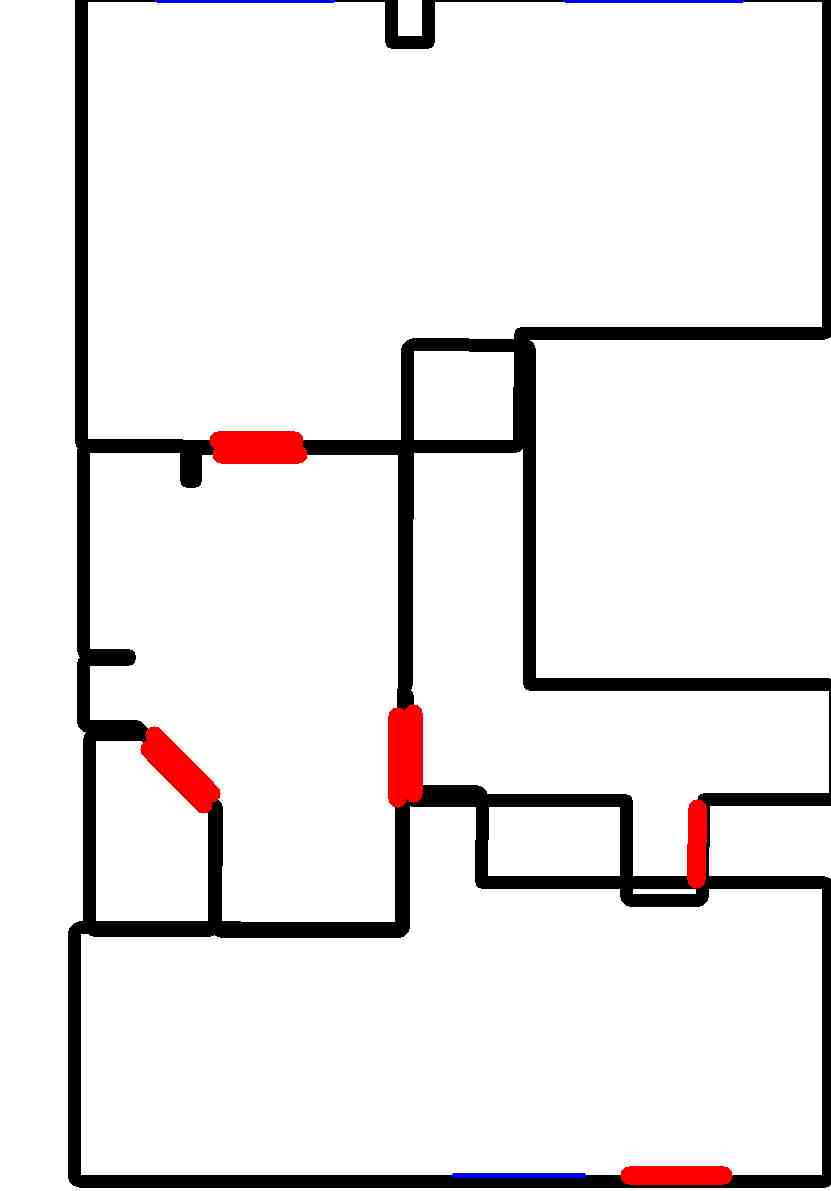}} &
\parbox[c]{\linewidth}{%
  \centering
  \includegraphics[width=\linewidth]{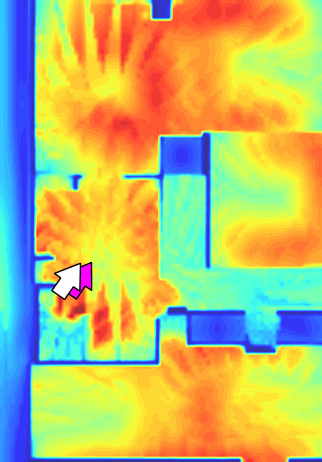}} &
\parbox[c]{\linewidth}{%
  \centering
  \includegraphics[width=\linewidth]{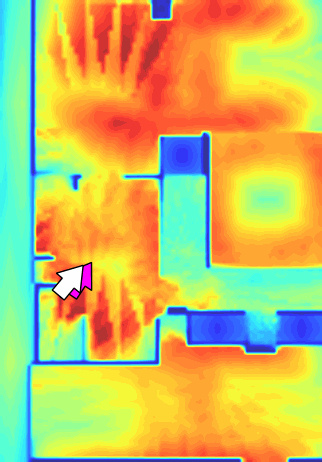}} &
\parbox[c]{\linewidth}{%
  \centering
  \includegraphics[width=\linewidth]{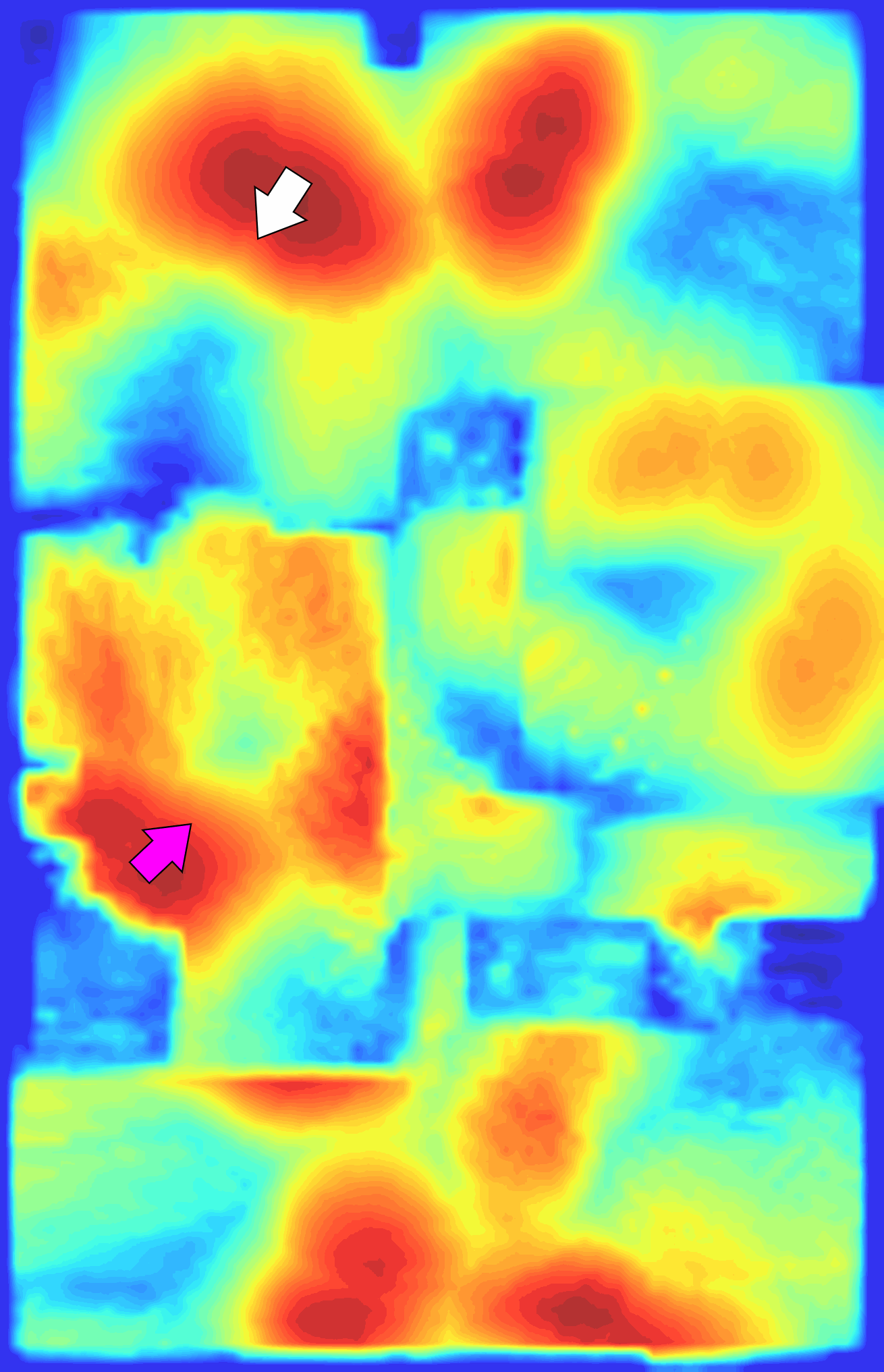}} \\
\noalign{\vskip 3pt}
\hline
\noalign{\vskip 3pt}
\parbox[c]{\linewidth}{%
  \centering
  \includegraphics[width=\linewidth]{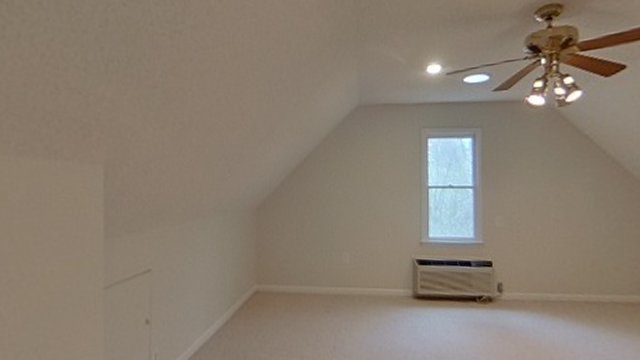}} &
\parbox[c]{\linewidth}{%
  \centering
  \includegraphics[width=\linewidth]{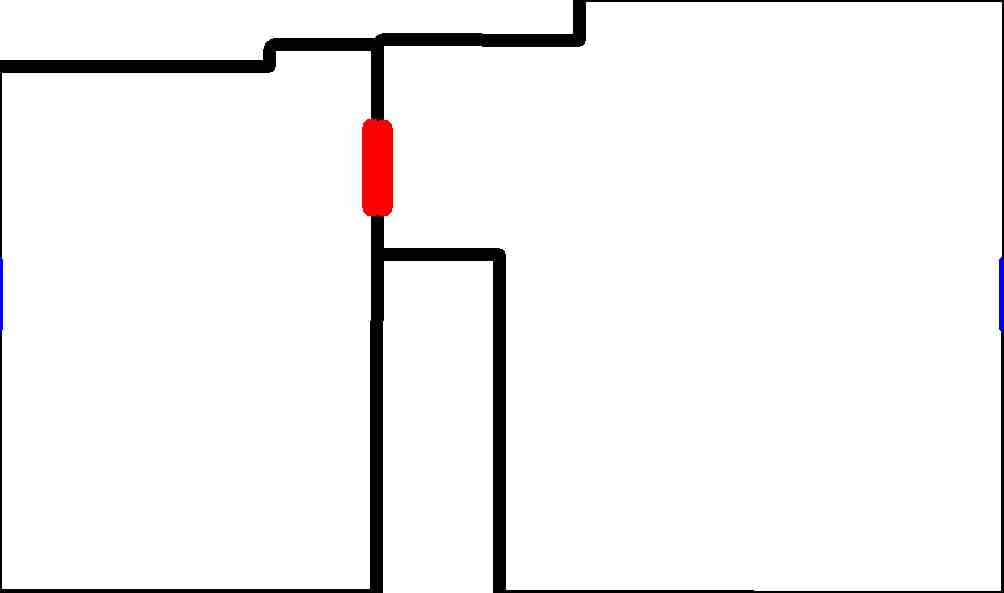}} &
\parbox[c]{\linewidth}{%
  \centering
  \includegraphics[width=\linewidth]{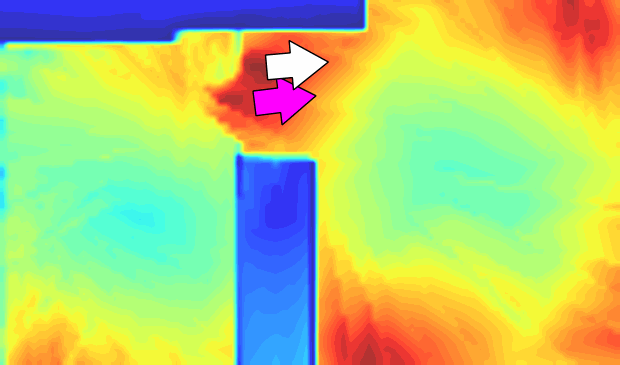}} &
\parbox[c]{\linewidth}{%
  \centering
  \includegraphics[width=\linewidth]{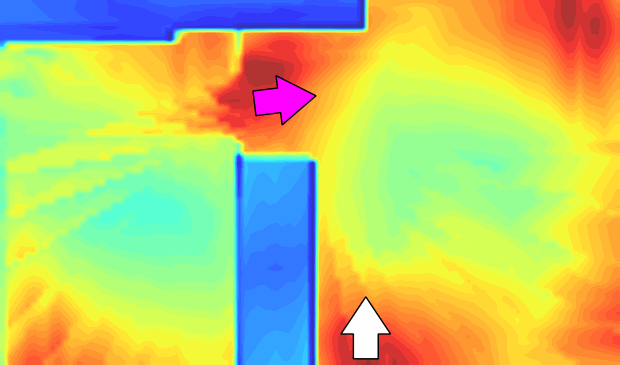}} &
\parbox[c]{\linewidth}{%
  \centering
  \includegraphics[width=\linewidth]{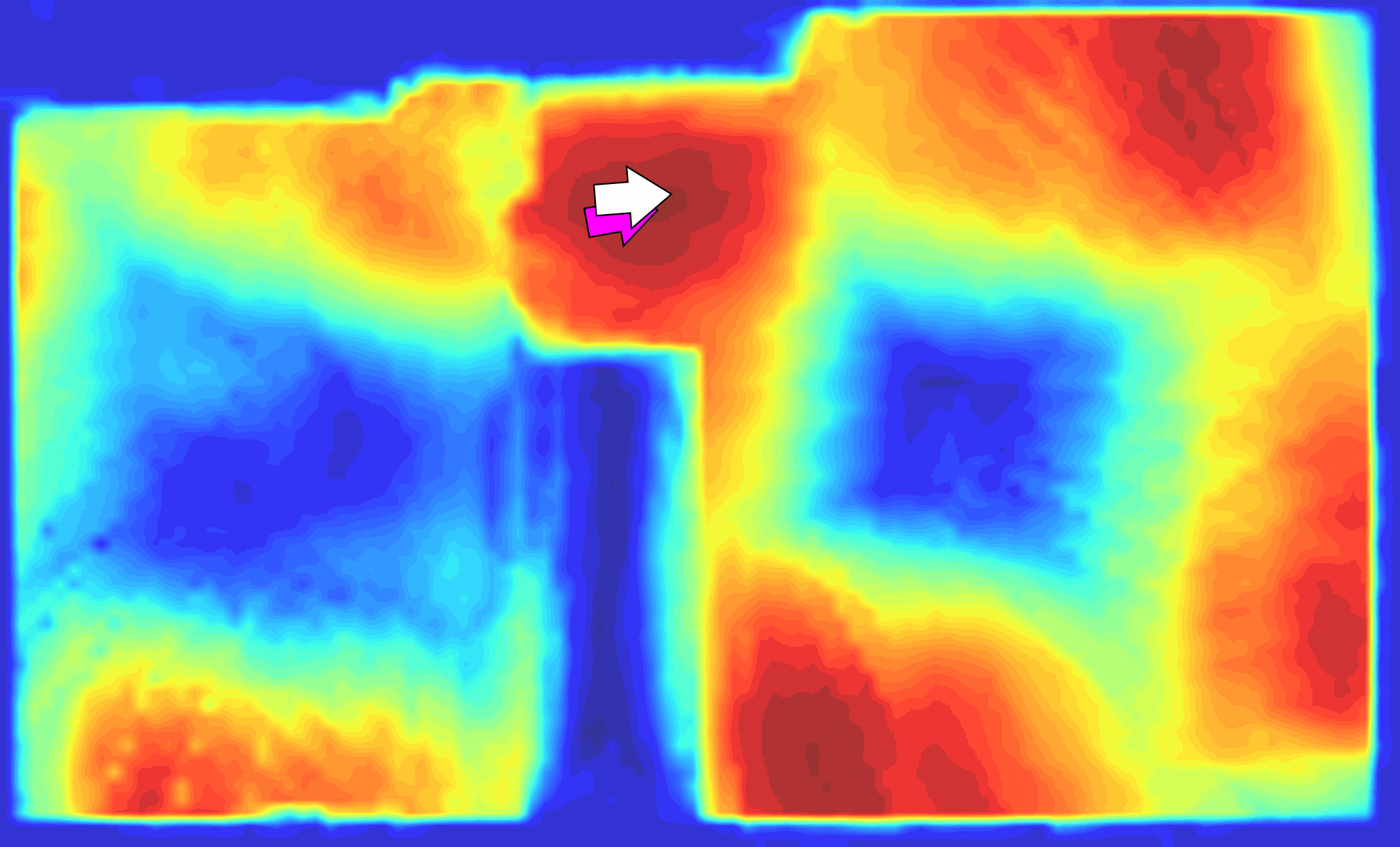}} \\
\noalign{\vskip 3pt}
\hline
\end{tabular}

    \caption{\textbf{Comparison to Baseline Methods:} Additional visualizations comparing our method with F3Loc and LASER on the ZiND dataset. Warmer colors correspond to regions with higher predicted probabilities. Overlaid on the estimated probabilities, we indicate the ground truth location (\textcolor{magenta}{magenta}) and the predicted location. Rays are: \textcolor{black}{wall}, \textcolor[HTML]{0000FF}{window}, and \textcolor[HTML]{FF0000}{door}.}
    \label{fig:method_compare_example}
\end{figure*}

Here, we present additional examples comparing our method against baseline approaches, specifically F3Loc and LASER, on the ZiND dataset. More visual examples of these comparisons are shown in Figure~\ref{fig:method_compare_example}. 

\section{Limitations}
\label{sec:limitation}
\begin{figure*}[htbp]
    \centering
    \centering
\renewcommand{\arraystretch}{1.2}
\begin{tabular}{%
  >{\centering\arraybackslash}m{0.25\textwidth}%
  >{\centering\arraybackslash}m{0.15\textwidth}%
  >{\centering\arraybackslash}m{0.15\textwidth}%
  >{\centering\arraybackslash}m{0.15\textwidth}%
  >{\centering\arraybackslash}m{0.15\textwidth}%
}
\hline
\textbf{Input Image} & \textbf{Floorplan} & \textbf{Ours} & \textbf{Ground Truth Rays} & \textbf{Predicted Rays} \\
\hline
\noalign{\vskip 3pt}
\parbox[c]{\linewidth}{%
  \centering
  \includegraphics[width=\linewidth]{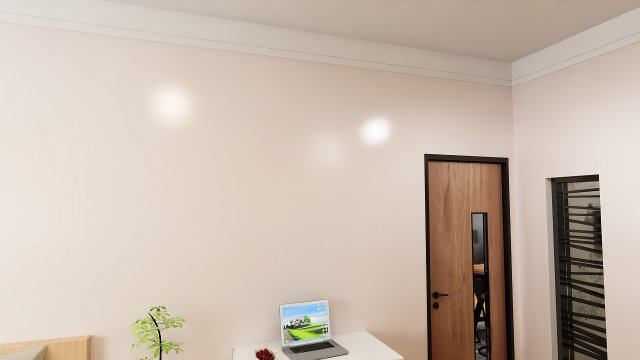}} &
\parbox[c]{\linewidth}{%
  \centering
  \includegraphics[width=\linewidth]{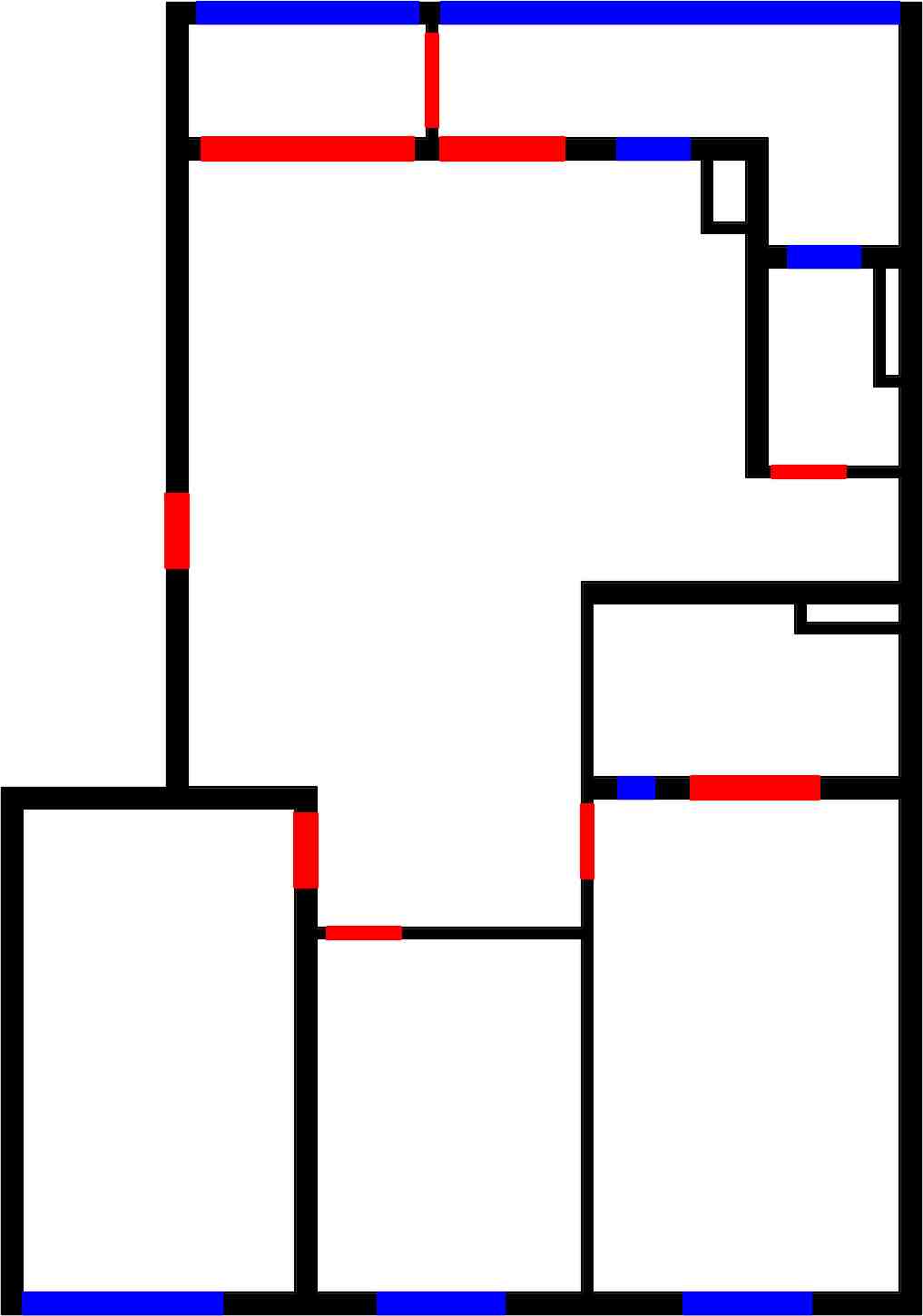}} &
\parbox[c]{\linewidth}{%
  \centering
  \includegraphics[width=\linewidth]{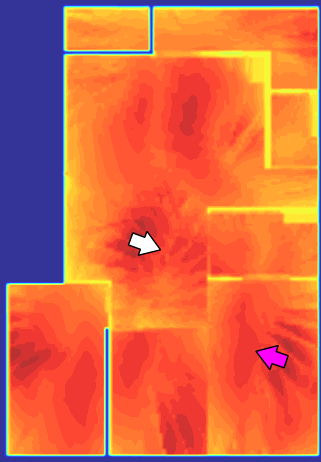}} &
\parbox[c]{\linewidth}{%
  \centering
  \includegraphics[width=\linewidth]{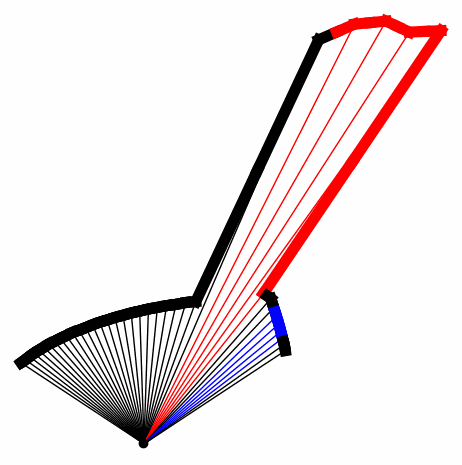}} &
\parbox[c]{\linewidth}{%
  \centering
  \includegraphics[width=\linewidth]{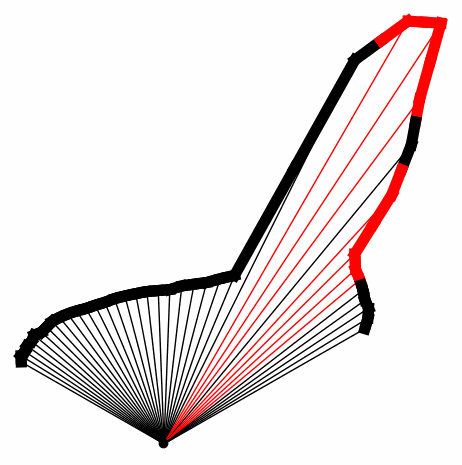}} \\
\noalign{\vskip 3pt}

\parbox[c]{\linewidth}{%
  \centering
  \includegraphics[width=\linewidth]{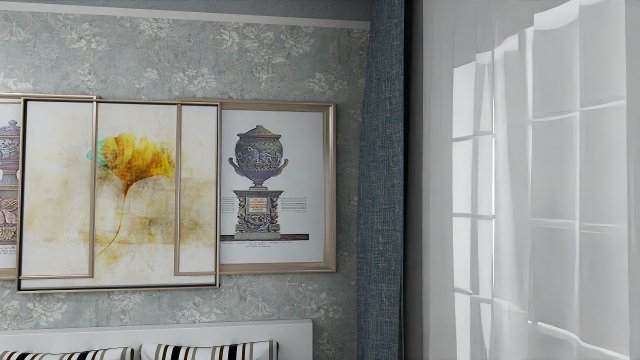}} &
\parbox[c]{\linewidth}{%
  \centering
  \includegraphics[width=\linewidth]{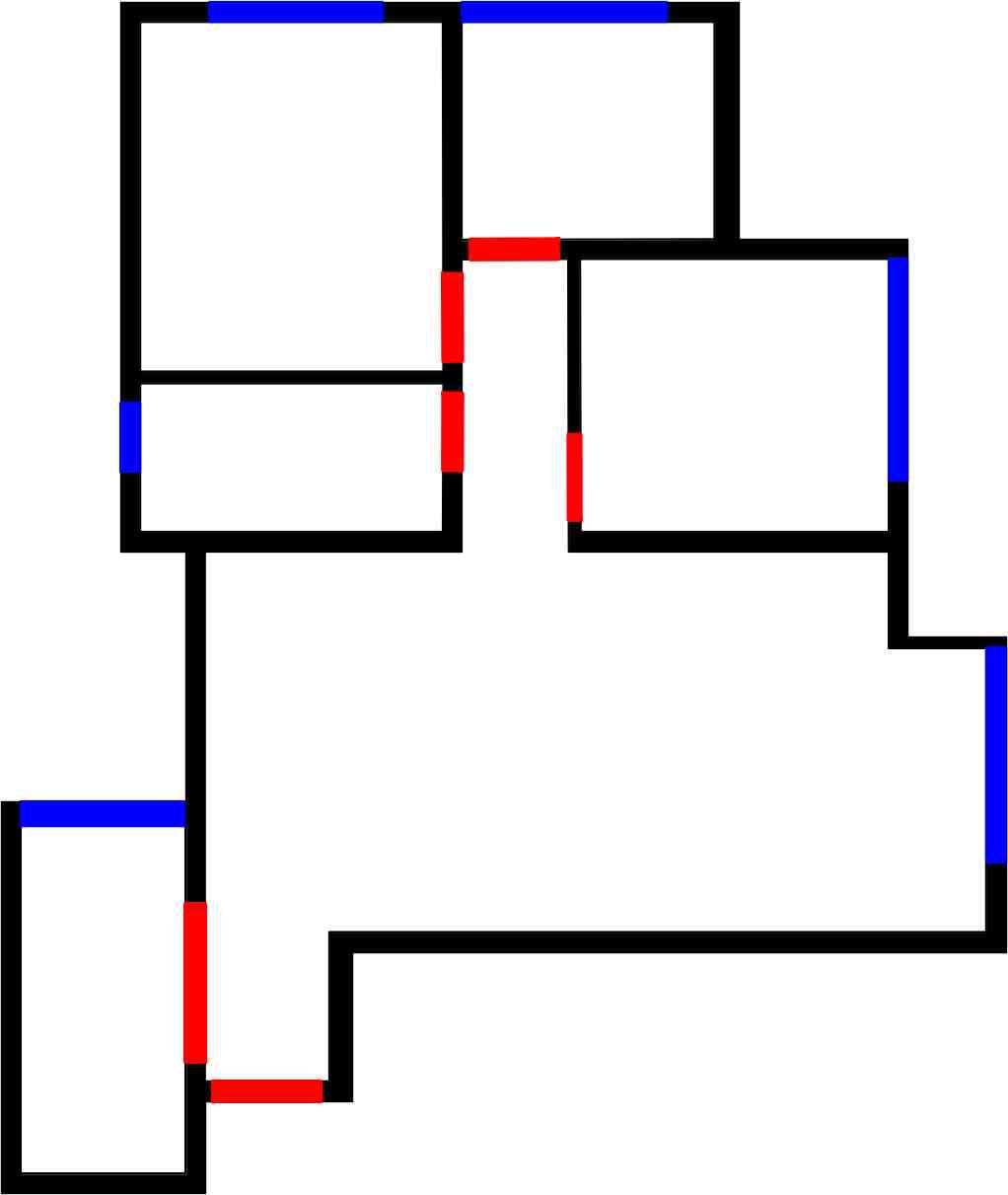}} &
\parbox[c]{\linewidth}{%
  \centering
  \includegraphics[width=\linewidth]{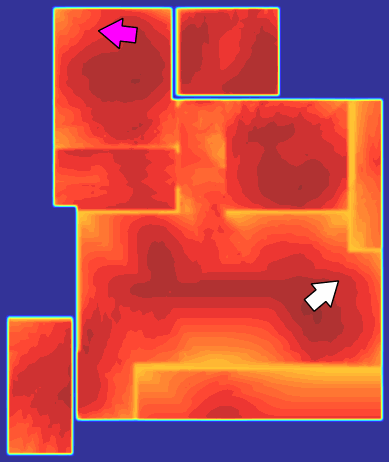}} &
\parbox[c]{\linewidth}{%
  \centering
  \includegraphics[width=\linewidth]{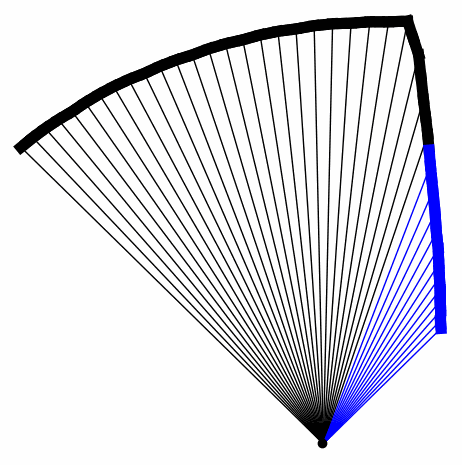}} &
\parbox[c]{\linewidth}{%
  \centering
  \includegraphics[width=\linewidth]{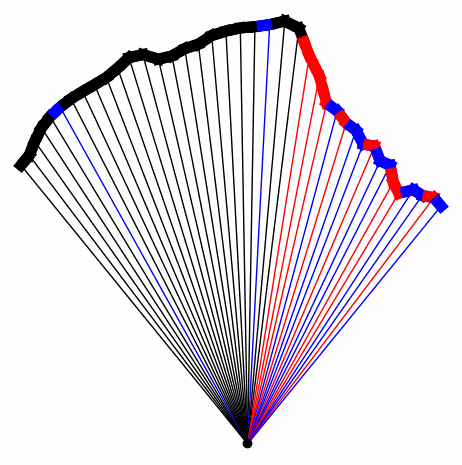}} \\
\noalign{\vskip 3pt}
\parbox[c]{\linewidth}{%
  \centering
  \includegraphics[width=\linewidth]{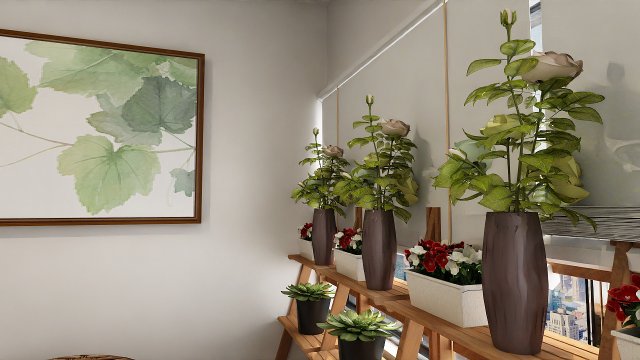}} &
\parbox[c]{\linewidth}{%
  \centering
  \includegraphics[width=\linewidth]{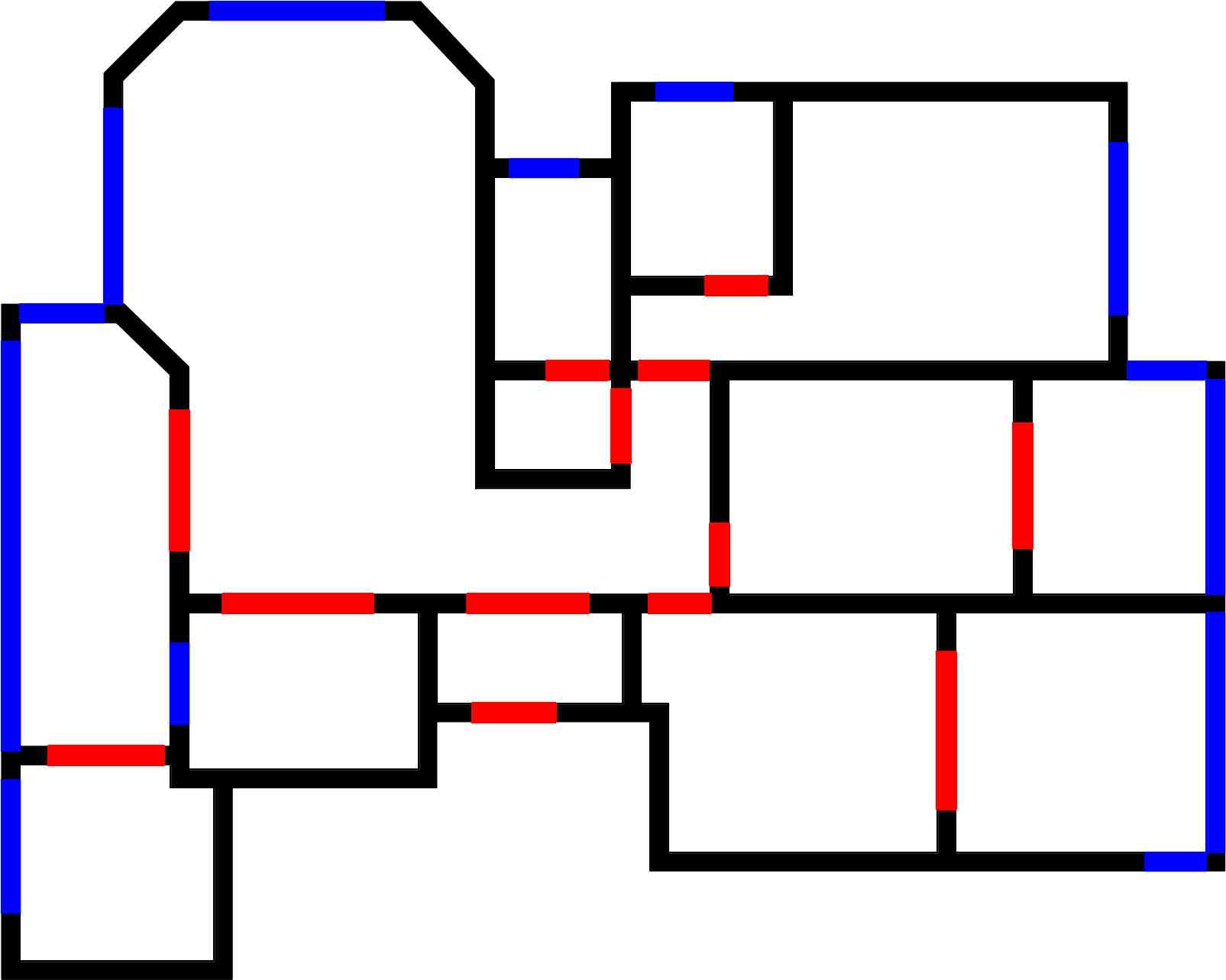}} &
\parbox[c]{\linewidth}{%
  \centering
  \includegraphics[width=\linewidth]{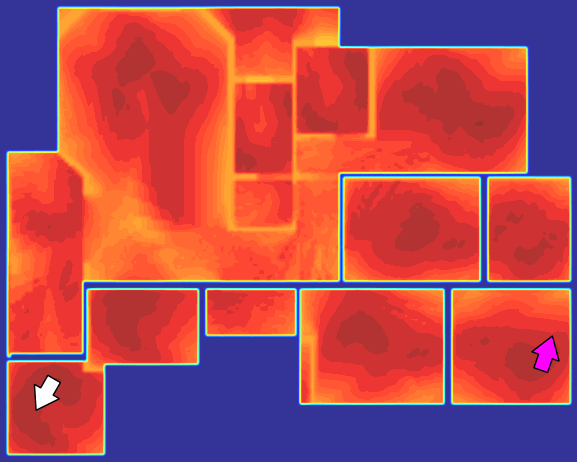}} &
\parbox[c]{\linewidth}{%
  \centering
  \includegraphics[width=\linewidth]{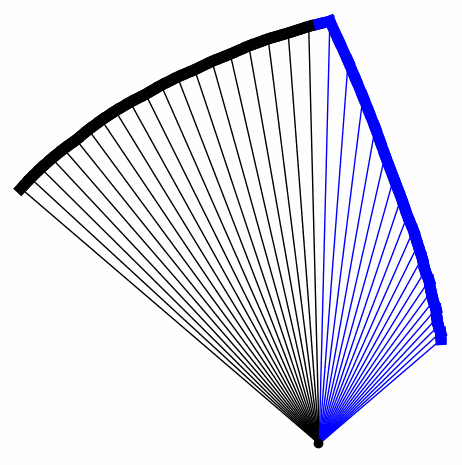}} &
\parbox[c]{\linewidth}{%
  \centering
  \includegraphics[width=\linewidth]{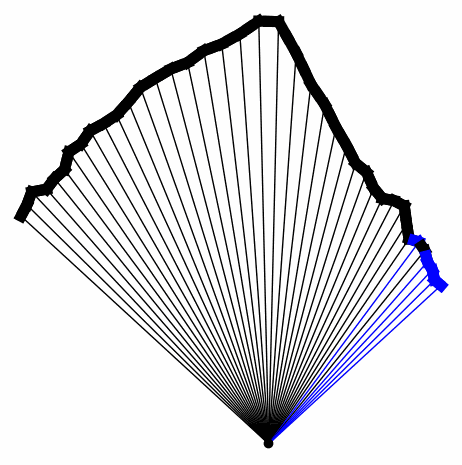}} \\
\noalign{\vskip 3pt}
\parbox[c]{\linewidth}{%
  \centering
  \includegraphics[width=\linewidth]{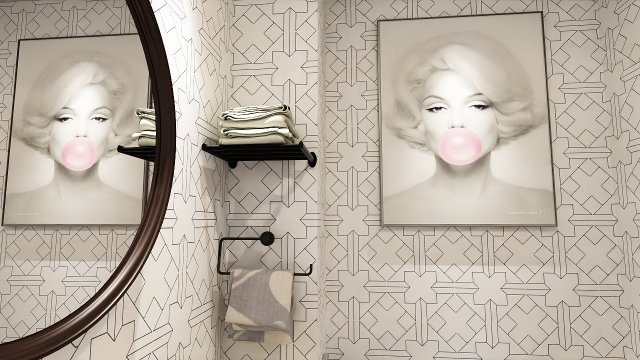}} &
\parbox[c]{\linewidth}{%
  \centering
  \includegraphics[width=\linewidth]{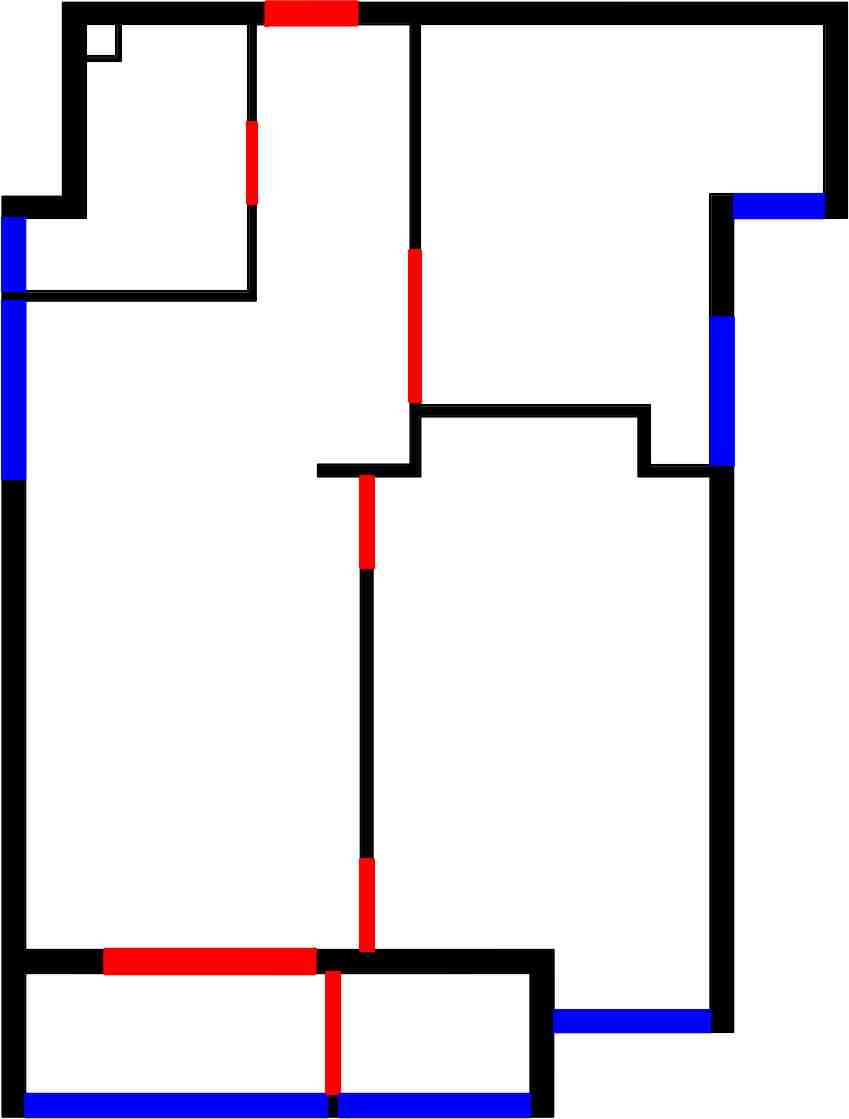}} &
\parbox[c]{\linewidth}{%
  \centering
  \includegraphics[width=\linewidth]{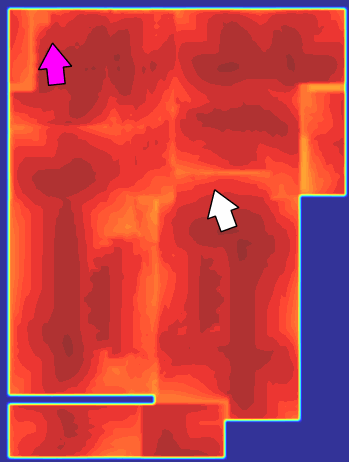}} &
\parbox[c]{\linewidth}{%
  \centering
  \includegraphics[width=\linewidth]{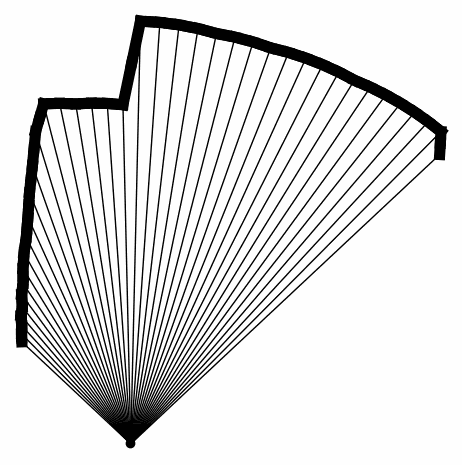}} &
\parbox[c]{\linewidth}{%
  \centering
  \includegraphics[width=\linewidth]{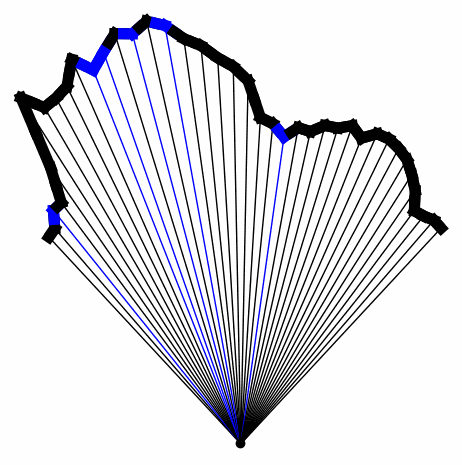}} \\
\noalign{\vskip 3pt}
\parbox[c]{\linewidth}{%
  \centering
  \includegraphics[width=\linewidth]{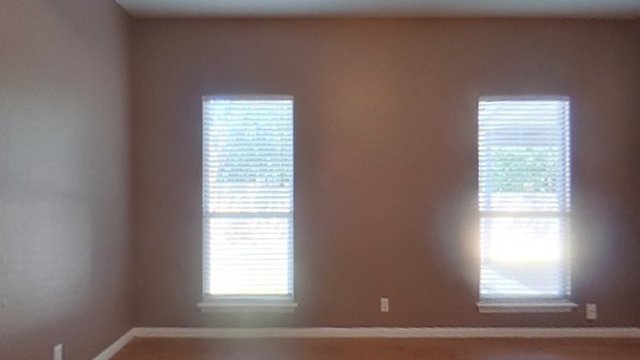}} &
\parbox[c]{\linewidth}{%
  \centering
  \includegraphics[width=\linewidth]{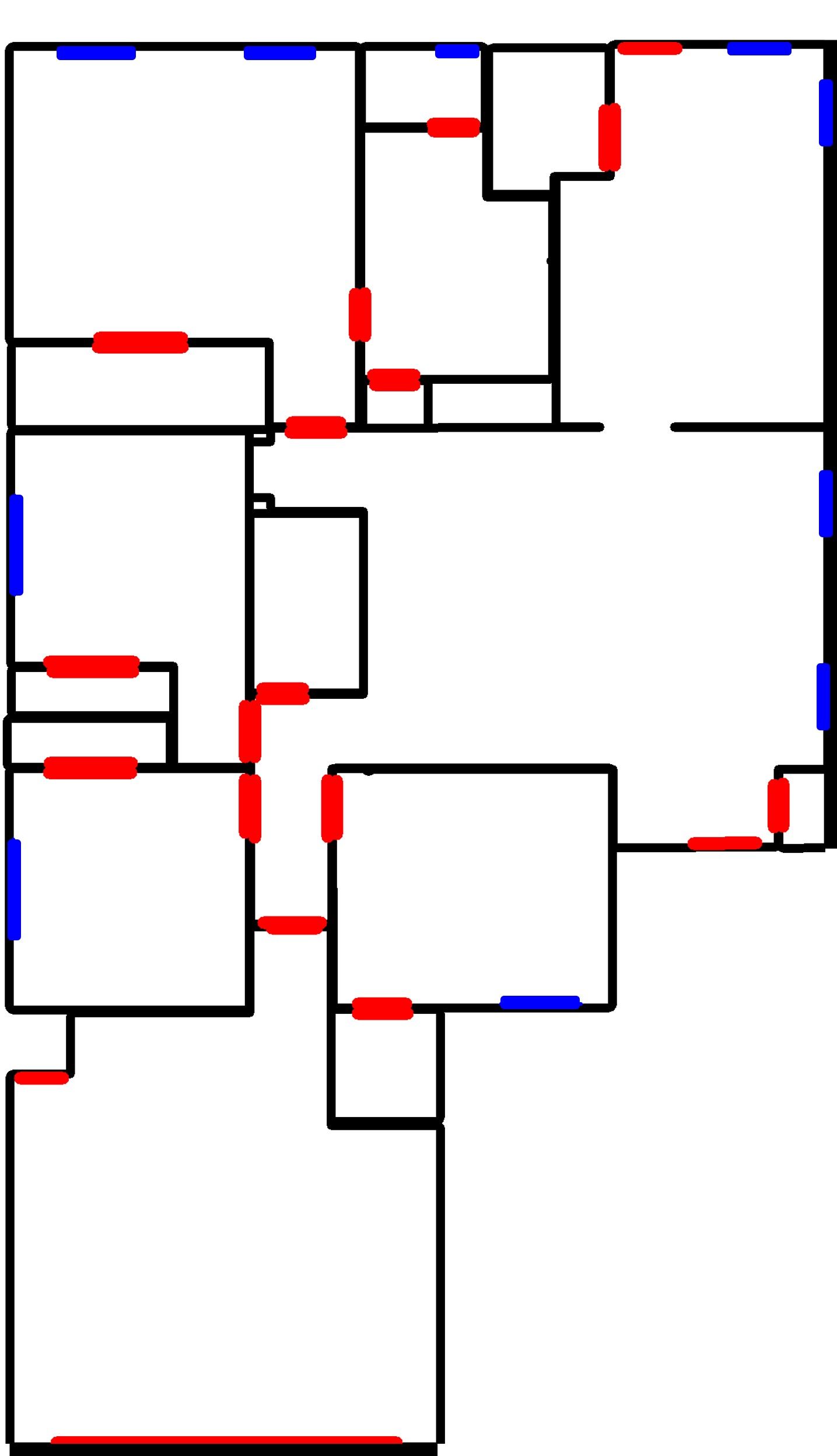}} &
\parbox[c]{\linewidth}{%
  \centering
  \includegraphics[width=\linewidth]{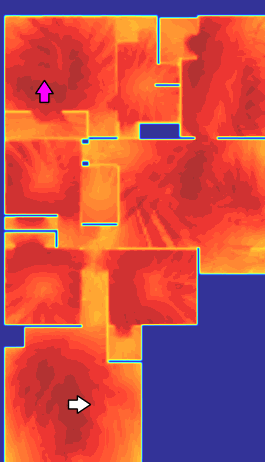}} &
\parbox[c]{\linewidth}{%
  \centering
  \includegraphics[width=\linewidth]{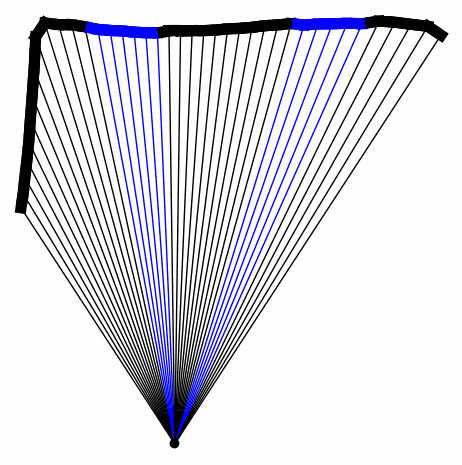}} &
\parbox[c]{\linewidth}{%
  \centering
  \includegraphics[width=\linewidth]{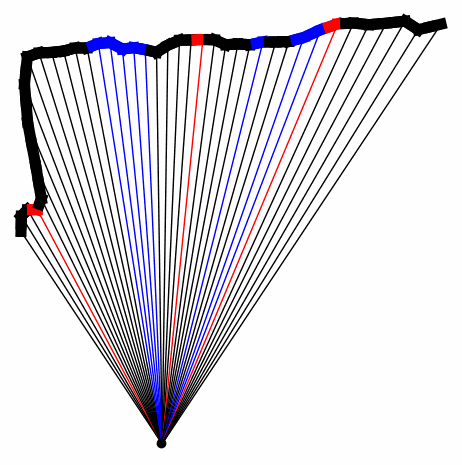}} \\
\noalign{\vskip 3pt}
\parbox[c]{\linewidth}{%
  \centering
  \includegraphics[width=\linewidth]{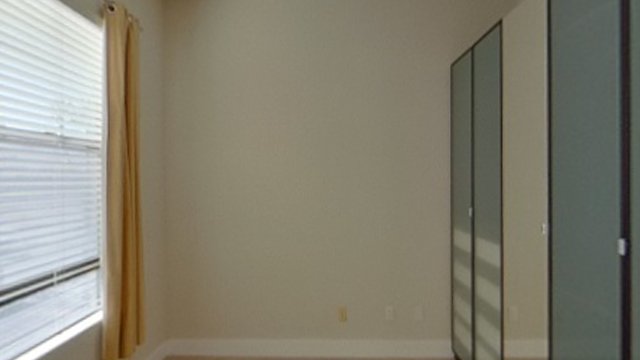}} &
\parbox[c]{\linewidth}{%
  \centering
  \includegraphics[width=\linewidth]{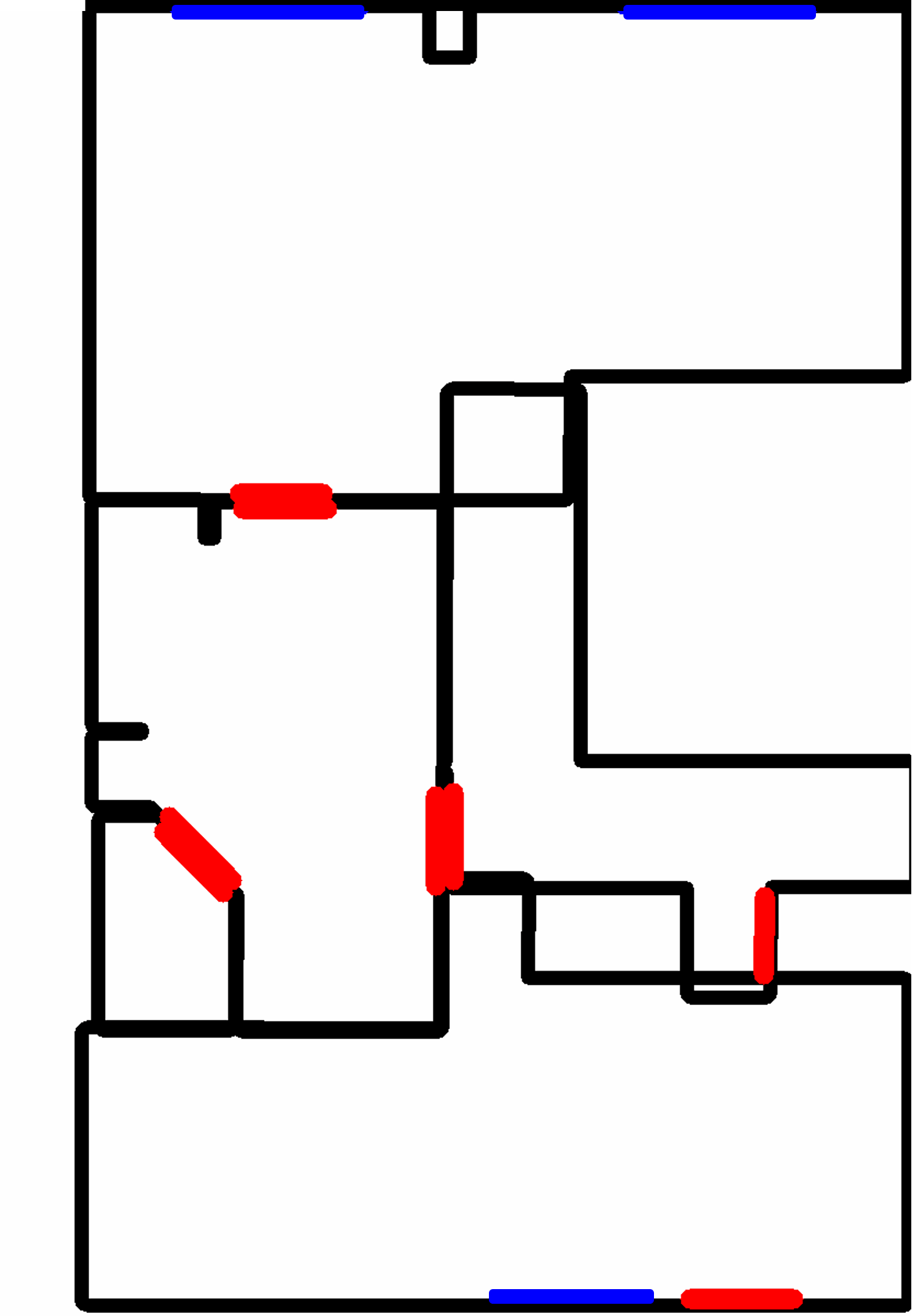}} &
\parbox[c]{\linewidth}{%
  \centering
  \includegraphics[width=\linewidth]{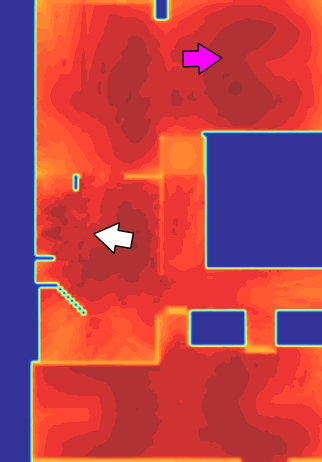}} &
\parbox[c]{\linewidth}{%
  \centering
  \includegraphics[width=\linewidth]{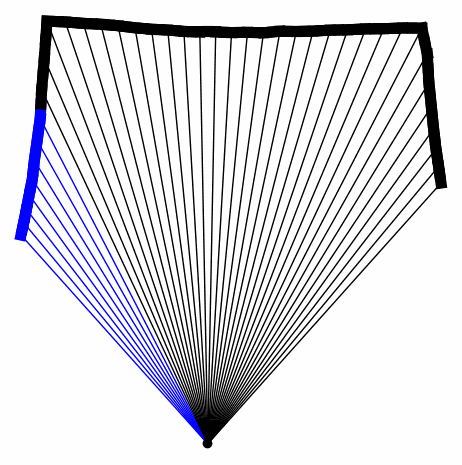}} &
\parbox[c]{\linewidth}{%
  \centering
  \includegraphics[width=\linewidth]{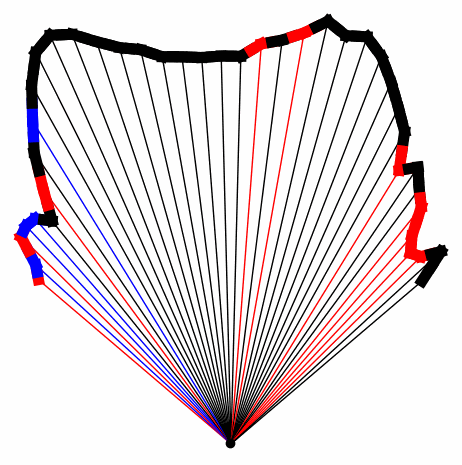}} \\
\noalign{\vskip 3pt}
\hline
\end{tabular}

        \caption{
        \textbf{Limitations}. Above we show several failure cases, where semantic misclassifications and structural ambiguities lead to localization errors; see Section \ref{sec:limitation} for additional details. Warmer colors again represent higher probabilities. \textcolor{magenta}{Magenta} marks the ground truth, and \textcolor{black}{white} indicates the estimated layout.  Rays are: \textcolor{black}{wall}, \textcolor[HTML]{0000FF}{window}, and \textcolor[HTML]{FF0000}{door}.
        }        
    \label{fig:bad_examples}
\end{figure*}

Figure~\ref{fig:bad_examples} illustrates several failure cases from both of the datasets where our approach struggles. In these examples, misclassifications of certain semantic labels or confusions between visually similar features, such as interpreting a window as a door (row 1) or mistaking the window size (row 3), can lead to localization errors. The figure displays both the ground truth rays and the predicted rays, highlighting the differences and emphasizing the critical role of precise semantic inference for robust indoor localization.

These limitations suggest that improvements in semantic segmentation and more sophisticated feature disambiguation techniques could enhance performance. We believe that addressing these issues can lead to further improvements in localization accuracy in future work.

\end{document}